\documentclass{article}

\usepackage[final]{_neurips2023/neurips_2023}

\usepackage{microtype}
\usepackage{graphicx}
\usepackage{subfigure}
\usepackage{booktabs} 

\usepackage{hyperref}

\usepackage{amsmath}
\usepackage{amssymb}
\usepackage{mathtools}
\usepackage{amsthm}

\usepackage[capitalize,noabbrev]{cleveref}

\theoremstyle{plain}
\newtheorem{theorem}{Theorem}[section]

\newtheorem{lemma}[theorem]{Lemma}

\theoremstyle{definition}

\newtheorem{assumption}[theorem]{Assumption}
\theoremstyle{remark}

\usepackage[textsize=tiny]{todonotes}

\newcommand{\ourmethod}[0]{HopCPT\xspace}
\newcommand{\enbpi}[0]{EnbPI\xspace}
\newcommand{\nexcp}[0]{NexCP\xspace}
\newcommand{\spic}[0]{SPCI\xspace}
\newcommand{\adaptiveci}[0]{AdaptiveCI\xspace}
\newcommand{\knn}[0]{kNN\xspace}

\newcommand{\repourl}{\url{https://github.com/ml-jku/HopCPT}}

\newcommand{\nextTableCaption}{}
\newcommand{\setTableCaption}[1]{\renewcommand{\nextTableCaption}{#1}}

\newcommand{\solarSmall}[0]{Solar (3M)\xspace}
\newcommand{\solarThreeY}[0]{Solar (3Y)\xspace}
\newcommand{\solarOneY}[0]{Solar (1Y)\xspace}
\newcommand{\airTen}[0]{Air Quality (10PM)\xspace}
\newcommand{\airTwenty}[0]{Air Quality (25PM)\xspace}
\newcommand{\sapflux}[0]{Sap flow\xspace}
\newcommand{\hydro}[0]{Streamflow\xspace}

\newcommand{\NRM}[1]{{{\left\| #1\right\|}}}
\newcommand{\soft}{\mathrm{softmax}}

\newcommand{\independent}{\perp\!\!\!\!\perp}

\newcommand{\muh}{\hat{\mu}}
\newcommand{\Ch}{\widehat{C}}
\newcommand{\quant}{\mathsf{Q}}
\newcommand{\dtv}{\mathsf{d}_{\mathsf{TV}}}

\newcommand{\PR}[1]{\mathbf{\mathrm{Pr}}\left\{{#1}\right\}}

\newcommand\Ba{\bm{a}}

\newcommand\Bx{\bm{x}}

\newcommand\Bz{\bm{z}}

\newcommand\BA{\bm{A}}

\newcommand\BW{\bm{W}}
\newcommand\BX{\bm{X}}

\newcommand\BZ{\bm{Z}}

\newcommand\Bep{\bm{\epsilon}}

\newcommand\Bxi{\bm{\xi}}

 \newcommand{\dR}{\mathbb{R}}

\newcommand{\rC}{\mathrm{C}}  
\newcommand{\rE}{\mathrm{E}}

\newcommand\EXP[1]{\mathbb{E}\left[{#1}\right]}

\newcommand\argmin{\mathop{\mathrm{argmin}\,}}

\newcommand\nn{\mathrm{new}}

\usepackage{xcolor}
\usepackage{amsmath,amssymb}
\usepackage{todonotes}
\usepackage{bm}
\usepackage{stmaryrd}
\usepackage{multirow}
\usepackage{tcolorbox}
\usepackage{mathtools}
\usepackage{booktabs}
\usepackage{xspace}
\usepackage{siunitx}
\usepackage{enumitem}

\title{Conformal Prediction for Time Series with \\ Modern Hopfield Networks}

\author{%
    Andreas Auer \footnotemark[1] \quad
    Martin Gauch\footnotemark[1]~$~^{,}$\footnotemark[2] \quad
    Daniel Klotz\footnotemark[1] \quad
    Sepp Hochreiter\footnotemark[1] \\
  \footnotemark[1]~~ELLIS Unit Linz and LIT AI Lab, Institute for Machine Learning,\\ Johannes Kepler University Linz, Austria\\
  \footnotemark[2]~~Google Research, Zürich, Switzerland\\
}

\usepackage{selectp}

\begin{document}

\maketitle


\begin{abstract}

To quantify uncertainty, conformal prediction methods 
are gaining continuously more interest and have already been
successfully applied to various domains.
However, they are difficult to apply to time series 
as the autocorrelative structure of time series 
violates basic assumptions required by conformal prediction.
We propose \ourmethod, a novel conformal prediction approach for time series
that not only copes with temporal structures but leverages them.
We show that our approach is theoretically well justified for
time series where temporal dependencies are present.
In experiments, 
we demonstrate that our new approach outperforms state-of-the-art 
conformal prediction methods on multiple real-world time series datasets from four different domains.
\end{abstract}
\vspace{-4mm}

\section{Introduction}

Uncertainty estimates are imperative to make actionable predictions for complex time-dependent systems \citep[e.g.,][]{gneiting2014probforecasting, zhu2017deep}. 
This is particularly evident for environmental phenomena such as flood forecasting \citep[e.g.,][]{krzysztofowicz2001case}, since they exhibit pronounced seasonality.
Conformal Prediction \citep[CP,][]{vovk1999machine} provides uncertainty estimates based on prediction intervals. 
It achieves finite-sample marginal coverage with almost no assumptions, except that the data is exchangeable \citep{vovk2005algorithmic, vovk2012conditional}. 
However, CP for time series is not trivial because temporal dependencies often violate the exchangeability assumption.

\begin{figure}[t]
\includegraphics[width=13cm]{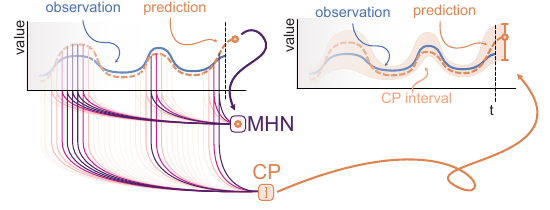}
\centering
\caption{Schematic illustration of \ourmethod. The Modern Hopfield Network (MHN) identifies regimes similar to the current one and up-weights them (colored lines). The weighted information enriches the conformal prediction (CP) procedure so that prediction intervals can be derived.}
\label{fig:figure1}
\end{figure}

\noindent
\textbf{\ourmethod.} \;\,
For time series models that predict a given system, the errors are typically specific to the input describing the current situation.
To apply CP, \ourmethod uses continuous Modern Hopfield Networks (MHNs). 
The MHN uses large weights for past situations that were similar to the current one
and weights close to zero for past situations that were dissimilar to the current one (Figure~\ref{fig:figure1}).
The CP procedure uses the weighted errors to construct strong uncertainty estimates close to the desired coverage level.
This exploits that similar situations (which we call a regime) tend to follow the same error distribution.
\ourmethod achieves new state-of-the-art efficiency, even under non-exchangeability and on large datasets. 

\noindent
\textbf{Our main contributions are:}
\vspace{-1.725mm}
\begin{enumerate}
    \item We propose \ourmethod, a CP method for time series, a domain where CP struggled so far.
    \item We introduce the concept of error regimes to CP, which softens the excitability requirements and enables efficient CP for time series.
    \item \ourmethod uses MHN for a similarity-based sample reweighting.
    In contrast to existing approaches, \ourmethod can learn from large datasets and predict intervals at arbitrary coverage levels without retraining.
    \item \ourmethod achieves state-of-the-art results for conformal time series prediction tasks from various real-world applications.
    \item HopCPT is the first algorithm with coverage guarantees that was applied to hydrological prediction applications --- a domain where uncertainty plays a key role in tasks such as flood forecasting and hydropower management.
\end{enumerate}

\subsection{Related Work}\label{sec:related-work}
\paragraph{Regimes.} In a world with non-linear dynamics, 
different environmental conditions lead to different error characteristics 
of models that predict
based on these conditions.
If we do not account for these different conditions, 
temporal changes may lead to unnecessarily large prediction intervals, i.e., to high uncertainty.
For example, solar energy production is high and stable on a sunny day, 
fluctuates during cloudy days, and is zero at night.
Often, the current environmental condition
was already observed at previous points in time. 
The error at these time steps is therefore assumed to 
have the same distribution as the current error.
Following \citet{quandt1958estimation} and \citet{hamilton1990analysis}, 
we call the sets of time steps with similar 
environmental conditions \textit{regimes}.
Although conditional CP is in general impossible \citep{foygel2021limits}, 
we show that conditioning on such regimes 
can lead to better prediction intervals 
while preserving the specified coverage.
\citet{hamilton1990analysis} models time series regimes as a discrete Markov process and conditions a classical autoregressive model on the regime states.
\citet{sanquer2012smooth} use a smooth transition approach to model multi-regime time series. 
\citet{tajeuna2021modeling} propose an approach to discover and model regime shifts in an ecosystem that comprises multiple time series. 
Further, \citet{masserano2022adaptive} handle distribution shifts by retraining a forecasting model with training data from a non-uniform adaptive sampling.
Although these approaches are not in a CP setting, their work is similar in spirit, as they also follow the general idea to condition on parts of the time series with similar regimes.

\paragraph{CP and extensions.} For thorough introductions to CP, we refer the reader to the foundational work of \citet{vovk1999machine} and a recent introductory paper by \citet{angelopoulos2021gentle}.
There exist a variety of extensions for CP that go ``beyond exchangeability'' \citep{vovk2005algorithmic}.
For example, \citet{papadopoulos2011reliable} apply CP to a nearest neighbor regression setting, \citet{teng2022predictive} apply CP to the feature space of models, \citet{angelopoulos2020uncertainty} use CP to generate uncertainty sets for image classification tasks, and \citet{toccaceli2017conformal} use a label-conditional variant to apply CP to biological activity prediction.
Of specific interest to us is the research regarding non-exchangeable data of  \citet{tibshirani2019conformal} and \citet{barber2022conformal}.
Both handle potential shifts between the calibration and test set by reweighting the data points.
\citet{tibshirani2019conformal} restrict themselves to settings with full knowledge about the change in distribution; \citet{barber2022conformal} rely on fixed weights.
In our work, we refrain from this assumption because such information is typically not available in time series prediction.
Another important research direction is the work on normalized conformity scores \citep[see][and references therein]{fontana2023conformal}. 
In this setting, the goal is to adapt the conformal bounds through a scaling factor in the nonconformity function.
The work on normalized conformity scores does not explicitly tailor their approaches to time series. 

\paragraph{CP for time series.} \citet{gibbs2021adaptive} and \citet{zaffran2022adaptive} account for shifts in sequential data by continuously adapting an internal coverage target.
Adaption-based approaches like these are orthogonal to \ourmethod and can serve as an enhancement.
\citet{stankeviciute2021conformal} use CP in conjunction with recurrent neural networks in a multi-step prediction setting, assuming that the series of observations is independent.
Thus, no weighting of the scores is required. 
\citet{Sun2022CopulaCP} introduce CopulaCPTS which applies CP to time series with multivariate targets.
They conformalize their prediction based on a copula of the target variables and adapt their calibration set in each step.
\citet{jensen2022ensemble} use a bootstrap ensemble to enable CP on time series. 
\nexcp \citep{barber2022conformal} uses exponential decay as the weighting method, arguing that the recent past is more likely to be of the same error distribution. 
\ourmethod can learn this strategy, but does not a priori commit to it. 
\citet{xu2021conformal} propose \enbpi, which uses quantiles of the $k$ most recent errors for the prediction interval.
Additionally, they introduce a novel leave-one-out ensembling technique.
This is specifically geared to settings with scarce data and difficult to use for larger datasets, which is why we do not apply it in our experiments. 
\enbpi is designed around the notion that near-term errors are often independent and identically distributed and therefore exchangeable.
\spic \citep{xu2022sequential} softens this requirement by exploiting the autocorrelative structure with a random forest.
However, it re-calculates the random forest model at each time step, which is a computational burden that prohibits its application to large datasets.
Our approach relaxes the requirement even further, as we do not assume that the data for the interval computations pertains to the $k$ most recent errors. 

\paragraph{Non-CP methods} Beside CP there exist a wide range of approaches for uncertainty-aware time series prediction.
For example, Mixture Density Networks \citep{bishop1994mixture} directly estimate the parameters of a mixture of distributions.
However, they require a distribution assumption and do not provide any theoretical guarantees.
Gaussian Processes \citep[e.g.,][]{zhu2023bayesian, corani2021time, sun2022recurrent} model time series by calculating posterior functions based on the samples and a prior but are computationally limited for large and high dimensional datasets.

\paragraph{Continuous Modern Hopfield Networks.} 
MHN are energy-based associative memory networks.
They advance conventional Hopfield Networks \citep{Hopfield:82} by 
introducing continuous queries and states via a new energy function.
The new energy function leads to exponential storage capacity, 
while retrieval is possible with a one-step update \citep{Ramsauer:21}.
Examples for successful applications of MHN are \citet{Widrich:20nips, furst2021cloob,  dong2022retrosynthesis, klambauer2022cloome, paischer2022history, schafl2022hopular}; and \citet{xu2022txt2img}. 
MHN are related to Transformers \citep{Vaswani:17} as their attention mechanism is closely related to the association mechanism in MHN.
In fact, \citet{Ramsauer:21} show that Transformers attention is a special case of the MHN association, at which we arrive when the queries and states are mapped to an associative Hopfield space with the dimensions $d_k$, and the inverse softmax temperature is set to $\beta = \frac{1}{\sqrt{d_k}}$.
However, we use the framework of MHN because we want to highlight the associative memory mechanism as \ourmethod directly ingests encoded observations.
This perspective further allows \ourmethod to update the memory for each new observation.
For more details, we refer to Appendix~\ref{sec:appendix-hopfield} in the supplementary material.

\subsection{Setting}\label{sec:setting}

Our setting consists of a multivariate time series $\{(\Bx_t, y_t)\}$, $t=1,\dotsc,T$, with a feature vector $\Bx_t \in \dR^m$, a target variable $y_t \in \dR$, and a given black-box prediction model $\muh$ that generates a point prediction $\hat{y}_t = \muh(\BX_t)$.
The input feature matrix $\BX_{t+1}$ can include all previous and current feature vectors $\{\Bx_i\}^{t+1}_{i=1}$, as well as all previous targets $\{y_i\}^{t}_{i=1}$.
Our goal is to construct a corresponding prediction interval $\Ch_t^{\alpha}(\BZ_{t+1})$ --- a set that includes $y_{t+1}$ with at least a specified probability $1 - \alpha$.
In its basic form, $\BZ_{t+1}$ will only contain  $\hat{y}_{t+1}$, but it can also inherit $\BX_{t+1} $ or other useful features.
Following \citet{vovk2005algorithmic}, we define the \textit{coverage} as
\begin{equation}
\PR{Y_{t+1} \in \Ch_t^{\alpha}(\BZ_{t+1})} \geq 1-\alpha,
\label{eq:cp_prop}
\end{equation}

where $Y_{t+1}$ is the random variable of the prediction.
An infinitely wide prediction interval is 100\% reliable, but not informative of the uncertainty.
Thus, CP aims to minimize the width of the prediction interval $\Ch^{\alpha}_t$, while preserving the coverage.
A smaller prediction interval is called a more \textit{efficient} interval \citep{vovk2005algorithmic} and usually evaluated as the mean of the interval width over the prediction period (\textit{PI-Width}).

Standard split conformal prediction takes a calibration set of size $n$ which has not been used to train the prediction model $\muh$.
For each data sample, it calculates the so-called non-conformity score \citep{vovk2005algorithmic}.
In a regression setting, this score often simply corresponds to the absolute error of the prediction \citep[e.g.,][]{barber2022conformal}.
The prediction interval is then calculated based on the empirical $1-\alpha$ quantile $\quant_{1-\alpha}$ of the calibration scores:
\begin{equation}
    \Ch_n^{\alpha}(\BZ_{t+1}) = \muh(\BX_{t+1}) \pm \quant_{1-\alpha} (\{|y_i - \muh(\BX_i)|\}_{i=1}^n).
\end{equation}

If the data is exchangeable and $\muh$ treats the data points symmetrically, the errors on the test set follow the distribution from the calibration.
Hence, the empirical quantiles on the calibration and test set will be approximately equal and it is guaranteed that the interval provides the desired coverage. 

The \textit{actual marginal miscoverage $\alpha^\star$} is based on the observed samples of a test set.
If the \textit{specified miscoverage} $\alpha$ differs from the $\alpha^\star$ in the evaluation, we denote the difference as the coverage gap $ \Delta \ \text{Cov} = \alpha - \alpha^\star$.

The remainder of this manuscript is structured as follows: 
In Section~\ref{sec:our-method}, we present \ourmethod alongside a theoretical motivation and a synthetic example that demonstrates the advantages of the approach.
In Section~\ref{sec:experiments}, we evaluate the performance against state-of-the-art CP approaches and discuss the results.
Section~\ref{sec:conclusion} gives our conclusions and provides an outlook on potential future work.

\section{\ourmethod}\label{sec:our-method}
\ourmethod\footnote{\repourl} combines conformal-style quantile estimation with learned similarity-based MHN retrieval.

\subsection{Theoretical Motivation}\label{sec:theory}

%
%
The theoretical motivation for the MHN retrieval stems from  \citet{barber2022conformal}, who introduced CP with weighted quantiles.
In the split conformal setting, the according prediction interval is calculated as
\begin{equation}\label{eq:cov-bound-interval}
    \Ch_t^{\alpha}(\BX_{t+1}) = \muh(\BX_{t+1}) \pm \quant_{1-\alpha} \left(\sum_{i=1}^t a_i \delta_{\epsilon_i} + a_{t+1} \delta_{+\infty} \right),
\end{equation}

where $\muh$ represents an existing point prediction model, $\quant_{\tau}$ is the $\tau$-quantile of a distribution, and $\delta_{\epsilon_i}$ is a point mass at $|\epsilon_i|$ (i.e., a probability distribution with all its mass at $|\epsilon_i|$), where $\epsilon_i$ are the errors of the existing prediction model defined by $ \epsilon_i = y_i - \muh(\BX_{i})$.
The normalized weight $a_i$ of data sample $i$ is
\begin{equation}\label{eq:weighted-a}
    a_i = \begin{cases}
        \frac{1}{\omega_1 + \dotsc + \omega_t + 1} & \text{if} \ i = t+1, \\
        \frac{\omega_i}{\omega_1 + \dotsc + \omega_t + 1} & \text{else},
    \end{cases}
\end{equation}

where $\omega_i$ are the un-normalized weights of the samples.
In the case of $\omega_1 = \dotsc = \omega_t = 1$, this corresponds to standard split CP.
Given this framework, \citet{barber2022conformal} show that $\Delta \ \text{Cov}$ can be bounded in a non-exchangeable data setting:
Let $D=((\BX_1, Y_1), \dotsc, (\BX_{t+1}, Y_{t+1}))$ be a dataset where the last entry represents the test sample, and $D^i$ be a permutation of $D$ which exchanges the test sample at $t+1$ with the $i$-th sample.
Then, $\Delta \ \text{Cov}$ can be bounded from below by the weighted sum of the total variation distances $d_\text{TV}$ between these permutations:
\vspace{-1mm}
\begin{equation}\label{eq:cov-bound}
    \Delta \ \text{Cov} \geq - \sum_{i=1}^t a_i \cdot \dtv(D,D^i)
\end{equation}

%
%
\vspace{-1mm}
If $D$ is a composite of multiple regimes and the test sample is from the same regime as the calibration sample $i$, then the distance between $D$ and $D^i$ is small.
Conversely, the distance might be big if the calibration sample is from a different regime.
In \ourmethod, the MHN association resembles direct estimates of $a_i$ --- dynamically assigning high values to samples from similar regimes.

Appendix~\ref{sec:theory-appendix} provides an extended theoretical discussion that relates \ourmethod to the work of \cite{barber2022conformal} and \citet{xu2021conformal}, who provide the basis for our theoretical analyses of the CP interval.
The latter compute individual quantiles for the upper and lower bound of the prediction interval on the errors themselves, in contrast to standard CP which uses only the highest absolute error quantile.
This can provide more efficient intervals in case $E[\epsilon] \neq 0$ within the corresponding error distribution.
Applying this principle to \ourmethod where the error distribution is conditional to the error regime and assuming that \ourmethod can successfully identify these regimes (Appendix: Assumption~\ref{app:local-error-assumption}), we arrive at a conditional asymptotic coverage bound (Appendix: Theorem~\ref{thm:cond_cov_xu}) and an asymptotic marginal coverage bound (Appendix: Theorem~\ref{thm:marg_cov_xu}).

\subsection{Associative Soft-selection for CP}\label{sec:mhn}
We use a MHN to identify parts of the time series where the conditional error distribution is similar:
For time step $t+1$, we query the memory of the past and look for matching patterns.
The MHN then provides an association vector $\Ba_{t+1}$ that allows to soft-select the relevant periods of the memory.
The selection procedure is analogous to a $k$-nearest neighbor classifier for hard selection, but it has the advantage that the similarity measure can be learned. Formally, the soft-selection is defined as:
\begin{equation}\label{eq:hopfield-association}
    \Ba_{t+1} = \text{softmax}\big(\beta \, m(\BZ_{t+1}) \, \BW_q \, \BW_k \, m(\BZ_{1:t}) \big),
\end{equation}

where $m$ is an encoding network (Section~\ref{sec:encoding-network}) that transforms the raw time series features of the current step $\BZ_{t+1}$ to the query pattern and the memory $\BZ_{1:t}$ to the stored key patterns;
$\BW_q$ and $\BW_k$ are the learned transformations which are applied before associating the query with the memory;
$\beta$ is a hyperparameter that controls the softmax temperature. 
As mentioned above, \ourmethod uses the softmax to amplify the impact of the data samples that are likely to follow a similar error distribution and to reduce the impact of samples that follow different distributions (see Section~\ref{sec:theory}).
This error weighting leads not only to more efficient prediction intervals in our experiments, but can also reduce the miscoverage (Section~\ref{sec:results-discuss}).

With the soft-selected time steps we can derive the CP interval using the observed errors $\epsilon$.
Following \citet{xu2021conformal}, we use individual quantiles for the upper and lower bound of the prediction interval, calculated from the errors themselves.
\ourmethod computes the prediction interval $\Ch^{\alpha}_t$ for time step $t+1$ in the following way:
\begin{equation}\label{eq:myconfbounds}
\begin{aligned}
\Ch_t^{\alpha}(\BZ_{t+1}) = %
   \Bigl[ & \muh(\BX_{t+1}) + q(\frac{\alpha}{2}, \BZ_{t+1}), \muh(\BX_{t+1}) + q(1-\frac{\alpha}{2}, \BZ_{t+1}) \Bigr],
\end{aligned}
\end{equation}

where $q(\tau, \BZ_{t+1}) = \quant_{\tau} \left(E_{t+1} \right)$ and $E_{t+1}$ is a multiset created by drawing $n$ times from $[\epsilon_i]_{i=1}^t$ with corresponding probabilities $[a_{t+1,i}]_{i=1}^t$.

\subsection{Encoding Network}\label{sec:encoding-network}

We embed $\BZ_t$ using a 2-layer fully connected network $m^L$ with ReLU activations and enhance the representation by temporal encoding features $\Bz_{T,t}^{\text{time}}$. The full encoding network can be represented as $m(\BZ_t)=[m^L(\BZ_t) \ || \ \Bz_{T,t}^{\text{time}}]$, 
where $\Bz_{T,t}^{\text{time}}$ is a simple temporal encoding that makes time dependent notions of similarity learnable (e.g., the windowed approach of \enbpi or the exponentially decaying weighting scheme of \nexcp). Specifically, we use $z_{T,t}^{\text{time}} = \frac{t}{T}$.
In general, we find that our method is not very sensitive to the exact encoding strategy (e.g., number of layers), which is why we kept to a simple choice throughout all experiments.


\subsection{Training Procedure}

We partition the split conformal calibration data into training and validation sets.
Training MHN with quantiles is difficult, which is why we use an auxiliary task:
Instead of applying the association mechanism from Equation~\ref{eq:hopfield-association} directly, we use the absolute errors as the value patterns of the MHN.
This way, the MHN learns to align errors from time steps with similar regime properties.
Intuitively, the observed errors from these time steps should work best to predict the current absolute error.
We use the mean squared error as loss function (Equation~\ref{eq:train-loss}).
To allow for efficient training, we simultaneously calculate the association of all $T$ time steps within a training split with each other.
We mask the association from a time step to itself.
The resulting association from each step to each step is $\BA_{1:T,1:T}$, and the loss function $\mathcal{L}$ is
\begin{equation}\label{eq:train-loss}
    \mathcal{L} = T^{-1} * \lVert(|\bm{\epsilon}_{1:T}| - \BA_{1:T,1:T}  |\bm{\epsilon}_{1:T}|)^2\rVert_1.
\end{equation}

This incentivizes the network to learn a representation such that the resulting soft-selection focuses on time steps with a similar error distribution.
Alternatively, one could use a loss based on sampling from the softmax.
This would correspond more closely to the inference procedure.
However, it makes training less efficient because each training step only carries information about a single value of $\epsilon_i$. 
In contrast, $\mathcal{L}$ leads to more sample-efficient training.
Choosing $\mathcal{L}$ assumes that it leads to a retrieval of errors from appropriate regimes instead of mixing unrelated errors.
Section~\ref{sec:results-discuss} and Appendix~\ref{sec:appendix-full-tables} provide evidence that this holds empirically.


\subsection{Synthetic Example}\label{sec:synthetic-example} 
\begin{figure}[t]
\includegraphics[width=13.5cm]{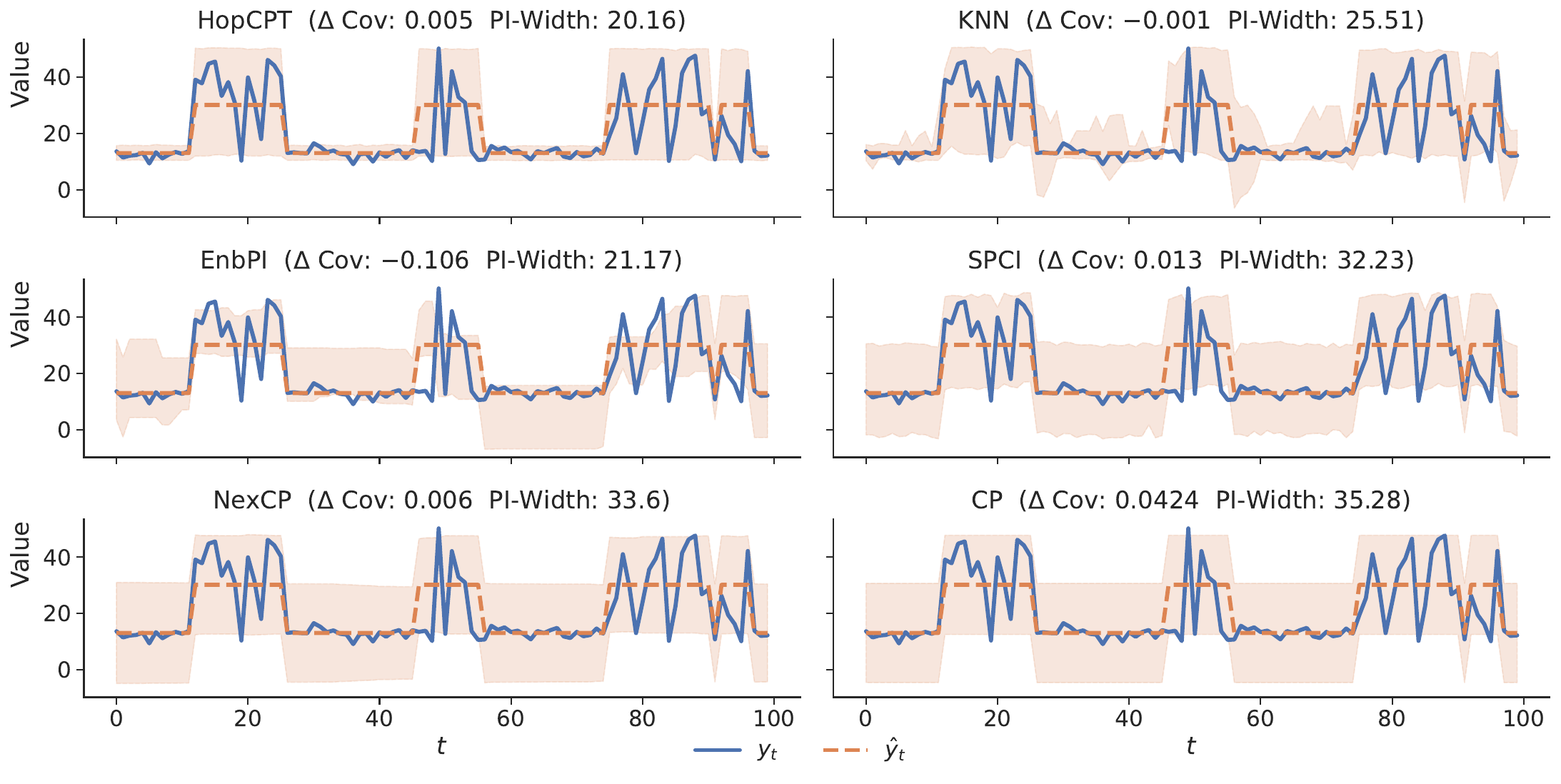}
\centering
\caption{Synthetic example evaluation. \ourmethod has the smallest  prediction interval width (PI-width), while maintaining the warranted coverage (i.e., a $\Delta \ \text{Cov}$ that is positive and close to zero).}
\label{fig:toy-example}
\end{figure}

The following synthetic example illustrates the advantages of the \ourmethod association mechanism.
We model a bivariate time series $D=\{(x_t, y_t)\}^T_{t=1}$.
While $y_t$ serves as the target variable, $x_t$ represents the feature in our prediction.
The series is composed of two different regimes:
target values are generated by $y_t=10+x_t+\mathcal{N}(0,\frac{x_t}{2})$ or $y_t=10 + x_t + U(-x_t,x_t)$.
$x_t$ is constant within a regime ($x=3$ and $x=21$, respectively).
The regimes alternate. 
For each regime, we sample the number of time steps from the discrete uniform distribution $\mathcal {U}(1,25)$.
We create 1,000 time steps and split the data equally into training, calibration, and test sets.
We use a ridge regression model as prediction model $\muh$.
\ourmethod can identify the time steps of  relevant regimes and therefore creates efficient prediction intervals while still preserving the coverage (Figure~\ref{fig:toy-example}).
\enbpi, \spic, and \nexcp focus only on the recent time steps and thus fail to base their intervals on information from the correct regime.
Whenever the regime changes from small to high errors, \enbpi propagates the small error signal and therefore loses coverage.
Similarly, its prediction intervals for the small error regime are inefficient (Figure~\ref{fig:toy-example}, row 3).
\spic, \nexcp, and standard CP cannot properly select the relevant time steps, either.
They do not lose coverage, but produce wide intervals for all time steps (Figure~\ref{fig:toy-example}, rows 4 and 5).
Lastly, if we replace the MHN with a \knn, it can retrieve information from similar regimes. However, its naive retrieval mechanism fails to focus on the informative features because it cannot learn them (Figure~\ref{fig:toy-example}, row 2).

\subsection{Limitations}
\ourmethod builds upon the formal guarantees of CP, but it is able to relax the data exchangeability assumption of CP which are problematic for time series data.
That said, \ourmethod still relies on assumptions on how it learns to identify the respective error regimes and exchangablity within regimes (see Section~\ref{sec:theory} and Appendix~\ref{sec:theory-appendix}).
Our extensive empirical evaluation suggests that these assumptions hold in practice.
\ourmethod is generally well-suited for very long time series because the memory requirement in the inference scales only linearly with the memory size.
Nevertheless, for extremely large datasets it can be the case that not all historical time steps may be kept in the Hopfield memory anymore.
\ourmethod can, similar to existing methods, disregard time steps by removing the oldest entries from the memory, or alternatively use a sub-sampling strategy.
On the other hand, for datasets with very scarce data it might be difficult to learn a useful embedding of the time steps.
In that case, it could be better to use kNN rather than learning the similarity with MHN (Appendix \ref{sec:knn-comparison}).
Additional considerations regarding the potential social impact of the method are summarized in Appendix~\ref{sec:social-impact}.

\section{Experiments}\label{sec:experiments} 
This section provides a comparative evaluation of \ourmethod and analyzes its association mechanism.

\subsection{Setup}

\paragraph{Datasets.}
We use datasets from four different domains:
(a) Three solar radiation datasets from the US National Solar Radiation Database \citep{sengupta2018national}.
The smallest one consists of 8 time series from different locations over a period of 84 days.
This dataset is also used in \citet{xu2021conformal, xu2022sequential}.
In addition, we evaluate on a 1-year and a 3-year dataset, with 50 time series each. 
(b) An air quality dataset from Beijing, China \citep{zhang2017cautionary}.
It consists of 12 time series, each from a different measurement station, over a period of 4 years.
The dataset has two prediction targets, the PM10 \citep[as in][]{xu2021conformal, xu2022sequential} and PM2.5 concentrations, which we evaluate separately.  
(c) Sap flow\footnote{Sap flow refers to the movement of water within a plant. In the environmental sciences, it is commonly used as a proxy for plant transpiration.} measurements from the Sapfluxnet data project \citep{poyatos2021global}.
Since the individual measurement series are considerably heterogeneous in length, we use a subset of 24 time series, each with between 15,000 and 20,000 data points and varying sampling rates.
(d) \hydro, a dataset of water flow measurements and corresponding meteorologic observations from 531 rivers across the continental United States \citep{newman2015development,addor2017camels}.
The measurements span 28 years at a daily time scale.
For more detailed information about the datasets see Appendix~\ref{sec:dataset-details}.

\paragraph{Prediction models.}
We use four prediction models for the solar radiation, the air quality, and the sap flux datasets to ensure that our results generalize:
a random forest, a LightGBM, a ridge regression, and a Long Short-Term Memory (LSTM) \citep[LSTM;][]{Hochreiter:97} model.
For the former three models we follow the related work \citep{xu2021conformal,xu2022sequential, barber2022conformal} and train a separate prediction model for each individual time series.
The random forest and LightGBM models are implemented with the darts library \citep{JMLR:v23:21-1177}, the ridge regression model with sklearn \citep{scikit-learn}.
For the LSTM model, we instead train a global model on all time series of a dataset, as is standard for state-of-the-art deep learning models \citep[e.g.,][]{Oreshkin2020N-BEATS:, salinas2020deepar, smyl2020hybrid}.
The LSTM is implemented with PyTorch \citep{paszke2019pytorch}.
For the streamflow dataset, we deviate from this scheme and instead only use the state-of-the-art model, which is also an LSTM network \citep[][see Appendix~\ref{sec:dataset-details}]{kratzert2020note}.

\setTableCaption{Performance of the evaluated CP algorithms for the \solarThreeY, \airTen, \sapflux, and \hydro (SF) datasets. The specified miscoverage level is $\alpha=0.1$ for all experiments. The column FC specifies the prediction algorithm used for the experiment (Forest: Random Forest, LGBM: LightGBM, Ridge: Ridge Regression, LSTM: LSTM neural network). Bold numbers correspond to the best result for the respective metric in the experiment (PI-Width and Winkler score), given that $\Delta \text{Cov} \ge - 0.25 \alpha$, i.e., the specific algorithm reached at least approximate coverage (the result is grayed otherwise). The error term represents the standard deviation over repeated runs with different seeds (results without an error term are from deterministic models).}


\begin{table*}
\centering
\caption{\nextTableCaption}
\label{tab:main-results-alpha-0.1}
\vskip 0.15in
\small
\begin{tabular}{lllrrrrrr|r}
\toprule
 &  & \text{UC} & \rotatebox{90}{\text{\ourmethod}} & \rotatebox{90}{\text{SPCI}} & \rotatebox{90}{\text{EnbPI}} & \rotatebox{90}{\text{NexCP}} & \rotatebox{90}{\text{CopulaCPTS}} & \rotatebox{90}{\text{CP/CF-RNN}} & \rotatebox{90}{\text{AdaptiveCI}} \\
\text{} & \text{} &  &  &  &  &  &  &  &  \\
\midrule
\multirow[c]{12}{*}{\rotatebox{90}{\text{Solar 3Y}}} & \multirow[c]{3}{*}{\rotatebox{90}{\text{Forest}}} & \text{$\Delta$ Cov} & $0.029^{\pm 0.012}$ & $0.012^{\pm 0.000}$ & $\textcolor{gray}{\text{-}0.031}$ & $\text{-}0.002$ & $0.005$ & $0.004$ &   \\
\rotatebox{90}{} &  & \text{PI-Width} & $\textbf{39.0}^{\pm 6.2}$ & $103.1^{\pm 0.1}$ & $\textcolor{gray}{131.1}$ & $166.6$ & $174.9$ & $174.6$ &   \\
\rotatebox{90}{} &  & \text{Winkler} & $\textbf{0.73}^{\pm 0.20}$ & $1.74^{\pm 0.00}$ & $\textcolor{gray}{2.47}$ & $2.53$ & $2.75$ & $2.76$ &   \\
\cmidrule{2-10}
\rotatebox{90}{} & \multirow[c]{3}{*}{\rotatebox{90}{\text{LGBM}}} & \text{$\Delta$ Cov} & $0.001^{\pm 0.003}$ & $0.014^{\pm 0.000}$ & $\text{-}0.023$ & $\text{-}0.002$ & $0.006$ & $0.006$ & $0.001$ \\
\rotatebox{90}{} &  & \text{PI-Width} & $\textbf{37.7}^{\pm 0.7}$ & $102.2^{\pm 0.1}$ & $133.6$ & $159.9$ & $169.8$ & $170.2$ & $67.1$ \\
\rotatebox{90}{} &  & \text{Winkler} & $\textbf{0.57}^{\pm 0.01}$ & $1.75^{\pm 0.00}$ & $2.52$ & $2.55$ & $2.80$ & $2.81$ & $1.19$ \\
\cmidrule{2-10}
\rotatebox{90}{} & \multirow[c]{3}{*}{\rotatebox{90}{\text{Ridge}}} & \text{$\Delta$ Cov} & $0.040^{\pm 0.001}$ & $0.002^{\pm 0.000}$ & $\textcolor{gray}{\text{-}0.074}$ & $\text{-}0.001$ & $0.004$ & $0.005$ &   \\
\rotatebox{90}{} &  & \text{PI-Width} & $\textbf{44.9}^{\pm 0.5}$ & $108.2^{\pm 0.0}$ & $\textcolor{gray}{131.1}$ & $171.0$ & $166.0$ & $167.7$ &   \\
\rotatebox{90}{} &  & \text{Winkler} & $\textbf{0.64}^{\pm 0.00}$ & $1.82^{\pm 0.00}$ & $\textcolor{gray}{2.49}$ & $2.66$ & $2.73$ & $2.74$ &   \\
\cmidrule{2-10}
\rotatebox{90}{} & \multirow[c]{3}{*}{\rotatebox{90}{\text{LSTM}}} & \text{$\Delta$ Cov} & $0.001^{\pm 0.006}$ & $0.014^{\pm 0.000}$ & $\text{-}0.018$ & $\text{-}0.001$ & $0.007$ & $0.007$ &   \\
\rotatebox{90}{} &  & \text{PI-Width} & $\textbf{17.9}^{\pm 0.6}$ & $27.8^{\pm 0.0}$ & $24.6$ & $28.2$ & $31.9$ & $33.0$ &   \\
\rotatebox{90}{} &  & \text{Winkler} & $\textbf{0.30}^{\pm 0.01}$ & $0.62^{\pm 0.00}$ & $0.64$ & $0.63$ & $0.68$ & $0.70$ &   \\
\cmidrule{1-10} \cmidrule{2-10}
\multirow[c]{12}{*}{\rotatebox{90}{\text{Air 10 PM}}} & \multirow[c]{3}{*}{\rotatebox{90}{\text{Forest}}} & \text{$\Delta$ Cov} & $0.028^{\pm 0.019}$ & $0.008^{\pm 0.000}$ & $\textcolor{gray}{\text{-}0.066}$ & $\text{-}0.004$ & $\text{-}0.019$ & $\textcolor{gray}{\text{-}0.033}$ &   \\
\rotatebox{90}{} &  & \text{PI-Width} & $\textbf{93.9}^{\pm 11.1}$ & $118.5^{\pm 0.1}$ & $\textcolor{gray}{202.8}$ & $263.5$ & $243.1$ & $\textcolor{gray}{229.8}$ &   \\
\rotatebox{90}{} &  & \text{Winkler} & $\textbf{1.50}^{\pm 0.09}$ & $2.23^{\pm 0.00}$ & $\textcolor{gray}{4.16}$ & $4.03$ & $4.94$ & $\textcolor{gray}{4.98}$ &   \\
\cmidrule{2-10}
\rotatebox{90}{} & \multirow[c]{3}{*}{\rotatebox{90}{\text{LGBM}}} & \text{$\Delta$ Cov} & $0.017^{\pm 0.016}$ & $0.023^{\pm 0.000}$ & $\textcolor{gray}{\text{-}0.057}$ & $\text{-}0.004$ & $\text{-}0.017$ & $\textcolor{gray}{\text{-}0.028}$ & $\text{-}0.001$ \\
\rotatebox{90}{} &  & \text{PI-Width} & $\textbf{85.6}^{\pm 7.4}$ & $113.2^{\pm 0.1}$ & $\textcolor{gray}{178.3}$ & $224.8$ & $206.7$ & $\textcolor{gray}{196.5}$ & $186.4$ \\
\rotatebox{90}{} &  & \text{Winkler} & $\textbf{1.45}^{\pm 0.06}$ & $1.94^{\pm 0.00}$ & $\textcolor{gray}{3.69}$ & $3.64$ & $4.33$ & $\textcolor{gray}{4.36}$ & $3.00$ \\
\cmidrule{2-10}
\rotatebox{90}{} & \multirow[c]{3}{*}{\rotatebox{90}{\text{Ridge}}} & \text{$\Delta$ Cov} & $0.010^{\pm 0.007}$ & $0.024^{\pm 0.000}$ & $\textcolor{gray}{\text{-}0.045}$ & $\text{-}0.002$ & $0.010$ & $0.012$ &   \\
\rotatebox{90}{} &  & \text{PI-Width} & $\textbf{79.9}^{\pm 4.7}$ & $93.9^{\pm 0.1}$ & $\textcolor{gray}{120.0}$ & $153.3$ & $153.7$ & $155.3$ &   \\
\rotatebox{90}{} &  & \text{Winkler} & $\textbf{1.35}^{\pm 0.06}$ & $1.52^{\pm 0.00}$ & $\textcolor{gray}{2.60}$ & $2.68$ & $2.95$ & $2.96$ &   \\
\cmidrule{2-10}
\rotatebox{90}{} & \multirow[c]{3}{*}{\text{\rotatebox{90}{\text{LSTM}}}} & \text{$\Delta$ Cov} & $\text{-}0.002^{\pm 0.005}$ & $0.010^{\pm 0.000}$ & $\textcolor{gray}{\text{-}0.025}$ & $\text{-}0.002$ & $0.001$ & $0.004$ &   \\
\rotatebox{90}{} &  & \text{PI-Width} & $62.7^{\pm 1.5}$ & $62.2^{\pm 0.0}$ & $\textcolor{gray}{58.1}$ & $62.4$ & $\textbf{61.8}$ & $63.0$ &   \\
\rotatebox{90}{} &  & \text{Winkler} & $1.33^{\pm 0.01}$ & $\textbf{1.21}^{\pm 0.00}$ & $\textcolor{gray}{1.32}$ & $1.29$ & $1.34$ & $1.34$ &   \\
\cmidrule{1-10} \cmidrule{2-10}
\multirow[c]{12}{*}{\rotatebox{90}{\text{Sap flow}}} & \multirow[c]{3}{*}{\rotatebox{90}{\text{Forest}}} & \text{$\Delta$ Cov} & $0.009^{\pm 0.019}$ & $0.007^{\pm 0.000}$ & $\textcolor{gray}{\text{-}0.042}$ & $0.000$ & $0.014$ & $0.005$ &   \\
\rotatebox{90}{} &  & \text{PI-Width} & $\textbf{1078.7}^{\pm 73.7}$ & $1741.8^{\pm 2.4}$ & $\textcolor{gray}{3671.6}$ & $6137.1$ & $7131.1$ & $7201.5$ &   \\
\rotatebox{90}{} &  & \text{Winkler} & $\textbf{0.30}^{\pm 0.01}$ & $0.59^{\pm 0.00}$ & $\textcolor{gray}{1.24}$ & $1.56$ & $1.76$ & $1.80$ &   \\
\cmidrule{2-10}
\rotatebox{90}{} & \multirow[c]{3}{*}{\rotatebox{90}{\text{LGBM}}} & \text{$\Delta$ Cov} & $\text{-}0.016^{\pm 0.013}$ & $0.003^{\pm 0.000}$ & $\textcolor{gray}{\text{-}0.040}$ & $\text{-}0.003$ & $0.006$ & $\text{-}0.007$ & $0.010$ \\
\rotatebox{90}{} &  & \text{PI-Width} & $\textbf{875.5}^{\pm 49.8}$ & $1582.3^{\pm 1.0}$ & $\textcolor{gray}{2924.1}$ & $4805.3$ & $5588.5$ & $5614.8$ & $6273.5$ \\
\rotatebox{90}{} &  & \text{Winkler} & $\textbf{0.27}^{\pm 0.00}$ & $0.49^{\pm 0.00}$ & $\textcolor{gray}{0.96}$ & $1.25$ & $1.43$ & $1.46$ & $1.50$ \\
\cmidrule{2-10}
\rotatebox{90}{} & \multirow[c]{3}{*}{\rotatebox{90}{\text{Ridge}}} & \text{$\Delta$ Cov} & $\text{-}0.025^{\pm 0.023}$ & $\textcolor{gray}{\text{-}0.241}^{\pm 0.000}$ & $\textcolor{gray}{\text{-}0.041}$ & $\text{-}0.015$ & $\textcolor{gray}{\text{-}0.251}$ & $\textcolor{gray}{\text{-}0.358}$ &   \\
\rotatebox{90}{} &  & \text{PI-Width} & $\textbf{1639.6}^{\pm 93.2}$ & $\textcolor{gray}{2060.5}^{\pm 1.6}$ & $\textcolor{gray}{3117.5}$ & $10628.9$ & $\textcolor{gray}{8943.3}$ & $\textcolor{gray}{7148.7}$ &   \\
\rotatebox{90}{} &  & \text{Winkler} & $\textbf{0.46}^{\pm 0.05}$ & $\textcolor{gray}{2.52}^{\pm 0.00}$ & $\textcolor{gray}{0.92}$ & $2.42$ & $\textcolor{gray}{3.31}$ & $\textcolor{gray}{5.85}$ &   \\
\cmidrule{2-10}
\rotatebox{90}{} & \multirow[c]{3}{*}{\text{\rotatebox{90}{\text{LSTM}}}} & \text{$\Delta$ Cov} & $0.004^{\pm 0.004}$ & $0.004^{\pm 0.000}$ & $\text{-}0.019$ & $\text{-}0.000$ & $\text{-}0.022$ & $\textcolor{gray}{\text{-}0.042}$ &   \\
\rotatebox{90}{} &  & \text{PI-Width} & $\textbf{594.3}^{\pm 7.7}$ & $628.2^{\pm 0.9}$ & $768.0$ & $990.0$ & $903.9$ & $\textcolor{gray}{817.2}$ &   \\
\rotatebox{90}{} &  & \text{Winkler} & $\textbf{0.19}^{\pm 0.01}$ & $0.24^{\pm 0.00}$ & $0.28$ & $0.32$ & $0.35$ & $\textcolor{gray}{0.36}$ &   \\
\cmidrule{1-10} \cmidrule{2-10}

\multirow[c]{3}{*}{\text{SF}} & \multirow[c]{3}{*}{\text{\rotatebox{90}{\text{LSTM}}}} & \text{$\Delta$ Cov} & $0.001^{\pm 0.041}$ & $0.027^{\pm 0.000}$ & $\textcolor{gray}{\text{-}0.054}$ & $\text{-}0.000$ & $0.005$ & $0.009$ &   \\
 &  & \text{PI-Width} & $\textbf{1.39}^{\pm 0.17}$ & $1.58^{\pm 0.00}$ & $\textcolor{gray}{1.55}$ & $1.94$ & $1.99$ & $2.08$ &   \\
 &  & \text{Winkler} & $\textbf{0.79}^{\pm 0.03}$ & $0.91^{\pm 0.00}$ & $\textcolor{gray}{1.27}$ & $1.21$ & $1.28$ & $1.29$ &   \\
\bottomrule
\end{tabular}
\end{table*}

\paragraph{Compared approaches.}
We compare \ourmethod to different state-of-the-art CP approaches for time series data: \enbpi \citep{xu2021conformal}, \spic \citep{xu2022sequential}, \nexcp \citep{barber2022conformal}, CopulaCPTS \citep{Sun2022CopulaCP}, and AdaptiveCI\footnote{\adaptiveci works on top of an existing quantile prediction method. Hence, we exclusively make comparisons based on the LightGBM models that can predict quantiles instead of point estimates. The approach is orthogonal to the remaining compared models and could be combined with \ourmethod.} \citep{gibbs2021adaptive}. 
In addition, the results of standard split CP (CP) serve as a baseline, which, for the LSTM base predictor, corresponds to CF-RNN \citep{stankeviciute2021conformal} in our setting (one-step, univariate target variable).
Appendix~\ref{sec:hyperparam} describes our hyperparameter search for each method. 
For \spic, an adaption of the original algorithm was necessary to provide scalability to larger datasets.
Appendix~\ref{sec:spic-retrained} contains more details and an empirical justification.
Appendix~\ref{sec:adaptive-cp} evaluates the addition of \adaptiveci \citep{gibbs2021adaptive} as an enhancement to \ourmethod and the other time series CP methods.
Appendix~\ref{sec:appendix-full-tables} gives a comparison to non-CP methods.
Lastly, Appendix~\ref{sec:knn-comparison} presents a supplemental comparison to \knn that shows the superiority of the learned similarity representation in \ourmethod.

\newpage

\paragraph{Metrics.} In our analyses, we compute $\Delta \ \text{Cov}$, PI-Width, and the Winkler score \citep{winkler1972decision} per time series and miscoverage level.
The Winkler score jointly elicitates miscoverage and interval width in a single metric:
\begin{align}\label{eq:winkler-score}
    \text{WS}_{\alpha}(\BZ_{t+1}, y_{t+1}) =
    \begin{cases}
        \text{IW}_t^{\alpha}(\BZ_{t+1}) + \frac{2}{\alpha} (y_{t+1} - \Ch_t^{\alpha,u}(\BZ_{t+1})) & \text{if} \ y_{t+1} > \Ch_t^{\alpha,u}(\BZ_{t+1}),\\
        \text{IW}_t^{\alpha}(\BZ_{t+1}) + \frac{2}{\alpha} (\Ch_t^{\alpha,l}(\BZ_{t+1}) - y_{t+1}) & \text{if} \ y_{t+1} <  \Ch_t^{\alpha,l}(\BZ_{t+1}),\\
        \text{IW}_t^{\alpha}(\BZ_{t+1}) &  \text{else}.
    \end{cases}
\end{align}

The score is calculated per time step $t$ and miscoverage level $\alpha$.
It corresponds to the interval width $\text{IW}_t^{\alpha}=\Ch_t^{\alpha,u} - \Ch_t^{\alpha,l}$ whenever the observed value $y_t$ is between the upper bound $\Ch_t^{\alpha,u}(\BZ_{t+1})$ and the lower bound $\Ch_t^{\alpha,l}(\BZ_{t+1})$ of $\Ch_t^{\alpha}(\BZ_{t+1})$.
If $y_{t+1}$ is outside these bounds, a penalty is added to the interval width.
We evaluate the mean Winkler score over all time steps.

We repeated each experiment with 12 different seeds.
For brevity, we only show the mean performance of one dataset per domain for $\alpha = 0.1$ in the main paper \citep[which is the most commonly reported value in the CP literature; e.g.,][]{xu2022sequential, barber2022conformal, gibbs2021adaptive}.
Appendix~\ref{sec:appendix-full-tables} presents additional results for all datasets and more $\alpha$ levels.

\subsection{Results \& Discussion}\label{sec:results-discuss}

%
%
\ourmethod has the most efficient prediction intervals for each domain --- with only one exception (Table~\ref{tab:main-results-alpha-0.1}; significance tested with a Mann--Whitney $U$ test at $p<0.005$) for the evaluated miscoverage level ($\alpha = 0.1$). 
In multiple experiments (\solarThreeY, \solarOneY), the \ourmethod prediction intervals are less than half as wide as those of the approach with the second-smallest PI-Width.
The second-most efficient intervals are predicted most often by \spic.
This ranking also holds for the Winkler score, where \ourmethod achieves the best (i.e., lowest) Winkler scores and \spic ranks second in most experiments.
Notably, these results reflect the increasing requirements posed by each uncertainty estimation method on the dataset (Section~\ref{sec:related-work}).

Furthermore, \ourmethod outperforms the other methods regarding both Winkler score and PI-Width when we evaluate over additional datasets and at different miscoverage levels (see Appendix~\ref{sec:appendix-full-tables}).
\ourmethod is the best-performing approach in the  vast majority of cases.
The variation in size of the different solar datasets (results for \solarOneY and \solarSmall in Appendix~\ref{sec:appendix-full-tables}) shows that \ourmethod especially increases its lead when more data is available.
We hypothesize that this is because the MHN can learn more generalizable retrieval patterns with more data.
However, data scarcity is not a limitation of similarity-based CP in general.
In fact, a simplified and almost parameter-free variation of \ourmethod, which replaces the MHN with \knn retrieval, performs best on the smallest solar dataset, as we demonstrate in Appendix~\ref{sec:knn-comparison}.

Most approaches have a coverage gap close to zero in almost all experiments --- they approximately achieve the specified marginal coverage level.
This is also reflected by the fact that the ranking of Winkler scores and PI-Widths agree in most evaluations.
Appendix~\ref{sec:local-coverage} provides a supplementary analysis of the local coverage. 
The results for non-CP methods show a different picture (see Appendix ~\ref{sec:appendix-full-tables} and Table~\ref{tab:non-cp-comp-0}).
Here, the non-CP models most often exhibit very narrow prediction intervals at the cost of high miscoverage.

%
%
Interestingly, standard CP also achieves good coverage for all experiments at the cost of inefficient prediction intervals.
There is one notable exception:
for the \sapflux dataset with ridge regression, we find $\Delta \ \text{Cov} = - 0.358$.
In this case, we argue that the bad performance of standard CP is driven by a strong violation of the (required) exchangeability assumption.
Specifically, a trend in the errors leads to a distribution shift over time (as illustrated in Appendix~\ref{sec:appendix-full-tables}, Figure~\ref{fig:eps-drift}).
\ourmethod and \nexcp handle this shift without substantial loss in coverage. 
The inferior coverage of \spic is likely influenced by the modification to larger datasets (see Appendix~\ref{sec:spic-retrained}).

Performance gaps between the approaches differ considerably across datasets, but are generally consistent across prediction models.
The biggest differences between the best and worst methods exist in \solarThreeY and \sapflux, likely due to the distinctive regimes in these datasets.
The smallest (but still significant) differences are visible in the \hydro data.  
On this dataset, we also evaluated the state-of-the-art non-CP uncertainty model in the domain \citep[][see Appendix~\ref{sec:appendix-full-tables}]{klotz2022uncertainty} and found that \ourmethod outperforms it with respect to efficiency.   

In terms of computational resources, \spic is most demanding followed by \ourmethod, as both methods require a learning step in contrast to the other methods.
A detailed overview can be found in Appendix~\ref{sec:comutatation-resources}.

\begin{figure*}[t]
\includegraphics[width=13cm]{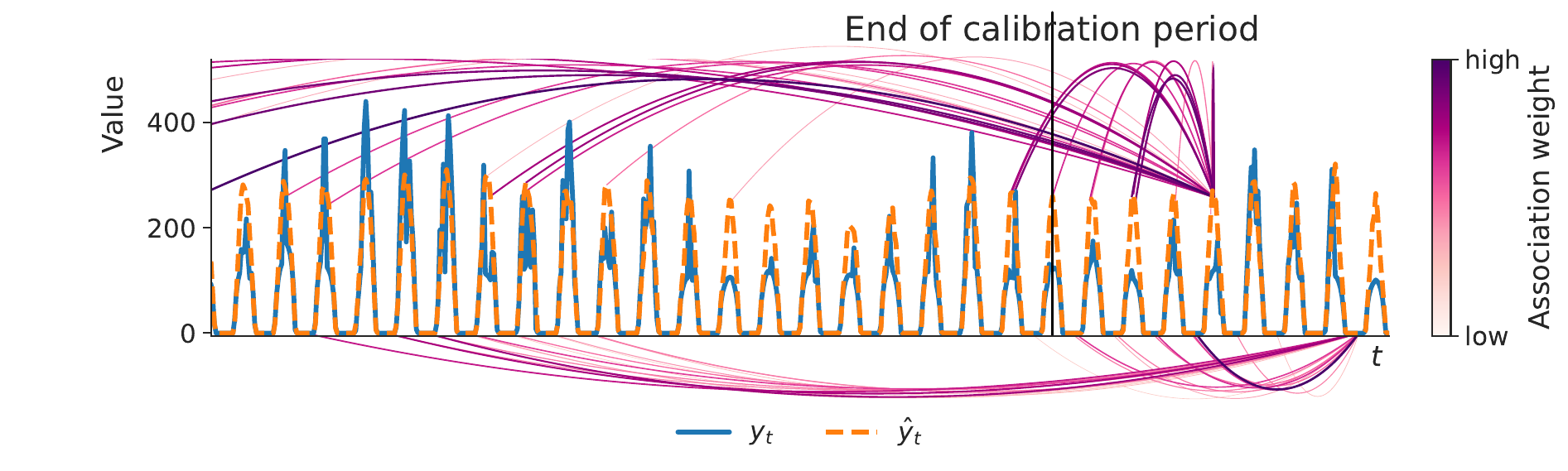}
\centering
\caption{Exemplary visualization of the 30 highest association weights that the MHN places on previous time steps (some are before the visualized part of the memory). \ourmethod retrieves similar peak values when estimating at a peak, and it retrieves similar small values when estimating at a small value.}
\label{fig:association-viz}
\end{figure*}

%
%
To assess whether \ourmethod learns meaningful associations within regimes, we conducted a qualitative study on the \solarThreeY dataset.
Figure~\ref{fig:association-viz} shows that \ourmethod retrieves the most highly weighted errors from time steps with similar regimes.
The illustrated weighting at the time step with a low prediction value retrieves previous time steps which are also in low-valued regimes.
Similarly, Figure~\ref{fig:association-eps-viz} (Appendix~\ref{sec:appendix-full-tables}) suggests that the learned distinction corresponds to the error regimes, which is crucial for \ourmethod.

\section{Conclusions}\label{sec:conclusion}

We have introduced \ourmethod, a novel CP approach for time series tasks.
\ourmethod uses continuous Modern Hopfield Networks to construct prediction intervals based on previously seen events with similar error distribution regimes.
We exploit that similar features lead to similar errors.
Associating features with errors identifies
 regimes with similar error distributions.
\ourmethod learns this association in the Modern Hopfield Network, 
which dynamically adjusts its focus on stored previous features 
according to the current regime.

Our experiments with established and novel datasets 
show that \ourmethod achieves state of the art:
It generates more efficient prediction intervals than existing CP methods and approximately preserves coverage, even in non-exchangeable scenarios
like time series.
\ourmethod comes with formal guarantees within the CP framework for uncertainty estimation in real-world applications such as streamflow prediction.
Furthermore, \ourmethod scales well to large datasets, provides multiple coverage levels after calibration, and shares information across individual time series within a dataset.

Future work comprises:
(a)~Drawing from multiple time series during the inference phase.
\ourmethod is already trained on the whole dataset at once, but it might be advantageous to leverage more information during inference as well. 
(b)~Investigating direct training objectives from which learning-based CP might benefit.
(c)~Using \ourmethod beyond time series, as non-exchangeability is also an issue 
in other domains which limits the applicability of existing CP methods.

\section*{Acknowledgements}
\small
We would like to thank Angela Bitto-Nemling for discussions during the development of the method.
The ELLIS Unit Linz, the LIT AI Lab, the Institute for Machine Learning, are supported by the Federal State Upper Austria. We thank the projects AI-MOTION (LIT-2018-6-YOU-212), DeepFlood (LIT-2019-8-YOU-213), Medical Cognitive Computing Center (MC3), INCONTROL-RL (FFG-881064), PRIMAL (FFG-873979), S3AI (FFG-872172), DL for GranularFlow (FFG-871302), EPILEPSIA (FFG-892171), AIRI FG 9-N (FWF-36284, FWF-36235), AI4GreenHeatingGrids(FFG- 899943), INTEGRATE (FFG-892418), ELISE (H2020-ICT-2019-3 ID: 951847), Stars4Waters (HORIZON-CL6-2021-CLIMATE-01-01). We thank Audi.JKU Deep Learning Center, TGW LOGISTICS GROUP GMBH, Silicon Austria Labs (SAL), FILL Gesellschaft mbH, Anyline GmbH, Google, ZF Friedrichshafen AG, Robert Bosch GmbH, UCB Biopharma SRL, Merck Healthcare KGaA, Verbund AG, GLS (Univ. Waterloo) Software Competence Center Hagenberg GmbH, T\"{U}V Austria, Frauscher Sensonic, TRUMPF and the NVIDIA Corporation.  The computational results presented have been achieved in part using the Vienna Scientific Cluster (VSC) and the In-Memory Supercomputer MACH-2 (operated by the  Scientific Computing Administration of JKU Linz).
\normalsize

\bibliography{literature, literature-cp}

\begin{thebibliography}{77}
\providecommand{\natexlab}[1]{#1}
\providecommand{\url}[1]{\texttt{#1}}
\expandafter\ifx\csname urlstyle\endcsname\relax
  \providecommand{\doi}[1]{doi: #1}\else
  \providecommand{\doi}{doi: \begingroup \urlstyle{rm}\Url}\fi

\bibitem[Abbott \& Arian(1987)Abbott and Arian]{Abbott:87}
Abbott, L.~F. and Arian, Y.
\newblock Storage capacity of generalized networks.
\newblock \emph{Physical Review A}, 36:\penalty0 5091--5094, 1987.
\newblock \doi{10.1103/PhysRevA.36.5091}.

\bibitem[Addor et~al.(2017)Addor, Newman, Mizukami, and Clark]{addor2017camels}
Addor, N., Newman, A.~J., Mizukami, N., and Clark, M.~P.
\newblock The {CAMELS} data set: catchment attributes and meteorology for
  large-sample studies.
\newblock \emph{Hydrology and Earth System Sciences (HESS)}, 21\penalty0
  (10):\penalty0 5293--5313, 2017.

\bibitem[Angelopoulos et~al.(2020)Angelopoulos, Bates, Malik, and
  Jordan]{angelopoulos2020uncertainty}
Angelopoulos, A., Bates, S., Malik, J., and Jordan, M.~I.
\newblock Uncertainty sets for image classifiers using conformal prediction.
\newblock \emph{arXiv preprint arXiv:2009.14193}, 2020.

\bibitem[Angelopoulos \& Bates(2021)Angelopoulos and
  Bates]{angelopoulos2021gentle}
Angelopoulos, A.~N. and Bates, S.
\newblock A gentle introduction to conformal prediction and distribution-free
  uncertainty quantification.
\newblock \emph{arXiv preprint arXiv:2107.07511}, 2021.

\bibitem[Baldi \& Venkatesh(1987)Baldi and Venkatesh]{Baldi:87}
Baldi, P. and Venkatesh, S.~S.
\newblock Number of stable points for spin-glasses and neural networks of
  higher orders.
\newblock \emph{Physical Review Letters}, 58:\penalty0 913--916, 1987.
\newblock \doi{10.1103/PhysRevLett.58.913}.

\bibitem[Bishop(1994)]{bishop1994mixture}
Bishop, C.~M.
\newblock Mixture density networks.
\newblock Technical report, Neural Computing Research Group, 1994.

\bibitem[Caputo \& Niemann(2002)Caputo and Niemann]{Caputo:02}
Caputo, B. and Niemann, H.
\newblock Storage capacity of kernel associative memories.
\newblock In \emph{Proceedings of the International Conference on Artificial
  Neural Networks (ICANN)}, pp.\  51–56, Berlin, Heidelberg, 2002.
  Springer-Verlag.

\bibitem[Chen et~al.(1986)Chen, Lee, Sun, Lee, Maxwell, and Giles]{Chen:86}
Chen, H.~H., Lee, Y.~C., Sun, G.~Z., Lee, H.~Y., Maxwell, T., and Giles, C.~L.
\newblock High order correlation model for associative memory.
\newblock \emph{AIP Conference Proceedings}, 151\penalty0 (1):\penalty0 86--99,
  1986.
\newblock \doi{10.1063/1.36224}.

\bibitem[Corani et~al.(2021)Corani, Benavoli, and Zaffalon]{corani2021time}
Corani, G., Benavoli, A., and Zaffalon, M.
\newblock Time series forecasting with gaussian processes needs priors.
\newblock In \emph{Joint European Conference on Machine Learning and Knowledge
  Discovery in Databases}, pp.\  103--117. Springer, 2021.

\bibitem[Dong et~al.(2022)Dong, Chen, and Wang]{dong2022retrosynthesis}
Dong, Z., Chen, Z., and Wang, Q.
\newblock Retrosynthesis prediction based on graph relation network.
\newblock In \emph{2022 15th International Congress on Image and Signal
  Processing, BioMedical Engineering and Informatics (CISP-BMEI)}, pp.\  1--5.
  IEEE, 2022.

\bibitem[Fontana et~al.(2023)Fontana, Zeni, and Vantini]{fontana2023conformal}
Fontana, M., Zeni, G., and Vantini, S.
\newblock Conformal prediction: a unified review of theory and new challenges.
\newblock \emph{Bernoulli}, 29\penalty0 (1):\penalty0 1--23, 2023.

\bibitem[Fox \& Rubin(1964)Fox and Rubin]{fox1964admissibility}
Fox, M. and Rubin, H.
\newblock Admissibility of quantile estimates of a single location parameter.
\newblock \emph{The Annals of Mathematical Statistics}, pp.\  1019--1030, 1964.

\bibitem[Foygel~Barber et~al.(2021)Foygel~Barber, Candes, Ramdas, and
  Tibshirani]{foygel2021limits}
Foygel~Barber, R., Candes, E.~J., Ramdas, A., and Tibshirani, R.~J.
\newblock The limits of distribution-free conditional predictive inference.
\newblock \emph{Information and Inference: A Journal of the IMA}, 10\penalty0
  (2):\penalty0 455--482, 2021.

\bibitem[Foygel~Barber et~al.(2022)Foygel~Barber, Candes, Ramdas, and
  Tibshirani]{barber2022conformal}
Foygel~Barber, R., Candes, E.~J., Ramdas, A., and Tibshirani, R.~J.
\newblock Conformal prediction beyond exchangeability.
\newblock \emph{arXiv preprint arXiv:2202.13415}, 2022.

\bibitem[F{\"u}rst et~al.(2022)F{\"u}rst, Rumetshofer, Lehner, Tran, Tang,
  Ramsauer, Kreil, Kopp, Klambauer, Bitto-Nemling, and
  Hochreiter]{furst2021cloob}
F{\"u}rst, A., Rumetshofer, E., Lehner, J., Tran, V.~T., Tang, F., Ramsauer,
  H., Kreil, D.~P., Kopp, M.~K., Klambauer, G., Bitto-Nemling, A., and
  Hochreiter, S.
\newblock {CLOOB:} {Modern} {Hopfield} {Networks} with info{LOOB} outperform
  {CLIP}.
\newblock In Oh, A.~H., Agarwal, A., Belgrave, D., and Cho, K. (eds.),
  \emph{Advances in Neural Information Processing Systems}, 2022.

\bibitem[Gal \& Ghahramani(2016)Gal and Ghahramani]{gal2016dropout}
Gal, Y. and Ghahramani, Z.
\newblock Dropout as a bayesian approximation: Representing model uncertainty
  in deep learning.
\newblock In \emph{international conference on machine learning}, pp.\
  1050--1059. PMLR, 2016.

\bibitem[Gardner(1987)]{Gardner:87}
Gardner, E.
\newblock Multiconnected neural network models.
\newblock \emph{Journal of Physics A}, 20\penalty0 (11):\penalty0 3453--3464,
  1987.
\newblock \doi{10.1088/0305-4470/20/11/046}.

\bibitem[Gibbs \& Candes(2021)Gibbs and Candes]{gibbs2021adaptive}
Gibbs, I. and Candes, E.~J.
\newblock Adaptive conformal inference under distribution shift.
\newblock \emph{Advances in Neural Information Processing Systems},
  34:\penalty0 1660--1672, 2021.

\bibitem[Gneiting \& Katzfuss(2014)Gneiting and
  Katzfuss]{gneiting2014probforecasting}
Gneiting, T. and Katzfuss, M.
\newblock Probabilistic forecasting.
\newblock \emph{Annual Review of Statistics and Its Application}, 1\penalty0
  (1):\penalty0 125--151, 2014.
\newblock \doi{10.1146/annurev-statistics-062713-085831}.

\bibitem[Hamilton(1990)]{hamilton1990analysis}
Hamilton, J.~D.
\newblock Analysis of time series subject to changes in regime.
\newblock \emph{Journal of econometrics}, 45\penalty0 (1-2):\penalty0 39--70,
  1990.

\bibitem[Herzen et~al.(2022)Herzen, L{\"a}ssig, Piazzetta, Neuer, Tafti,
  Raille, Van~Pottelbergh, Pasieka, Skrodzki, Huguenin,
  et~al.]{JMLR:v23:21-1177}
Herzen, J., L{\"a}ssig, F., Piazzetta, S.~G., Neuer, T., Tafti, L., Raille, G.,
  Van~Pottelbergh, T., Pasieka, M., Skrodzki, A., Huguenin, N., et~al.
\newblock Darts: User-friendly modern machine learning for time series.
\newblock \emph{Journal of Machine Learning Research}, 23\penalty0
  (124):\penalty0 1--6, 2022.

\bibitem[Hochreiter \& Schmidhuber(1997)Hochreiter and
  Schmidhuber]{Hochreiter:97}
Hochreiter, S. and Schmidhuber, J.
\newblock Long short-term memory.
\newblock \emph{Neural Comput.}, 9\penalty0 (8):\penalty0 1735--1780, 1997.

\bibitem[Hopfield(1982)]{Hopfield:82}
Hopfield, J.~J.
\newblock Neural networks and physical systems with emergent collective
  computational abilities.
\newblock \emph{Proceedings of the National Academy of Sciences}, 79\penalty0
  (8):\penalty0 2554--2558, 1982.

\bibitem[Hopfield(1984)]{Hopfield:84}
Hopfield, J.~J.
\newblock Neurons with graded response have collective computational properties
  like those of two-state neurons.
\newblock \emph{Proceedings of the National Academy of Sciences}, 81\penalty0
  (10):\penalty0 3088--3092, 1984.
\newblock \doi{10.1073/pnas.81.10.3088}.

\bibitem[Horn \& Usher(1988)Horn and Usher]{Horn:88}
Horn, D. and Usher, M.
\newblock Capacities of multiconnected memory models.
\newblock \emph{Journal of Physics France}, 49\penalty0 (3):\penalty0 389--395,
  1988.
\newblock \doi{10.1051/jphys:01988004903038900}.

\bibitem[Jensen et~al.(2022)Jensen, Bianchi, and Anfinsen]{jensen2022ensemble}
Jensen, V., Bianchi, F.~M., and Anfinsen, S.~N.
\newblock Ensemble conformalized quantile regression for probabilistic time
  series forecasting.
\newblock \emph{arXiv preprint arXiv:2202.08756}, 2022.

\bibitem[Klotz et~al.(2022)Klotz, Kratzert, Gauch, Keefe~Sampson, Brandstetter,
  Klambauer, Hochreiter, and Nearing]{klotz2022uncertainty}
Klotz, D., Kratzert, F., Gauch, M., Keefe~Sampson, A., Brandstetter, J.,
  Klambauer, G., Hochreiter, S., and Nearing, G.
\newblock Uncertainty estimation with deep learning for rainfall--runoff
  modeling.
\newblock \emph{Hydrology and Earth System Sciences}, 26\penalty0 (6):\penalty0
  1673--1693, 2022.
\newblock \doi{10.5194/hess-26-1673-2022}.

\bibitem[Kratzert et~al.(2019)Kratzert, Klotz, Shalev, Klambauer, Hochreiter,
  and Nearing]{kratzert2019regional}
Kratzert, F., Klotz, D., Shalev, G., Klambauer, G., Hochreiter, S., and
  Nearing, G.
\newblock Towards learning universal, regional, and local hydrological
  behaviors via machine learning applied to large-sample datasets.
\newblock \emph{Hydrology and Earth System Sciences}, 23\penalty0
  (12):\penalty0 5089--5110, 2019.
\newblock \doi{10.5194/hess-23-5089-2019}.

\bibitem[Kratzert et~al.(2021)Kratzert, Klotz, Hochreiter, and
  Nearing]{kratzert2020note}
Kratzert, F., Klotz, D., Hochreiter, S., and Nearing, G.~S.
\newblock A note on leveraging synergy in multiple meteorological data sets
  with deep learning for rainfall--runoff modeling.
\newblock \emph{Hydrology and Earth System Sciences}, 25\penalty0 (5):\penalty0
  2685--2703, 2021.

\bibitem[Kratzert et~al.(2022)Kratzert, Gauch, Nearing, and
  Klotz]{kratzert2022nh}
Kratzert, F., Gauch, M., Nearing, G., and Klotz, D.
\newblock {NeuralHydrology} --- a {Python} library for deep learning research
  in hydrology.
\newblock \emph{Journal of Open Source Software}, 7\penalty0 (71):\penalty0
  4050, 2022.
\newblock \doi{10.21105/joss.04050}.

\bibitem[Krotov \& Hopfield(2016)Krotov and Hopfield]{Krotov:16}
Krotov, D. and Hopfield, J.~J.
\newblock Dense associative memory for pattern recognition.
\newblock In Lee, D.~D., Sugiyama, M., Luxburg, U.~V., Guyon, I., and Garnett,
  R. (eds.), \emph{Advances in Neural Information Processing Systems}, pp.\
  1172--1180. Curran Associates, Inc., 2016.

\bibitem[Krzysztofowicz(2001)]{krzysztofowicz2001case}
Krzysztofowicz, R.
\newblock The case for probabilistic forecasting in hydrology.
\newblock \emph{Journal of hydrology}, 249\penalty0 (1-4):\penalty0 2--9, 2001.

\bibitem[Lei \& Wasserman(2014)Lei and Wasserman]{Lei2014DistributionfreePB}
Lei, J. and Wasserman, L.~A.
\newblock Distribution‐free prediction bands for non‐parametric regression.
\newblock \emph{Journal of the Royal Statistical Society: Series B (Statistical
  Methodology)}, 76, 2014.

\bibitem[Loshchilov \& Hutter(2019)Loshchilov and Hutter]{loshchilov2019adamw}
Loshchilov, I. and Hutter, F.
\newblock Decoupled weight decay regularization.
\newblock In \emph{7th International Conference on Learning Representations,
  {ICLR} 2019, New Orleans, LA, USA, May 6-9, 2019}. OpenReview.net, 2019.

\bibitem[Masserano et~al.(2022)Masserano, Rangapuram, Kapoor, Nirwan, Park, and
  Bohlke-Schneider]{masserano2022adaptive}
Masserano, L., Rangapuram, S.~S., Kapoor, S., Nirwan, R.~S., Park, Y., and
  Bohlke-Schneider, M.
\newblock Adaptive sampling for probabilistic forecasting under distribution
  shift.
\newblock In \emph{NeurIPS 2022 Workshop on Distribution Shifts: Connecting
  Methods and Applications}, 2022.

\bibitem[Maurer et~al.(2002)Maurer, Wood, Adam, Lettenmaier, and
  Nijssen]{maurer2002long}
Maurer, E.~P., Wood, A.~W., Adam, J.~C., Lettenmaier, D.~P., and Nijssen, B.
\newblock A long-term hydrologically based dataset of land surface fluxes and
  states for the conterminous {United States}.
\newblock \emph{Journal of climate}, 15\penalty0 (22):\penalty0 3237--3251,
  2002.

\bibitem[Newman et~al.(2015)Newman, Clark, Sampson, Wood, Hay, Bock, Viger,
  Blodgett, Brekke, Arnold, Hopson, and Duan]{newman2015development}
Newman, A.~J., Clark, M.~P., Sampson, K., Wood, A., Hay, L.~E., Bock, A.,
  Viger, R.~J., Blodgett, D., Brekke, L., Arnold, J.~R., Hopson, T., and Duan,
  Q.
\newblock Development of a large-sample watershed-scale hydrometeorological
  data set for the contiguous {USA}: data set characteristics and assessment of
  regional variability in hydrologic model performance.
\newblock \emph{Hydrology and Earth System Sciences}, 19\penalty0 (1):\penalty0
  209--223, 2015.
\newblock \doi{10.5194/hess-19-209-2015}.

\bibitem[Newman et~al.(2017)Newman, Mizukami, Clark, Wood, Nijssen, and
  Nearing]{newman2017benchmarking}
Newman, A.~J., Mizukami, N., Clark, M.~P., Wood, A.~W., Nijssen, B., and
  Nearing, G.
\newblock Benchmarking of a physically based hydrologic model.
\newblock \emph{Journal of Hydrometeorology}, 18\penalty0 (8):\penalty0
  2215--2225, 2017.
\newblock \doi{10.1175/JHM-D-16-0284.1}.

\bibitem[Oreshkin et~al.(2020)Oreshkin, Carpov, Chapados, and
  Bengio]{Oreshkin2020N-BEATS:}
Oreshkin, B.~N., Carpov, D., Chapados, N., and Bengio, Y.
\newblock N-beats: Neural basis expansion analysis for interpretable time
  series forecasting.
\newblock In \emph{International Conference on Learning Representations}, 2020.
\newblock URL \url{https://openreview.net/forum?id=r1ecqn4YwB}.

\bibitem[Paischer et~al.(2022)Paischer, Adler, Patil, Bitto-Nemling,
  Holzleitner, Lehner, Eghbal-zadeh, and Hochreiter]{paischer2022history}
Paischer, F., Adler, T., Patil, V., Bitto-Nemling, A., Holzleitner, M., Lehner,
  S., Eghbal-zadeh, H., and Hochreiter, S.
\newblock History compression via language models in reinforcement learning.
\newblock In Chaudhuri, K., Jegelka, S., Song, L., Szepesvari, C., Niu, G., and
  Sabato, S. (eds.), \emph{Proceedings of the 39th International Conference on
  Machine Learning}, volume 162 of \emph{Proceedings of Machine Learning
  Research}, pp.\  17156--17185. PMLR, 17--23 Jul 2022.

\bibitem[Papadopoulos \& Haralambous(2011)Papadopoulos and
  Haralambous]{papadopoulos2011reliable}
Papadopoulos, H. and Haralambous, H.
\newblock Reliable prediction intervals with regression neural networks.
\newblock \emph{Neural Networks}, 24\penalty0 (8):\penalty0 842--851, 2011.

\bibitem[Paszke et~al.(2019)Paszke, Gross, Massa, Lerer, Bradbury, Chanan,
  Killeen, Lin, Gimelshein, Antiga, et~al.]{paszke2019pytorch}
Paszke, A., Gross, S., Massa, F., Lerer, A., Bradbury, J., Chanan, G., Killeen,
  T., Lin, Z., Gimelshein, N., Antiga, L., et~al.
\newblock Pytorch: An imperative style, high-performance deep learning library.
\newblock \emph{Advances in neural information processing systems}, 32, 2019.

\bibitem[Pedregosa et~al.(2011)Pedregosa, Varoquaux, Gramfort, Michel, Thirion,
  Grisel, Blondel, Prettenhofer, Weiss, Dubourg, Vanderplas, Passos,
  Cournapeau, Brucher, Perrot, and Duchesnay]{scikit-learn}
Pedregosa, F., Varoquaux, G., Gramfort, A., Michel, V., Thirion, B., Grisel,
  O., Blondel, M., Prettenhofer, P., Weiss, R., Dubourg, V., Vanderplas, J.,
  Passos, A., Cournapeau, D., Brucher, M., Perrot, M., and Duchesnay, E.
\newblock Scikit-learn: Machine learning in {P}ython.
\newblock \emph{Journal of Machine Learning Research}, 12:\penalty0 2825--2830,
  2011.

\bibitem[Pinsker(1964)]{pinsker1964information}
Pinsker, M.~S.
\newblock \emph{Information and information stability of random variables and
  processes}.
\newblock Holden-Day, 1964.

\bibitem[Poyatos et~al.(2021)Poyatos, Granda, Flo, Adams, Adorj{\'a}n,
  Aguad{\'e}, Aidar, Allen, Alvarado-Barrientos, Anderson-Teixeira,
  et~al.]{poyatos2021global}
Poyatos, R., Granda, V., Flo, V., Adams, M.~A., Adorj{\'a}n, B., Aguad{\'e},
  D., Aidar, M.~P., Allen, S., Alvarado-Barrientos, M.~S., Anderson-Teixeira,
  K.~J., et~al.
\newblock Global transpiration data from sap flow measurements: the sapfluxnet
  database.
\newblock \emph{Earth system science data}, 13\penalty0 (6):\penalty0
  2607--2649, 2021.

\bibitem[Psaltis \& Cheol(1986)Psaltis and Cheol]{Psaltis:86}
Psaltis, D. and Cheol, H.~P.
\newblock Nonlinear discriminant functions and associative memories.
\newblock \emph{AIP Conference Proceedings}, 151\penalty0 (1):\penalty0
  370--375, 1986.
\newblock \doi{10.1063/1.36241}.

\bibitem[Quandt(1958)]{quandt1958estimation}
Quandt, R.~E.
\newblock The estimation of the parameters of a linear regression system
  obeying two separate regimes.
\newblock \emph{Journal of the american statistical association}, 53\penalty0
  (284):\penalty0 873--880, 1958.

\bibitem[Ramsauer et~al.(2021)Ramsauer, Sch\"{a}fl, Lehner, Seidl, Widrich,
  Gruber, Holzleitner, Pavlovi{\'c}, Sandve, Greiff, Kreil, Kopp, Klambauer,
  Brandstetter, and Hochreiter]{Ramsauer:21}
Ramsauer, H., Sch\"{a}fl, B., Lehner, J., Seidl, P., Widrich, M., Gruber, L.,
  Holzleitner, M., Pavlovi{\'c}, M., Sandve, G.~K., Greiff, V., Kreil, D.,
  Kopp, M., Klambauer, G., Brandstetter, J., and Hochreiter, S.
\newblock {Hopfield} networks is all you need.
\newblock In \emph{9th International Conference on Learning Representations
  (ICLR)}, 2021.

\bibitem[Salinas et~al.(2020)Salinas, Flunkert, Gasthaus, and
  Januschowski]{salinas2020deepar}
Salinas, D., Flunkert, V., Gasthaus, J., and Januschowski, T.
\newblock Deepar: Probabilistic forecasting with autoregressive recurrent
  networks.
\newblock \emph{International Journal of Forecasting}, 36\penalty0
  (3):\penalty0 1181--1191, 2020.

\bibitem[Sanchez-Fernandez et~al.(2022)Sanchez-Fernandez, Rumetshofer,
  Hochreiter, and Klambauer]{klambauer2022cloome}
Sanchez-Fernandez, A., Rumetshofer, E., Hochreiter, S., and Klambauer, G.
\newblock {CLOOME}: contrastive learning unlocks bioimaging databases for
  queries with chemical structures.
\newblock In \emph{NeurIPS 2022 Women in Machine Learning Workshop}, 2022.

\bibitem[Sanquer et~al.(2012)Sanquer, Chatelain, El-Guedri, and
  Martin]{sanquer2012smooth}
Sanquer, M., Chatelain, F., El-Guedri, M., and Martin, N.
\newblock A smooth transition model for multiple-regime time series.
\newblock \emph{IEEE transactions on signal processing}, 61\penalty0
  (7):\penalty0 1835--1847, 2012.

\bibitem[Sch{\"a}fl et~al.(2022)Sch{\"a}fl, Gruber, Bitto-Nemling, and
  Hochreiter]{schafl2022hopular}
Sch{\"a}fl, B., Gruber, L., Bitto-Nemling, A., and Hochreiter, S.
\newblock Hopular: {Modern} {Hopfield} {Networks} for tabular data.
\newblock \emph{arXiv preprint arXiv:2206.00664}, 2022.

\bibitem[Sengupta et~al.(2018)Sengupta, Xie, Lopez, Habte, Maclaurin, and
  Shelby]{sengupta2018national}
Sengupta, M., Xie, Y., Lopez, A., Habte, A., Maclaurin, G., and Shelby, J.
\newblock The national solar radiation data base ({NSRDB}).
\newblock \emph{Renewable and sustainable energy reviews}, 89:\penalty0 51--60,
  2018.

\bibitem[Smyl(2020)]{smyl2020hybrid}
Smyl, S.
\newblock A hybrid method of exponential smoothing and recurrent neural
  networks for time series forecasting.
\newblock \emph{International Journal of Forecasting}, 36\penalty0
  (1):\penalty0 75--85, 2020.

\bibitem[Stankevičiūtė et~al.(2021)Stankevičiūtė, M~Alaa, and van~der
  Schaar]{stankeviciute2021conformal}
Stankevičiūtė, K., M~Alaa, A., and van~der Schaar, M.
\newblock Conformal time-series forecasting.
\newblock \emph{Advances in Neural Information Processing Systems},
  34:\penalty0 6216--6228, 2021.

\bibitem[Sun \& Yu(2022)Sun and Yu]{Sun2022CopulaCP}
Sun, S. and Yu, R.
\newblock Copula conformal prediction for multi-step time series forecasting.
\newblock \emph{ArXiv}, abs/2212.03281, 2022.

\bibitem[Sun et~al.(2022)Sun, Kim, and Choi]{sun2022recurrent}
Sun, X., Kim, S., and Choi, J.-I.
\newblock Recurrent neural network-induced gaussian process.
\newblock \emph{Neurocomputing}, 509:\penalty0 75--84, 2022.

\bibitem[Tagasovska \& Lopez-Paz(2019)Tagasovska and
  Lopez-Paz]{tagasovska2019single}
Tagasovska, N. and Lopez-Paz, D.
\newblock Single-model uncertainties for deep learning.
\newblock \emph{Advances in Neural Information Processing Systems}, 32, 2019.

\bibitem[Tajeuna et~al.(2021)Tajeuna, Bouguessa, and Wang]{tajeuna2021modeling}
Tajeuna, E.~G., Bouguessa, M., and Wang, S.
\newblock Modeling regime shifts in multiple time series.
\newblock \emph{arXiv preprint arXiv:2109.09692}, 2021.

\bibitem[Teng et~al.(2022)Teng, Wen, Zhang, Bengio, Gao, and
  Yuan]{teng2022predictive}
Teng, J., Wen, C., Zhang, D., Bengio, Y., Gao, Y., and Yuan, Y.
\newblock Predictive inference with feature conformal prediction.
\newblock \emph{arXiv preprint arXiv:2210.00173}, 2022.

\bibitem[Thornton et~al.(1997)Thornton, Running, White,
  et~al.]{thornton1997generating}
Thornton, P.~E., Running, S.~W., White, M.~A., et~al.
\newblock Generating surfaces of daily meteorological variables over large
  regions of complex terrain.
\newblock \emph{Journal of hydrology}, 190\penalty0 (3-4):\penalty0 214--251,
  1997.

\bibitem[Tibshirani et~al.(2019)Tibshirani, Foygel~Barber, Candes, and
  Ramdas]{tibshirani2019conformal}
Tibshirani, R.~J., Foygel~Barber, R., Candes, E.~J., and Ramdas, A.
\newblock Conformal prediction under covariate shift.
\newblock \emph{Advances in Neural Information Processing Systems}, 32, 2019.

\bibitem[Toccaceli et~al.(2017)Toccaceli, Nouretdinov, and
  Gammerman]{toccaceli2017conformal}
Toccaceli, P., Nouretdinov, I., and Gammerman, A.
\newblock Conformal prediction of biological activity of chemical compounds.
\newblock \emph{Annals of Mathematics and Artificial Intelligence}, 81\penalty0
  (1):\penalty0 105--123, 2017.

\bibitem[Vaswani et~al.(2017)Vaswani, Shazeer, Parmar, Uszkoreit, Jones, Gomez,
  Kaiser, and Polosukhin]{Vaswani:17}
Vaswani, A., Shazeer, N., Parmar, N., Uszkoreit, J., Jones, L., Gomez, A.~N.,
  Kaiser, L., and Polosukhin, I.
\newblock Attention is all you need.
\newblock In Guyon, I., Luxburg, U.~V., Bengio, S., Wallach, H., Fergus, R.,
  Vishwanathan, S., and Garnett, R. (eds.), \emph{Advances in Neural
  Information Processing Systems 30}, pp.\  5998--6008. Curran Associates,
  Inc., 2017.

\bibitem[Vovk(2012)]{vovk2012conditional}
Vovk, V.
\newblock Conditional validity of inductive conformal predictors.
\newblock In \emph{Asian conference on machine learning}, pp.\  475--490. PMLR,
  2012.

\bibitem[Vovk et~al.(1999)Vovk, Gammerman, and Saunders]{vovk1999machine}
Vovk, V., Gammerman, A., and Saunders, C.
\newblock Machine-learning applications of algorithmic randomness.
\newblock In Bratko, I. and Dzeroski, S. (eds.), \emph{Proceedings of the
  Sixteenth International Conference on Machine Learning {(ICML} 1999), Bled,
  Slovenia, June 27 - 30, 1999}, pp.\  444--453. Morgan Kaufmann, 1999.

\bibitem[Vovk et~al.(2005)Vovk, Gammerman, and Shafer]{vovk2005algorithmic}
Vovk, V., Gammerman, A., and Shafer, G.
\newblock \emph{Algorithmic learning in a random world}.
\newblock Springer Science \& Business Media, 2005.

\bibitem[Widrich et~al.(2020)Widrich, Sch\"{a}fl, Pavlovi{\'c}, Ramsauer,
  Gruber, Holzleitner, Brandstetter, Sandve, Greiff, Hochreiter, and
  Klambauer]{Widrich:20nips}
Widrich, M., Sch\"{a}fl, B., Pavlovi{\'c}, M., Ramsauer, H., Gruber, L.,
  Holzleitner, M., Brandstetter, J., Sandve, G.~K., Greiff, V., Hochreiter, S.,
  and Klambauer, G.
\newblock Modern {Hopfield} {Networks} and attention for immune repertoire
  classification.
\newblock In \emph{Advances in Neural Information Processing Systems}. Curran
  Associates, Inc., 2020.

\bibitem[Winkler(1972)]{winkler1972decision}
Winkler, R.~L.
\newblock A decision-theoretic approach to interval estimation.
\newblock \emph{Journal of the American Statistical Association}, 67\penalty0
  (337):\penalty0 187--191, 1972.

\bibitem[Xia et~al.(2012)Xia, Mitchell, Ek, Sheffield, Cosgrove, Wood, Luo,
  Alonge, Wei, Meng, et~al.]{xia2012continental}
Xia, Y., Mitchell, K., Ek, M., Sheffield, J., Cosgrove, B., Wood, E., Luo, L.,
  Alonge, C., Wei, H., Meng, J., et~al.
\newblock Continental-scale water and energy flux analysis and validation for
  the {North American Land Data Assimilation System} project phase 2
  ({NLDAS-2}): 1. intercomparison and application of model products.
\newblock \emph{Journal of Geophysical Research: Atmospheres}, 117\penalty0
  (D3), 2012.

\bibitem[Xu \& Xie(2022{\natexlab{a}})Xu and Xie]{xu2021conformal}
Xu, C. and Xie, Y.
\newblock Conformal prediction for time series.
\newblock \emph{arXiv preprint arXiv:2010.09107}, 2022{\natexlab{a}}.

\bibitem[Xu \& Xie(2022{\natexlab{b}})Xu and Xie]{xu2022sequential}
Xu, C. and Xie, Y.
\newblock Sequential predictive conformal inference for time series.
\newblock \emph{arXiv preprint arXiv:2212.03463}, 2022{\natexlab{b}}.

\bibitem[Xu et~al.(2022)Xu, Yu, Ghamisi, Kopp, and Hochreiter]{xu2022txt2img}
Xu, Y., Yu, W., Ghamisi, P., Kopp, M., and Hochreiter, S.
\newblock {Txt2Img-MHN}: Remote sensing image generation from text using
  {Modern} {Hopfield} {Networks}.
\newblock \emph{arXiv preprint arXiv:2208.04441}, 2022.

\bibitem[Zaffran et~al.(2022)Zaffran, F{\'e}ron, Goude, Josse, and
  Dieuleveut]{zaffran2022adaptive}
Zaffran, M., F{\'e}ron, O., Goude, Y., Josse, J., and Dieuleveut, A.
\newblock Adaptive conformal predictions for time series.
\newblock In \emph{International Conference on Machine Learning}, pp.\
  25834--25866. PMLR, 2022.

\bibitem[Zhang et~al.(2017)Zhang, Guo, Dong, He, Xu, and
  Chen]{zhang2017cautionary}
Zhang, S., Guo, B., Dong, A., He, J., Xu, Z., and Chen, S.~X.
\newblock Cautionary tales on air-quality improvement in {Beijing}.
\newblock \emph{Proceedings of the Royal Society A: Mathematical, Physical and
  Engineering Sciences}, 473\penalty0 (2205):\penalty0 20170457, 2017.

\bibitem[Zhu \& Laptev(2017)Zhu and Laptev]{zhu2017deep}
Zhu, L. and Laptev, N.
\newblock Deep and confident prediction for time series at {Uber}.
\newblock In \emph{2017 IEEE International Conference on Data Mining Workshops
  (ICDMW)}, pp.\  103--110. IEEE, 2017.

\bibitem[Zhu et~al.(2023)Zhu, Huang, Lee, Ibrahim, and Bindel]{zhu2023bayesian}
Zhu, X., Huang, L., Lee, E.~H., Ibrahim, C.~A., and Bindel, D.
\newblock Bayesian transformed gaussian processes.
\newblock \emph{Transactions on Machine Learning Research}, 2023.

\end{thebibliography}
\bibliographystyle{_icml2023/icml2023}

\newpage
\appendix

\section{Extended Experimental Setup \& Results}

\subsection{Hyperparameter Search}\label{sec:hyperparam}

We conducted an individual hyperparameter grid search for each predictor--dataset combination.
For methods that require calibration data (\ourmethod, \adaptiveci, \nexcp), each calibration set was split in half:
One part served as actual calibration data while the other half was used for validation.\footnote{Note that this is only the case in the hyperparameter search. 
In the evaluation, the full calibration set was used for calibration.}
As \enbpi requires only the $k$ past points, we used the full calibration  set minus these $k$ points for validation, so that it could fully exploit the available data.
Table~\ref{tab:hyperparameter-grid} shows all sets of hyperparameters used for the search.

\begin{table}[h]
\centering
\caption{Parameters used in the hyperparameter search of the uncertainty models}
\label{tab:hyperparameter-grid}
\small
\vskip 0.15in
\begin{tabular}{@{}lll@{}}
\toprule
Method & Parameter & Value \\ \midrule
\multirow{3}{*}{\ourmethod}  & Learning Rate & 0.01, 0.001  \\
                             & Dropout       & 0, 0.25, 0.5 \\ 
                             & Time Encode   & yes/no  
                             \\ \midrule
\multirow{2}{*}{\adaptiveci} & Mode          & simple, momentum  \\
                             & $\gamma$      & 0.002, 0.005, 0.01, 0.02    \\ \midrule
\enbpi                       & Window Length & \begin{tabular}[c]{@{}l@{}}200, 150, 125, 100,\\ 75, 50, 25, 10\end{tabular}          \\ \midrule
\nexcp                       & $\rho$        & \begin{tabular}[c]{@{}l@{}}0.999, 0.995, 0.993,\\ 0.99, 0.98, 0.95, 0.90\end{tabular} \\ \midrule
\knn                       & k-Top Share        & \begin{tabular}[c]{@{}l@{}}0.025, 0.05, 0.10, 0.15, \\ 0.20, 0.25, 0.30, 0.35\end{tabular} \\
\midrule
\multirow{4}{*}{MCD}  & Learning Rate & 0.005, 0.001, 0.0001  \\
                       & Dropout       & 0.1, 0.25, 0.5 \\ 
                       & Batch Size    & 512, 256  \\
                       & Hidden Size   & 64, 128, 256 \\
\midrule
\multirow{4}{*}{Gaus}  & Learning Rate & 0.005, 0.001, 0.0001  \\
                       & Dropout       & 0, 0.1, 0.25 \\ 
                       & Batch Size    & 512, 256  \\
                       & Hidden Size   & 64, 128, 256 \\
\midrule
\multirow{5}{*}{MDN}  & Learning Rate & 0.005, 0.001, 0.0001  \\
                       & Dropout       & 0, 0.1, 0.25 \\ 
                       & Batch Size    & 512, 256  \\
                       & Hidden Size   & 64, 128, 256 \\
                       & \# Components & 3, 5, 7\\
\bottomrule
\end{tabular}
\end{table}

\paragraph{Model selection.} Model selection was done uniformly for all algorithms:
As a first step, all models with a negative $\Delta \ \text{Cov}$ on the validation set were excluded.
From the remaining models, the model with the smallest PI-Width was selected.
In cases where no model achieved non-negative $\Delta \ \text{Cov}$, the model with the highest $\Delta \ \text{Cov}$ was selected.

\paragraph{\ourmethod.} 
\ourmethod was trained for 3,000 epochs in each experiment.
We chose this number to make sure that the loss and model selection metric curves are already converged.
Throughout training, we validated every 5 epochs and selected the model that best fulfilled the model selection criteria described above.
AdamW \citep{loshchilov2019adamw} with standard parameters ($\beta_1=0.9$, $\beta_2=0.999$, $\delta=0.01$) was used as optimizer.
The learning rate was part of the hyperparameter search.
Depending on the dataset size, the batch size was set to $2$ or $4$, where a batch size of $n$ would mean that the full training part of the calibration set of $n$ time series is used.

\newpage
\paragraph{\spic.}
The computational demand of \spic did not allow to conduct a full hyperparameter search for the window length parameter (see more in Section~\ref{sec:spic-retrained}).
Since \spic applies a random forest on the window, one can assume that it is capable to find the relevant parts of the window.
On top of that, a longer window has less risk than a window that is too short and (potentially) cuts off relevant information.
Hence, we set the window length to 100 for all experiments, which corresponds to the longest setting in the original paper.
To check our reasoning, we evaluated the performance of \spic with window length 25 on all but the largest two datasets (Table~\ref{tab:spic-length-comparsion-solar} and Table~\ref{tab:spic-length-comparsion-other}).
We found hardly any differences except for the smallest datasets \solarSmall.
In that case we reason that the result is due to the limited calibration data in this setting.

\paragraph{MCD/MDN/Gauss.}
In Appendix~\ref{sec:appendix-full-tables} we compare \ourmethod to non-CP architectures that can provide prediction interval estimates.
Specifically, we use Monte Carlo Dropout \citep[MCD;][]{gal2016dropout}, a Mixture Density Network \citep[MDN;][]{bishop1994mixture}, and a Gauss model (i.e., an MDN with a single Gaussian density).
An LSTM serves as backbone for all models.
Therefore, there is no distinction between the predictor model and the uncertainty model.
To allow for a fair comparison to CP approaches, the models are trained on both the predictor training data and the calibration data combined.
None of these approaches are CP methods, and MCD was specifically devised to estimate only epistemic uncertainty.
Nevertheless, we consider these methods as useful comparisons, since they are widely used for uncertainty quantification.
The models were trained for 150 epochs with AdamW \citep{loshchilov2019adamw} (using its standard parameters: $\beta_1=0.9$, $\beta_2=0.999$, $\delta=0.001$). We chose this number to ensure that the loss and model selection metrics already converged.
Throughout the training, we validated after every epoch and selected the model that best fulfilled the model selection criteria described above.
The learning rate was part of the hyperparameter search.

\paragraph{Global LSTM.}
We conducted an individual hyperparameter tuning of the global LSTM prediction model for each dataset.
Table~\ref{tab:hyperparameter-lstm} shows the hyperparameters used for the grid search.
Similar to the LSTM uncertainty models, we trained for 150 epochs with an AdamW optimizer \citep{loshchilov2019adamw} and validated after each epoch.
We optimized for the mean squared error and selected the model with the smallest validation mean squared error.

\begin{table}[h]
\centering
\caption{Parameters used in the hyperparameter search of the global LSTM prediction model.}
\small
\label{tab:hyperparameter-lstm}
\vskip 0.15in
\begin{tabular}{ll}
\toprule
Parameter & Value \\
\midrule
Learning Rate & 0.005, 0.001, 0.0001  \\
                       Dropout       & 0, 0.1, 0.25 \\ 
                       Batch Size   & 512, 256  \\
                       Hidden Size & 64, 128, 256 \\
\bottomrule
\end{tabular}
\end{table}

\setTableCaption{Performance of \spic with an input window  of length 100 and length 25 on the solar datasets.
The differences in performance are small compared to the differences across methods (Section~\ref{sec:results-discuss}). The error term represents the standard deviation over repeated runs.}


\begin{table}
\centering
\caption{\nextTableCaption}
\label{tab:spic-length-comparsion-solar}
\vskip 0.15in
\small
\begin{tabular}{lllrr}
\toprule
 &  & Window & 100 & 25 \\
Data & FC &  &  &  \\
\midrule
\multirow[c]{12}{*}{\rotatebox{90}{Solar 1Y}} & \multirow[c]{3}{*}{Forest} & $\Delta$ Cov & $0.045^{\pm 0.000}$ & $0.047^{\pm 0.000}$ \\
\rotatebox{90}{} &  & PI-Width & $97.2^{\pm 0.1}$ & $97.8^{\pm 0.3}$ \\
\rotatebox{90}{} &  & Winkler & $1.26^{\pm 0.00}$ & $1.25^{\pm 0.00}$ \\
\cmidrule{2-5}
\rotatebox{90}{} & \multirow[c]{3}{*}{LGBM} & $\Delta$ Cov & $0.045^{\pm 0.000}$ & $0.047^{\pm 0.000}$ \\
\rotatebox{90}{} &  & PI-Width & $96.6^{\pm 0.2}$ & $97.6^{\pm 0.1}$ \\
\rotatebox{90}{} &  & Winkler & $1.26^{\pm 0.00}$ & $1.26^{\pm 0.00}$ \\
\cmidrule{2-5}
\rotatebox{90}{} & \multirow[c]{3}{*}{Ridge} & $\Delta$ Cov & $0.031^{\pm 0.000}$ & $0.030^{\pm 0.000}$ \\
\rotatebox{90}{} &  & PI-Width & $112.9^{\pm 0.2}$ & $113.7^{\pm 0.1}$ \\
\rotatebox{90}{} &  & Winkler & $1.40^{\pm 0.00}$ & $1.42^{\pm 0.00}$ \\
\cmidrule{2-5}
\rotatebox{90}{} & \multirow[c]{3}{*}{LSTM} & $\Delta$ Cov & $0.018^{\pm 0.000}$ & $0.018^{\pm 0.000}$ \\
\rotatebox{90}{} &  & PI-Width & $22.5^{\pm 0.0}$ & $22.2^{\pm 0.0}$ \\
\rotatebox{90}{} &  & Winkler & $0.41^{\pm 0.00}$ & $0.40^{\pm 0.00}$ \\
\cmidrule{1-5}
\multirow[c]{9}{*}{\rotatebox{90}{Solar Small}} & \multirow[c]{3}{*}{Forest} & $\Delta$ Cov & $-0.064^{\pm 0.002}$ & $-0.024^{\pm 0.000}$ \\
\rotatebox{90}{} &  & PI-Width & $38.8^{\pm 0.4}$ & $43.2^{\pm 0.2}$ \\
\rotatebox{90}{} &  & Winkler & $1.82^{\pm 0.01}$ & $1.53^{\pm 0.00}$ \\
\cmidrule{2-5}
\rotatebox{90}{} & \multirow[c]{3}{*}{LGBM} & $\Delta$ Cov & $-0.052^{\pm 0.002}$ & $-0.027^{\pm 0.001}$ \\
\rotatebox{90}{} &  & PI-Width & $37.4^{\pm 0.2}$ & $43.3^{\pm 0.1}$ \\
\rotatebox{90}{} &  & Winkler & $1.84^{\pm 0.01}$ & $1.63^{\pm 0.00}$ \\
\cmidrule{2-5}
\rotatebox{90}{} & \multirow[c]{3}{*}{Ridge} & $\Delta$ Cov & $-0.055^{\pm 0.002}$ & $-0.057^{\pm 0.001}$ \\
\rotatebox{90}{} &  & PI-Width & $51.9^{\pm 0.2}$ & $52.8^{\pm 0.2}$ \\
\rotatebox{90}{} &  & Winkler & $1.91^{\pm 0.02}$ & $1.83^{\pm 0.00}$ \\
\bottomrule
\end{tabular}
\end{table}

\setTableCaption{Performance of \spic with an input window  of length 100 and length 25 on the different non-solar datasets.
The differences in performance are small compared to the differences across methods (Section~\ref{sec:results-discuss}). The error term represents the standard deviation over repeated runs.}


\begin{table}
\centering
\caption{\nextTableCaption}
\label{tab:spic-length-comparsion-other}
\vskip 0.15in
\small
\begin{tabular}{lllrr}
\toprule
 &  & Window & 100 & 25 \\
Data & FC &  &  &  \\
\midrule
\multirow[c]{12}{*}{\rotatebox{90}{Air 10 PM}} & \multirow[c]{3}{*}{Forest} & $\Delta$ Cov & $0.008^{\pm 0.000}$ & $0.008^{\pm 0.000}$ \\
\rotatebox{90}{} &  & PI-Width & $118.5^{\pm 0.0}$ & $118.5^{\pm 0.1}$ \\
\rotatebox{90}{} &  & Winkler & $2.23^{\pm 0.00}$ & $2.23^{\pm 0.00}$ \\
\cmidrule{2-5}
\rotatebox{90}{} & \multirow[c]{3}{*}{LGBM} & $\Delta$ Cov & $0.024^{\pm 0.000}$ & $0.023^{\pm 0.000}$ \\
\rotatebox{90}{} &  & PI-Width & $113.2^{\pm 0.2}$ & $113.2^{\pm 0.1}$ \\
\rotatebox{90}{} &  & Winkler & $1.94^{\pm 0.00}$ & $1.94^{\pm 0.00}$ \\
\cmidrule{2-5}
\rotatebox{90}{} & \multirow[c]{3}{*}{Ridge} & $\Delta$ Cov & $0.024^{\pm 0.000}$ & $0.024^{\pm 0.000}$ \\
\rotatebox{90}{} &  & PI-Width & $93.9^{\pm 0.0}$ & $93.8^{\pm 0.0}$ \\
\rotatebox{90}{} &  & Winkler & $1.52^{\pm 0.00}$ & $1.52^{\pm 0.00}$ \\
\cmidrule{2-5}
\rotatebox{90}{} & \multirow[c]{3}{*}{LSTM} & $\Delta$ Cov & $0.010^{\pm 0.000}$ & $0.010^{\pm 0.000}$ \\
\rotatebox{90}{} &  & PI-Width & $62.3^{\pm 0.1}$ & $62.3^{\pm 0.0}$ \\
\rotatebox{90}{} &  & Winkler & $1.21^{\pm 0.00}$ & $1.21^{\pm 0.00}$ \\
\cmidrule{1-5}
\multirow[c]{12}{*}{\rotatebox{90}{Air 25 PM}} & \multirow[c]{3}{*}{Forest} & $\Delta$ Cov & $-0.009^{\pm 0.000}$ & $-0.009^{\pm 0.000}$ \\
\rotatebox{90}{} &  & PI-Width & $81.5^{\pm 0.0}$ & $81.4^{\pm 0.0}$ \\
\rotatebox{90}{} &  & Winkler & $2.02^{\pm 0.00}$ & $2.02^{\pm 0.00}$ \\
\cmidrule{2-5}
\rotatebox{90}{} & \multirow[c]{3}{*}{LGBM} & $\Delta$ Cov & $0.002^{\pm 0.001}$ & $0.001^{\pm 0.000}$ \\
\rotatebox{90}{} &  & PI-Width & $73.3^{\pm 0.0}$ & $73.3^{\pm 0.0}$ \\
\rotatebox{90}{} &  & Winkler & $1.84^{\pm 0.00}$ & $1.83^{\pm 0.00}$ \\
\cmidrule{2-5}
\rotatebox{90}{} & \multirow[c]{3}{*}{Ridge} & $\Delta$ Cov & $0.010^{\pm 0.000}$ & $0.010^{\pm 0.000}$ \\
\rotatebox{90}{} &  & PI-Width & $65.5^{\pm 0.0}$ & $65.4^{\pm 0.1}$ \\
\rotatebox{90}{} &  & Winkler & $1.40^{\pm 0.00}$ & $1.40^{\pm 0.00}$ \\
\cmidrule{2-5}
\rotatebox{90}{} & \multirow[c]{3}{*}{LSTM} & $\Delta$ Cov & $-0.017^{\pm 0.000}$ & $-0.017^{\pm 0.000}$ \\
\rotatebox{90}{} &  & PI-Width & $32.4^{\pm 0.0}$ & $32.3^{\pm 0.0}$ \\
\rotatebox{90}{} &  & Winkler & $0.93^{\pm 0.00}$ & $0.92^{\pm 0.00}$ \\
\cmidrule{1-5}
\multirow[c]{12}{*}{\rotatebox{90}{Sap flow}} & \multirow[c]{3}{*}{Forest} & $\Delta$ Cov & $0.007^{\pm 0.000}$ & $0.007^{\pm 0.000}$ \\
\rotatebox{90}{} &  & PI-Width & $1743.1^{\pm 1.4}$ & $1742.7^{\pm 1.2}$ \\
\rotatebox{90}{} &  & Winkler & $0.59^{\pm 0.00}$ & $0.59^{\pm 0.00}$ \\
\cmidrule{2-5}
\rotatebox{90}{} & \multirow[c]{3}{*}{LGBM} & $\Delta$ Cov & $0.003^{\pm 0.000}$ & $0.003^{\pm 0.000}$ \\
\rotatebox{90}{} &  & PI-Width & $1581.9^{\pm 0.7}$ & $1583.6^{\pm 1.3}$ \\
\rotatebox{90}{} &  & Winkler & $0.49^{\pm 0.00}$ & $0.49^{\pm 0.00}$ \\
\cmidrule{2-5}
\rotatebox{90}{} & \multirow[c]{3}{*}{Ridge} & $\Delta$ Cov & $-0.241^{\pm 0.000}$ & $-0.242^{\pm 0.000}$ \\
\rotatebox{90}{} &  & PI-Width & $2061.5^{\pm 1.4}$ & $2066.8^{\pm 1.2}$ \\
\rotatebox{90}{} &  & Winkler & $2.52^{\pm 0.00}$ & $2.52^{\pm 0.00}$ \\
\cmidrule{2-5}
\rotatebox{90}{} & \multirow[c]{3}{*}{LSTM} & $\Delta$ Cov & $0.004^{\pm 0.000}$ & $0.006^{\pm 0.000}$ \\
\rotatebox{90}{} &  & PI-Width & $628.6^{\pm 0.8}$ & $631.9^{\pm 0.2}$ \\
\rotatebox{90}{} &  & Winkler & $0.24^{\pm 0.00}$ & $0.24^{\pm 0.00}$ \\
\bottomrule
\end{tabular}
\end{table}

\subsection{\spic Retraining}\label{sec:spic-retrained}
As our results show, \spic is very competitive (see, for example, Section~\ref{sec:results-discuss}).
\citet{xu2022sequential} did, however, design \spic so that its random forest regressor is retrained after each prediction step.
This design choice is computationally prohibitive for larger datasets.
To nevertheless allow a comparison against \spic on the larger datasets, we modified the algorithm so that the retraining is skipped.
Experiments on the smallest dataset (Table~\ref{tab:spic-retrain-comparsion}) show only small performance decrease with the modified algorithm.
One limitation of the adapted algorithm is, however, that a strong shift in the error distribution(s) would potentially require a retraining on the new distribution.
A viable proxy to detect such a change in our setting is the coverage performance of standard CP. The reason for this is that standard CP predictions are solely based on the calibration data and can thus not account for shifts.
In the experiments (Section~\ref{sec:experiments} and Appendix~\ref{sec:appendix-full-tables}), standard CP achieves good coverage with only one exception (see Section~\ref{sec:experiments}).
Hence, we decided to include \spic (in the modified version) to enable a comparison to it on larger datasets.

\setTableCaption{Performance of the original \spic algorithm (Retrain) and the modified version (No Retrain) on the \solarSmall dataset.
Performance in terms of the evaluation metrics is similar compared to the differences between methods (Section~\ref{sec:results-discuss}).
The computational demand of the modified version is considerably lower. The error term represents the standard deviation over repeated runs.}


\begin{table}
\centering
\caption{\nextTableCaption}
\label{tab:spic-retrain-comparsion}
\vskip 0.15in
\small
\begin{tabular}{lllrr}
\toprule
 &  & UC & \text{No Retrain} & \text{Retrain} \\
FC & $\alpha$ &  &  &  \\
\midrule
\multirow[c]{9}{*}{\rotatebox{90}{Forest}} & \multirow[c]{3}{*}{0.05} & $\Delta$ Cov & $-0.074^{\pm 0.001}$ & $-0.049^{\pm 0.001}$ \\
\rotatebox{90}{} &  & PI-Width & $58.0^{\pm 0.2}$ & $62.4^{\pm 0.0}$ \\
\rotatebox{90}{} &  & Winkler & $2.62^{\pm 0.03}$ & $2.28^{\pm 0.01}$ \\
\cmidrule{2-5}
\rotatebox{90}{} & \multirow[c]{3}{*}{0.10} & $\Delta$ Cov & $-0.064^{\pm 0.002}$ & $-0.045^{\pm 0.003}$ \\
\rotatebox{90}{} &  & PI-Width & $38.8^{\pm 0.4}$ & $41.6^{\pm 0.0}$ \\
\rotatebox{90}{} &  & Winkler & $1.82^{\pm 0.01}$ & $1.67^{\pm 0.00}$ \\
\cmidrule{2-5}
\rotatebox{90}{} & \multirow[c]{3}{*}{0.15} & $\Delta$ Cov & $-0.050^{\pm 0.002}$ & $-0.038^{\pm 0.003}$ \\
\rotatebox{90}{} &  & PI-Width & $26.8^{\pm 0.2}$ & $28.9^{\pm 0.0}$ \\
\rotatebox{90}{} &  & Winkler & $1.44^{\pm 0.02}$ & $1.35^{\pm 0.00}$ \\
\cmidrule{1-5}
\multirow[c]{9}{*}{\rotatebox{90}{LGBM}} & \multirow[c]{3}{*}{0.05} & $\Delta$ Cov & $-0.062^{\pm 0.001}$ & $-0.061^{\pm 0.001}$ \\
\rotatebox{90}{} &  & PI-Width & $56.1^{\pm 0.4}$ & $58.2^{\pm 0.0}$ \\
\rotatebox{90}{} &  & Winkler & $2.65^{\pm 0.03}$ & $2.51^{\pm 0.01}$ \\
\cmidrule{2-5}
\rotatebox{90}{} & \multirow[c]{3}{*}{0.10} & $\Delta$ Cov & $-0.052^{\pm 0.002}$ & $-0.049^{\pm 0.001}$ \\
\rotatebox{90}{} &  & PI-Width & $37.4^{\pm 0.2}$ & $39.8^{\pm 0.0}$ \\
\rotatebox{90}{} &  & Winkler & $1.84^{\pm 0.01}$ & $1.78^{\pm 0.01}$ \\
\cmidrule{2-5}
\rotatebox{90}{} & \multirow[c]{3}{*}{0.15} & $\Delta$ Cov & $-0.042^{\pm 0.001}$ & $-0.037^{\pm 0.001}$ \\
\rotatebox{90}{} &  & PI-Width & $26.9^{\pm 0.3}$ & $28.3^{\pm 0.0}$ \\
\rotatebox{90}{} &  & Winkler & $1.47^{\pm 0.00}$ & $1.42^{\pm 0.00}$ \\
\cmidrule{1-5}
\multirow[c]{9}{*}{\rotatebox{90}{Ridge}} & \multirow[c]{3}{*}{0.05} & $\Delta$ Cov & $-0.063^{\pm 0.001}$ & $-0.056^{\pm 0.002}$ \\
\rotatebox{90}{} &  & PI-Width & $67.3^{\pm 0.2}$ & $68.7^{\pm 0.0}$ \\
\rotatebox{90}{} &  & Winkler & $2.47^{\pm 0.01}$ & $2.25^{\pm 0.01}$ \\
\cmidrule{2-5}
\rotatebox{90}{} & \multirow[c]{3}{*}{0.10} & $\Delta$ Cov & $-0.055^{\pm 0.002}$ & $-0.048^{\pm 0.001}$ \\
\rotatebox{90}{} &  & PI-Width & $51.9^{\pm 0.2}$ & $54.2^{\pm 0.0}$ \\
\rotatebox{90}{} &  & Winkler & $1.91^{\pm 0.02}$ & $1.77^{\pm 0.01}$ \\
\cmidrule{2-5}
\rotatebox{90}{} & \multirow[c]{3}{*}{0.15} & $\Delta$ Cov & $-0.039^{\pm 0.003}$ & $-0.036^{\pm 0.000}$ \\
\rotatebox{90}{} &  & PI-Width & $43.1^{\pm 0.3}$ & $45.0^{\pm 0.0}$ \\
\rotatebox{90}{} &  & Winkler & $1.59^{\pm 0.00}$ & $1.50^{\pm 0.00}$ \\
\bottomrule
\end{tabular}
\end{table}

\begin{figure}[t]
\includegraphics[width=10cm]{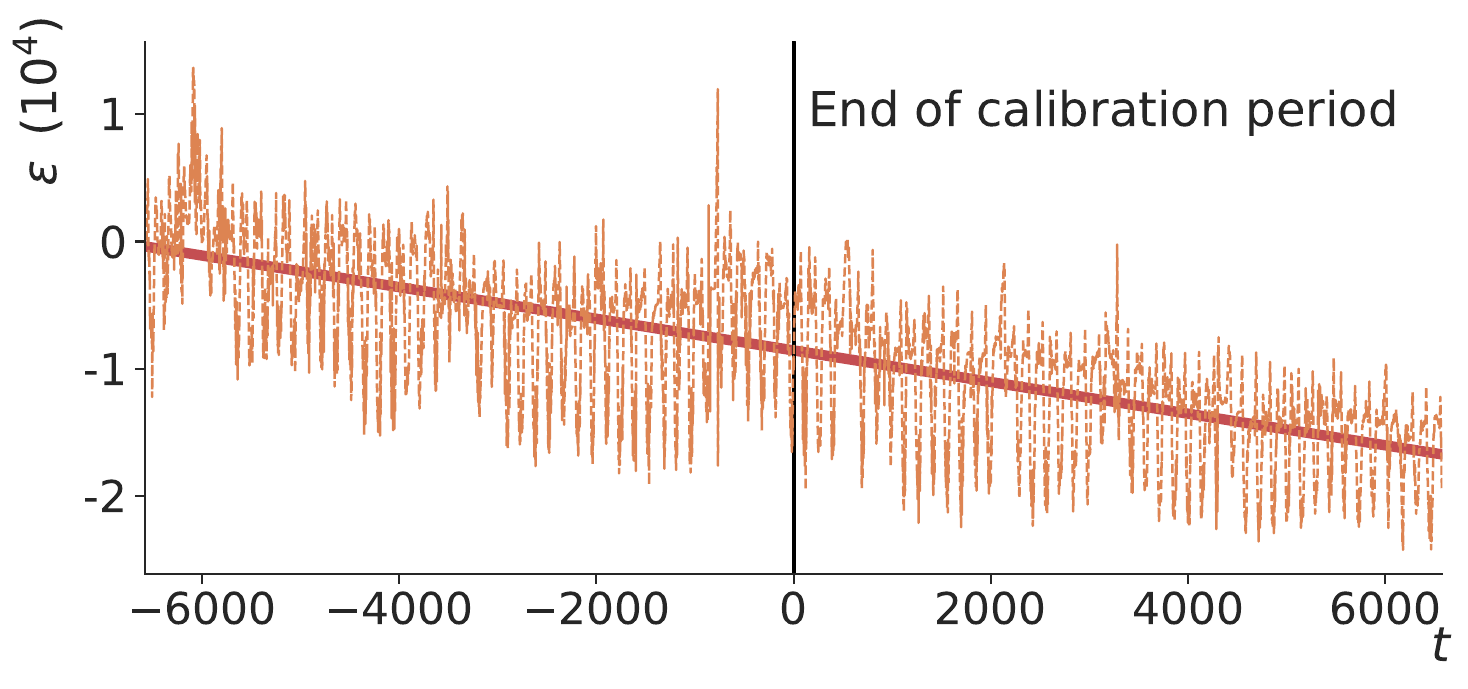}
\centering
\caption{Time series of the prediction error $\epsilon$ for the ridge regression model on the \sapflux dataset. The time series spans the calibration ($t < 0$) and test ($t \geq 0$) data.
The red line (fitted by the least squares method) shows a strong trend in the error distribution.}
\label{fig:eps-drift}
\end{figure}

\clearpage

\subsection{Additional Results} \label{sec:appendix-full-tables}
Tables~\ref{tab:nsdb-60m-2018-20-overall-v2}--\ref{tab:hydro-531-basin-list-overall-v2} show the results for miscoverage levels $\alpha \in \{0.05, 0.10, 0.15\}$ for all evaluated combinations of datasets and predictors.
Results for individual time series of the datasets are uploaded to the code repository (Appendix~\ref{sec:code}).

\paragraph{CMAL.} For the \hydro dataset, we additionally compare to CMAL, which is the state-of-the-art non-CP uncertainty estimation technique in the respective domain \citep{klotz2022uncertainty}.
CMAL is a mixture density network \citep{bishop1994mixture} based on an LSTM that predicts the parameters of asymmetric Laplacian distributions.
As our experiments use the same dataset, we adopt the hyperparameters from \citet{klotz2022uncertainty} but lower the learning rate to $0.0001$ because we train CMAL on more training samples (18 years, i.e., the training and calibration period combined, which allows for a fair comparison against the CP methods).
Despite the fact that CMAL is not based on the CP paradigm, it achieves good coverage results, however, at the cost of wide prediction intervals.
We argue that the Winkler score is low because CMAL considers all (and therefore also uncovered) samples during the optimization while CP methods do not consider them at all --- therefore, the distance of these points to the interval might be smaller (see Table~\ref{tab:hydro-531-basin-list-overall-v2}).

\begin{figure*}[t]
\includegraphics[width=\textwidth]{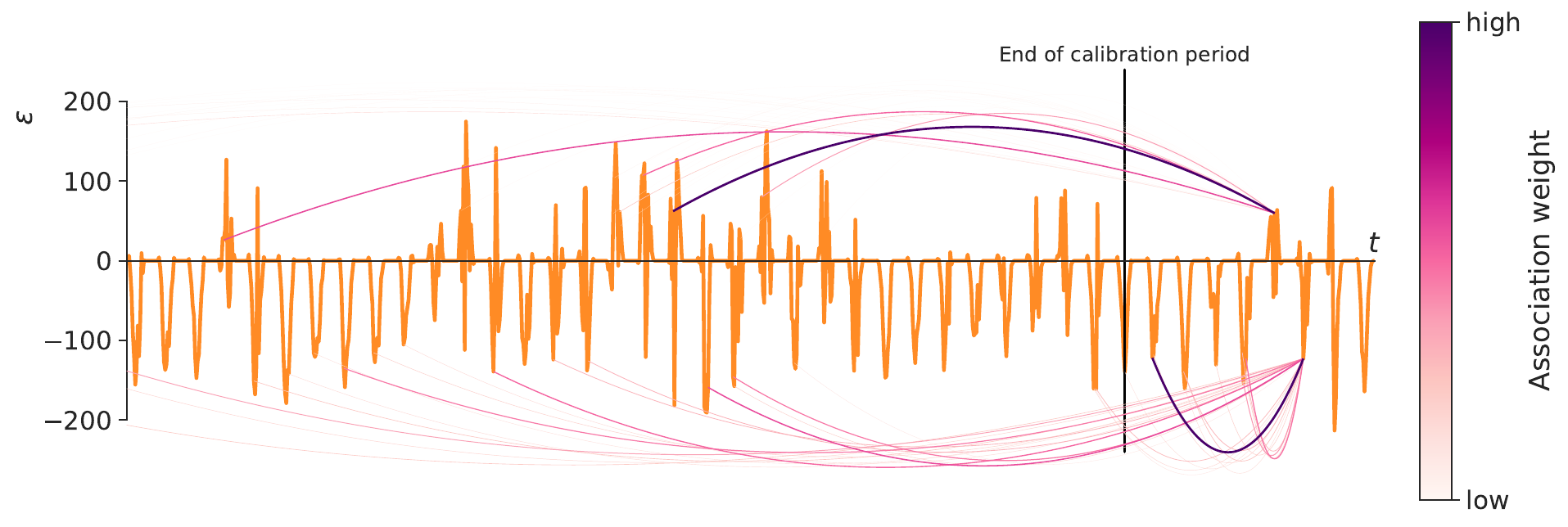}
\centering
\caption{Visualization of the 30 highest association weights that the Hopfield network places on previous time steps. \ourmethod retrieves similar error values when predicting at a time of high error, and it retrieves similar, previous small errors when predicting at a time step with small error.}
\label{fig:association-eps-viz}
\end{figure*}

\paragraph{Error trend.} Figure~\ref{fig:eps-drift} shows the prediction errors of the ridge regression model on the \sapflux dataset. The errors exhibit a strong trend and shift towards a negative expectation value.

\paragraph{Association weights.} Figure~\ref{fig:association-eps-viz} investigates the association patterns of \ourmethod and shows its capabilities to focus on time steps from similar error regimes.
The depicted time step in a regime with negative errors retrieves time steps with primarily negative errors and, likewise, the time step in a regime with positive errors retrieves time steps with primarily positive errors of the same magnitude.

\paragraph{Non-CP methods}

We also compare \ourmethod to non-CP methods to evaluate its performance within the broader field of uncertainty estimation for time series.
Specifically, we compare against a Monte Carlo Dropout (MCD); a Mixture Density Network (MDN), and a Gauss model (i.e., an MDN with a single Gaussian component), with an LSTM backbone (see Appendix~\ref{sec:hyperparam} for more details about the models and their hyperparameters).
This comparison shows that, although the non-CP methods often lead to very narrow prediction intervals, they fail to provide approximate coverage on most experiments (Table~\ref{tab:non-cp-comp-0}).

\newpage

\subsection{Local Coverage}\label{sec:local-coverage}
Standard CP only provides marginal coverage guarantees and a constant prediction interval width \citep{vovk2005algorithmic}.
In time series prediction tasks, this can lead to bad local coverage \citep{Lei2014DistributionfreePB}.
To evaluate whether the coverage approximately holds also locally, we evaluated $\Delta \ \text{Cov}$ on windows of size $k$.
To avoid compensation of negative $\Delta \ \text{Cov}$ in some windows by other windows with positive $\Delta \ \text{Cov}$, we upper-bounded each window's $\Delta \ \text{Cov}$ by zero before averaging over the bounded coverage gaps --- i.e., we calculated $\frac{1}{W} \sum_{w=1}^W - \Delta \  \text{Cov}_w^{\top 0}$, with 
\begin{align}
     \Delta \  \text{Cov}_w^{\top 0} = \begin{cases}
     \Delta \ \text{Cov}_w \; & \text{if} \ \Delta \ \text{Cov}_w \leq 0, \\
        0 \; & \text{else},
    \end{cases}
\end{align}

where $ \Delta \ \text{Cov}_w$ is the $\Delta \ \text{Cov}$ within window $w$.

Table~\ref{tab:rolling-coverage-v2} shows the results of this evaluation for window sizes $k \in \{10, 20, 50\}$ and miscoverage level $\alpha=0.1$.
Depending on the dataset, most often either \ourmethod or \spic perform best.
The overall rankings are only partly similar to the evaluation of the marginal coverage (Table~\ref{tab:main-results-alpha-0.1} and Appendix~\ref{sec:appendix-full-tables}).
Especially standard CP, which achieves competitive results in marginal coverage, falls short in this comparison.
Only for \solarSmall, where the approach achieves a high marginal $\Delta \ \text{Cov}$, it preserves the local coverage best.
Note that this comes with the drawback of very wide prediction intervals, i.e., bad efficiency.
Overall, the results show that \ourmethod and other time series CP methods improve the local coverage in non-exchangeable time series settings, compared to standard CP.

\subsection{Negative Results}
We also tested training with more involved training procedures than directly using the mean squared error.
However, based on our empirical findings, the vanilla method with mean squared error outperforms more complex approaches. In the following, we briefly describe those additional configurations (so that potential future research can avoid these paths):

\begin{itemize}
    \item \textbf{Pinball Loss}. We tried to train the MHN with a pinball loss \citep[the original use seems to stem from][we are however not aware of the naming origin]{fox1964admissibility} in many different variations (see points that follow), but consistently got worse results than with the mean squared error.
    \begin{itemize}
        \item Inspired by the ideas lined out by \citet{tagasovska2019single} we tried to use the miscoverage level as an additional input to the network, while also parameterizing the loss with it. 
        \item We tried to use the pinball loss to predict multiple miscoverage values at the same time to get a more informed representation of the distribution we want to approximate. 
    \end{itemize}
    \item \textbf{Softmax Probabilities}. We tried to use the softmax output of the MHN to directly estimate probabilities and use a maximum likelihood procedure. This did train, but it did not produce good results.
\end{itemize}

\subsection{Computational Resources}\label{sec:comutatation-resources}

We used different hardware setups in our experiments, however, most of them were executed on a machine with an Nvidia P100 GPU and a Xeon E5-2698 CPU.
The runtime differs greatly between different dataset sizes. We report the approximate run times for a single experiment (i.e., one seed, all evaluated coverage levels, not including training the prediction model $\muh$):

\begin{enumerate}
    \item \textbf{\ourmethod}: \solarThreeY 5h, \solarOneY 1h, \solarSmall 7min, \airTen 3h, \airTwenty 3h, \sapflux 2.5h, \hydro 12--20h
    \item  \textbf{\spic}: (adapted): \solarThreeY 5--10 days, \solarOneY 10--30h, \solarSmall 45min,  \airTen 13--30h, \airTwenty 13--30h, \sapflux 20--50h, \hydro 12--17 days
    \item \textbf{\enbpi}: all datasets under 6h
    \item \textbf{\nexcp}: all datasets under 45min
    \item \textbf{\adaptiveci}: all datasets under 45min
    \item \textbf{Standard CP}: all datasets under 45min
\end{enumerate}

\setTableCaption{Performance of the best two CP methods (\ourmethod and \spic) and state-of-the-art non-CP methods for the different datasets and levels of $\alpha$.
All approaches build upon an LSTM model.
Bold numbers correspond to the best result for the respective metric in the experiment (PI-Width and Winkler score), given that $\Delta \text{Cov} \ge - 0.25 \alpha$, i.e., the specific algorithm reached at least approximate coverage (the result is grayed out otherwise).
The error term represents the standard deviation over repeated runs with different seeds (results without an error term are from deterministic models).}



\begin{table*}
\centering
\caption{\nextTableCaption}
\label{tab:non-cp-comp-0}
\vskip 0.15in
\small
\begin{tabular}{llrrr|rrr}
\toprule
 &  &  & \text{\ourmethod} & \text{SPCI} & MCD & Gauss & MDN \\
\text{} & \text{$\alpha$} &  &  &  &  &  &  \\
\midrule
\multirow[c]{9}{*}{\rotatebox{90}{\text{Solar 3Y}}} & \multirow[c]{3}{*}{\rotatebox{90}{0.05}} & \text{$\Delta$ Cov} & $\text{-}0.003^{\pm 0.005}$ & $0.004^{\pm 0.000}$ & $\textcolor{gray}{\text{-}0.058}^{\pm 0.004}$ & $\textcolor{gray}{\text{-}0.331}^{\pm 0.243}$ & $\text{-}0.005^{\pm 0.013}$ \\
\rotatebox{90}{} &  & \text{PI-Width} & $22.7^{\pm 0.8}$ & $47.7^{\pm 0.0}$ & $\textcolor{gray}{17.5}^{\pm 0.1}$ & $\textcolor{gray}{25.9}^{\pm 4.1}$ & $\textbf{16.2}^{\pm 1.0}$ \\
\rotatebox{90}{} &  & \text{Winkler} & $0.38^{\pm 0.01}$ & $0.87^{\pm 0.00}$ & $\textcolor{gray}{0.75}^{\pm 0.01}$ & $\textcolor{gray}{0.46}^{\pm 0.10}$ & $\textbf{0.33}^{\pm 0.06}$ \\
\cmidrule{2-8}
\rotatebox{90}{} & \multirow[c]{3}{*}{\rotatebox{90}{0.10}} & \text{$\Delta$ Cov} & $0.001^{\pm 0.006}$ & $0.014^{\pm 0.000}$ & $\textcolor{gray}{\text{-}0.030}^{\pm 0.006}$ & $\textcolor{gray}{\text{-}0.310}^{\pm 0.238}$ & $\text{-}0.013^{\pm 0.047}$ \\
\rotatebox{90}{} &  & \text{PI-Width} & $17.9^{\pm 0.6}$ & $27.8^{\pm 0.0}$ & $\textcolor{gray}{14.8}^{\pm 0.1}$ & $\textcolor{gray}{21.8}^{\pm 3.5}$ & $\textbf{13.2}^{\pm 0.8}$ \\
\rotatebox{90}{} &  & \text{Winkler} & $0.30^{\pm 0.01}$ & $0.62^{\pm 0.00}$ & $\textcolor{gray}{0.49}^{\pm 0.00}$ & $\textcolor{gray}{0.37}^{\pm 0.06}$ & $\textbf{0.26}^{\pm 0.03}$ \\
\cmidrule{2-8}
\rotatebox{90}{} & \multirow[c]{3}{*}{\rotatebox{90}{0.15}} & \text{$\Delta$ Cov} & $0.002^{\pm 0.007}$ & $0.024^{\pm 0.000}$ & $0.001^{\pm 0.007}$ & $\textcolor{gray}{\text{-}0.320}^{\pm 0.166}$ & $\textcolor{gray}{\text{-}0.053}^{\pm 0.141}$ \\
\rotatebox{90}{} &  & \text{PI-Width} & $15.0^{\pm 0.4}$ & $18.1^{\pm 0.0}$ & $\textbf{12.9}^{\pm 0.1}$ & $\textcolor{gray}{19.1}^{\pm 3.0}$ & $\textcolor{gray}{11.4}^{\pm 0.8}$ \\
\rotatebox{90}{} &  & \text{Winkler} & $\textbf{0.26}^{\pm 0.00}$ & $0.48^{\pm 0.00}$ & $0.38^{\pm 0.00}$ & $\textcolor{gray}{0.32}^{\pm 0.04}$ & $\textcolor{gray}{0.23}^{\pm 0.02}$ \\
\cmidrule{1-8} \cmidrule{2-8}
\multirow[c]{9}{*}{\rotatebox{90}{\text{Solar 1Y}}} & \multirow[c]{3}{*}{\rotatebox{90}{0.05}} & \text{$\Delta$ Cov} & $0.019^{\pm 0.003}$ & $0.006^{\pm 0.000}$ & $\textcolor{gray}{\text{-}0.041}^{\pm 0.004}$ & $\textcolor{gray}{\text{-}0.109}^{\pm 0.191}$ & $\textcolor{gray}{\text{-}0.055}^{\pm 0.104}$ \\
\rotatebox{90}{} &  & \text{PI-Width} & $\textbf{21.4}^{\pm 0.7}$ & $37.0^{\pm 0.0}$ & $\textcolor{gray}{20.5}^{\pm 0.3}$ & $\textcolor{gray}{14.6}^{\pm 0.8}$ & $\textcolor{gray}{13.2}^{\pm 1.4}$ \\
\rotatebox{90}{} &  & \text{Winkler} & $\textbf{0.27}^{\pm 0.01}$ & $0.58^{\pm 0.00}$ & $\textcolor{gray}{0.58}^{\pm 0.01}$ & $\textcolor{gray}{0.22}^{\pm 0.01}$ & $\textcolor{gray}{0.20}^{\pm 0.03}$ \\
\cmidrule{2-8}
\rotatebox{90}{} & \multirow[c]{3}{*}{\rotatebox{90}{0.10}} & \text{$\Delta$ Cov} & $0.028^{\pm 0.010}$ & $0.018^{\pm 0.000}$ & $\text{-}0.011^{\pm 0.005}$ & $\textcolor{gray}{\text{-}0.134}^{\pm 0.253}$ & $\textcolor{gray}{\text{-}0.130}^{\pm 0.182}$ \\
\rotatebox{90}{} &  & \text{PI-Width} & $\textbf{16.0}^{\pm 0.6}$ & $22.5^{\pm 0.0}$ & $17.3^{\pm 0.2}$ & $\textcolor{gray}{12.3}^{\pm 0.7}$ & $\textcolor{gray}{10.8}^{\pm 1.2}$ \\
\rotatebox{90}{} &  & \text{Winkler} & $\textbf{0.22}^{\pm 0.01}$ & $0.41^{\pm 0.00}$ & $0.40^{\pm 0.01}$ & $\textcolor{gray}{0.18}^{\pm 0.01}$ & $\textcolor{gray}{0.17}^{\pm 0.02}$ \\
\cmidrule{2-8}
\rotatebox{90}{} & \multirow[c]{3}{*}{\rotatebox{90}{0.15}} & \text{$\Delta$ Cov} & $0.029^{\pm 0.017}$ & $0.032^{\pm 0.000}$ & $0.022^{\pm 0.006}$ & $\textcolor{gray}{\text{-}0.148}^{\pm 0.236}$ & $\textcolor{gray}{\text{-}0.185}^{\pm 0.224}$ \\
\rotatebox{90}{} &  & \text{PI-Width} & $\textbf{13.1}^{\pm 0.5}$ & $15.3^{\pm 0.0}$ & $15.2^{\pm 0.2}$ & $\textcolor{gray}{10.8}^{\pm 0.6}$ & $\textcolor{gray}{9.3}^{\pm 1.1}$ \\
\rotatebox{90}{} &  & \text{Winkler} & $\textbf{0.19}^{\pm 0.00}$ & $0.32^{\pm 0.00}$ & $0.32^{\pm 0.00}$ & $\textcolor{gray}{0.16}^{\pm 0.01}$ & $\textcolor{gray}{0.15}^{\pm 0.02}$ \\
\cmidrule{1-8} \cmidrule{2-8}
\multirow[c]{9}{*}{\rotatebox{90}{\text{Air 10 PM}}} & \multirow[c]{3}{*}{\rotatebox{90}{0.05}} & \text{$\Delta$ Cov} & $\text{-}0.001^{\pm 0.001}$ & $0.003^{\pm 0.000}$ & $\textcolor{gray}{\text{-}0.202}^{\pm 0.008}$ & $\textcolor{gray}{\text{-}0.191}^{\pm 0.024}$ & $\textcolor{gray}{\text{-}0.218}^{\pm 0.115}$ \\
\rotatebox{90}{} &  & \text{PI-Width} & $90.7^{\pm 1.9}$ & $\textbf{86.1}^{\pm 0.0}$ & $\textcolor{gray}{42.3}^{\pm 0.8}$ & $\textcolor{gray}{45.8}^{\pm 2.2}$ & $\textcolor{gray}{43.9}^{\pm 9.5}$ \\
\rotatebox{90}{} &  & \text{Winkler} & $1.83^{\pm 0.03}$ & $\textbf{1.63}^{\pm 0.00}$ & $\textcolor{gray}{2.36}^{\pm 0.03}$ & $\textcolor{gray}{2.08}^{\pm 0.12}$ & $\textcolor{gray}{2.22}^{\pm 0.59}$ \\
\cmidrule{2-8}
\rotatebox{90}{} & \multirow[c]{3}{*}{\rotatebox{90}{0.10}} & \text{$\Delta$ Cov} & $\text{-}0.002^{\pm 0.005}$ & $0.010^{\pm 0.000}$ & $\textcolor{gray}{\text{-}0.201}^{\pm 0.009}$ & $\textcolor{gray}{\text{-}0.211}^{\pm 0.026}$ & $\textcolor{gray}{\text{-}0.243}^{\pm 0.122}$ \\
\rotatebox{90}{} &  & \text{PI-Width} & $62.7^{\pm 1.5}$ & $\textbf{62.2}^{\pm 0.0}$ & $\textcolor{gray}{35.7}^{\pm 0.7}$ & $\textcolor{gray}{38.6}^{\pm 1.9}$ & $\textcolor{gray}{35.3}^{\pm 7.3}$ \\
\rotatebox{90}{} &  & \text{Winkler} & $1.33^{\pm 0.01}$ & $\textbf{1.21}^{\pm 0.00}$ & $\textcolor{gray}{1.49}^{\pm 0.01}$ & $\textcolor{gray}{1.42}^{\pm 0.06}$ & $\textcolor{gray}{1.48}^{\pm 0.30}$ \\
\cmidrule{2-8}
\rotatebox{90}{} & \multirow[c]{3}{*}{\rotatebox{90}{0.15}} & \text{$\Delta$ Cov} & $\text{-}0.002^{\pm 0.010}$ & $0.017^{\pm 0.000}$ & $\textcolor{gray}{\text{-}0.192}^{\pm 0.009}$ & $\textcolor{gray}{\text{-}0.218}^{\pm 0.027}$ & $\textcolor{gray}{\text{-}0.252}^{\pm 0.122}$ \\
\rotatebox{90}{} &  & \text{PI-Width} & $\textbf{49.4}^{\pm 1.7}$ & $50.2^{\pm 0.0}$ & $\textcolor{gray}{31.4}^{\pm 0.6}$ & $\textcolor{gray}{33.8}^{\pm 1.7}$ & $\textcolor{gray}{30.2}^{\pm 6.0}$ \\
\rotatebox{90}{} &  & \text{Winkler} & $1.09^{\pm 0.01}$ & $\textbf{1.01}^{\pm 0.00}$ & $\textcolor{gray}{1.16}^{\pm 0.01}$ & $\textcolor{gray}{1.15}^{\pm 0.04}$ & $\textcolor{gray}{1.17}^{\pm 0.20}$ \\
\cmidrule{1-8} \cmidrule{2-8}
\multirow[c]{9}{*}{\rotatebox{90}{\text{Air 25 PM}}} & \multirow[c]{3}{*}{\rotatebox{90}{0.05}} & \text{$\Delta$ Cov} & $0.005^{\pm 0.005}$ & $\textcolor{gray}{\text{-}0.015}^{\pm 0.000}$ & $\textcolor{gray}{\text{-}0.100}^{\pm 0.022}$ & $\textcolor{gray}{\text{-}0.227}^{\pm 0.018}$ & $\textcolor{gray}{\text{-}0.209}^{\pm 0.119}$ \\
\rotatebox{90}{} &  & \text{PI-Width} & $\textbf{57.1}^{\pm 5.6}$ & $\textcolor{gray}{44.9}^{\pm 0.0}$ & $\textcolor{gray}{31.7}^{\pm 3.5}$ & $\textcolor{gray}{27.1}^{\pm 1.4}$ & $\textcolor{gray}{26.9}^{\pm 6.1}$ \\
\rotatebox{90}{} &  & \text{Winkler} & $\textbf{1.19}^{\pm 0.08}$ & $\textcolor{gray}{1.29}^{\pm 0.00}$ & $\textcolor{gray}{1.41}^{\pm 0.05}$ & $\textcolor{gray}{1.58}^{\pm 0.08}$ & $\textcolor{gray}{1.57}^{\pm 0.47}$ \\
\cmidrule{2-8}
\rotatebox{90}{} & \multirow[c]{3}{*}{\rotatebox{90}{0.10}} & \text{$\Delta$ Cov} & $0.007^{\pm 0.008}$ & $\text{-}0.017^{\pm 0.000}$ & $\textcolor{gray}{\text{-}0.095}^{\pm 0.027}$ & $\textcolor{gray}{\text{-}0.247}^{\pm 0.020}$ & $\textcolor{gray}{\text{-}0.234}^{\pm 0.129}$ \\
\rotatebox{90}{} &  & \text{PI-Width} & $40.7^{\pm 4.3}$ & $\textbf{32.4}^{\pm 0.0}$ & $\textcolor{gray}{26.6}^{\pm 3.0}$ & $\textcolor{gray}{22.8}^{\pm 1.2}$ & $\textcolor{gray}{21.6}^{\pm 4.9}$ \\
\rotatebox{90}{} &  & \text{Winkler} & $\textbf{0.88}^{\pm 0.05}$ & $0.93^{\pm 0.00}$ & $\textcolor{gray}{0.95}^{\pm 0.03}$ & $\textcolor{gray}{1.06}^{\pm 0.04}$ & $\textcolor{gray}{1.05}^{\pm 0.24}$ \\
\cmidrule{2-8}
\rotatebox{90}{} & \multirow[c]{3}{*}{\rotatebox{90}{0.15}} & \text{$\Delta$ Cov} & $0.005^{\pm 0.011}$ & $\text{-}0.016^{\pm 0.000}$ & $\textcolor{gray}{\text{-}0.086}^{\pm 0.030}$ & $\textcolor{gray}{\text{-}0.253}^{\pm 0.020}$ & $\textcolor{gray}{\text{-}0.243}^{\pm 0.131}$ \\
\rotatebox{90}{} &  & \text{PI-Width} & $32.5^{\pm 3.7}$ & $\textbf{26.1}^{\pm 0.0}$ & $\textcolor{gray}{23.2}^{\pm 2.6}$ & $\textcolor{gray}{20.0}^{\pm 1.0}$ & $\textcolor{gray}{18.4}^{\pm 4.2}$ \\
\rotatebox{90}{} &  & \text{Winkler} & $\textbf{0.74}^{\pm 0.04}$ & $0.76^{\pm 0.00}$ & $\textcolor{gray}{0.76}^{\pm 0.02}$ & $\textcolor{gray}{0.85}^{\pm 0.02}$ & $\textcolor{gray}{0.84}^{\pm 0.16}$ \\
\cmidrule{1-8} \cmidrule{2-8}
\multirow[c]{9}{*}{\rotatebox{90}{\text{Sap flow}}} & \multirow[c]{3}{*}{\rotatebox{90}{0.05}} & \text{$\Delta$ Cov} & $0.001^{\pm 0.002}$ & $0.000^{\pm 0.000}$ & $\textcolor{gray}{\text{-}0.150}^{\pm 0.007}$ & $\textcolor{gray}{\text{-}0.072}^{\pm 0.018}$ & $\textcolor{gray}{\text{-}0.059}^{\pm 0.035}$ \\
\rotatebox{90}{} &  & \text{PI-Width} & $\textbf{783.5}^{\pm 9.2}$ & $897.7^{\pm 0.9}$ & $\textcolor{gray}{421.6}^{\pm 5.2}$ & $\textcolor{gray}{494.7}^{\pm 26.0}$ & $\textcolor{gray}{494.6}^{\pm 41.2}$ \\
\rotatebox{90}{} &  & \text{Winkler} & $\textbf{0.25}^{\pm 0.01}$ & $0.33^{\pm 0.00}$ & $\textcolor{gray}{0.49}^{\pm 0.01}$ & $\textcolor{gray}{0.29}^{\pm 0.02}$ & $\textcolor{gray}{0.27}^{\pm 0.06}$ \\
\cmidrule{2-8}
\rotatebox{90}{} & \multirow[c]{3}{*}{\rotatebox{90}{0.10}} & \text{$\Delta$ Cov} & $0.004^{\pm 0.004}$ & $0.004^{\pm 0.000}$ & $\textcolor{gray}{\text{-}0.139}^{\pm 0.007}$ & $\textcolor{gray}{\text{-}0.067}^{\pm 0.022}$ & $\textcolor{gray}{\text{-}0.067}^{\pm 0.037}$ \\
\rotatebox{90}{} &  & \text{PI-Width} & $\textbf{594.3}^{\pm 7.7}$ & $628.2^{\pm 0.9}$ & $\textcolor{gray}{355.2}^{\pm 4.3}$ & $\textcolor{gray}{416.5}^{\pm 21.9}$ & $\textcolor{gray}{391.7}^{\pm 31.4}$ \\
\rotatebox{90}{} &  & \text{Winkler} & $\textbf{0.19}^{\pm 0.01}$ & $0.24^{\pm 0.00}$ & $\textcolor{gray}{0.30}^{\pm 0.01}$ & $\textcolor{gray}{0.20}^{\pm 0.01}$ & $\textcolor{gray}{0.19}^{\pm 0.03}$ \\
\cmidrule{2-8}
\rotatebox{90}{} & \multirow[c]{3}{*}{\rotatebox{90}{0.15}} & \text{$\Delta$ Cov} & $0.005^{\pm 0.005}$ & $0.006^{\pm 0.000}$ & $\textcolor{gray}{\text{-}0.122}^{\pm 0.007}$ & $\textcolor{gray}{\text{-}0.059}^{\pm 0.024}$ & $\textcolor{gray}{\text{-}0.068}^{\pm 0.039}$ \\
\rotatebox{90}{} &  & \text{PI-Width} & $\textbf{489.9}^{\pm 7.0}$ & $493.2^{\pm 0.5}$ & $\textcolor{gray}{311.4}^{\pm 3.8}$ & $\textcolor{gray}{364.9}^{\pm 19.2}$ & $\textcolor{gray}{331.5}^{\pm 25.8}$ \\
\rotatebox{90}{} &  & \text{Winkler} & $\textbf{0.17}^{\pm 0.01}$ & $0.20^{\pm 0.00}$ & $\textcolor{gray}{0.23}^{\pm 0.01}$ & $\textcolor{gray}{0.17}^{\pm 0.01}$ & $\textcolor{gray}{0.16}^{\pm 0.02}$ \\
\bottomrule
\end{tabular}
\end{table*}

\setTableCaption{Negative average of zero-upper-bounded $\Delta \ \text{Cov}$ of rolling windows with miscoverage level $\alpha=0.1$. We evaluate each dataset--predictor combination at the windows sizes $k \in \{10, 20, 50\}$.}


\begin{table}
\centering
\caption{\nextTableCaption}
\label{tab:rolling-coverage-v2}
\vskip 0.15in
\small
\begin{tabular}{llllllll}
\toprule
 &  & & \rotatebox{90}{\ourmethod} & \rotatebox{90}{SPCI} & \rotatebox{90}{EnbPI} & \rotatebox{90}{NexCP} & \rotatebox{90}{CP} \\
Data & FC & k &  &  &  &  &  \\
\midrule
\multirow[c]{9}{*}{\rotatebox{90}{Solar 3Y}} & \multirow[c]{3}{*}{\rotatebox{90}{Forest}} & 10 & \boldmath$.030$ & $.040$ & $.081$ & $.059$ & $.059$ \\
\rotatebox{90}{} & \rotatebox{90}{} & 20 & \boldmath$.020$ & $.022$ & $.055$ & $.034$ & $.040$ \\
\rotatebox{90}{} & \rotatebox{90}{} & 50 & \boldmath$.012$ & $.016$ & $.042$ & $.021$ & $.032$ \\
\cline{2-8}
\rotatebox{90}{} & \multirow[c]{3}{*}{\rotatebox{90}{LGBM}} & 10 & $.051$ & \boldmath$.040$ & $.076$ & $.059$ & $.058$ \\
\rotatebox{90}{} & \rotatebox{90}{} & 20 & $.040$ & \boldmath$.022$ & $.052$ & $.035$ & $.039$ \\
\rotatebox{90}{} & \rotatebox{90}{} & 50 & $.031$ & \boldmath$.015$ & $.038$ & $.022$ & $.031$ \\
\cline{2-8}
\rotatebox{90}{} & \multirow[c]{3}{*}{\rotatebox{90}{Ridge}} & 10 & \boldmath$.023$ & $.049$ & $.099$ & $.055$ & $.056$ \\
\rotatebox{90}{} & \rotatebox{90}{} & 20 & \boldmath$.014$ & $.035$ & $.083$ & $.039$ & $.040$ \\
\rotatebox{90}{} & \rotatebox{90}{} & 50 & \boldmath$.007$ & $.028$ & $.075$ & $.027$ & $.030$ \\
\cline{1-8}
\multirow[c]{9}{*}{\rotatebox{90}{Solar 1Y}} & \multirow[c]{3}{*}{\rotatebox{90}{Forest}} & 10 & $.023$ & $.023$ & $.070$ & $.055$ & \boldmath$.019$ \\
\rotatebox{90}{} & \rotatebox{90}{} & 20 & $.015$ & $.012$ & $.043$ & $.031$ & \boldmath$.010$ \\
\rotatebox{90}{} & \rotatebox{90}{} & 50 & $.010$ & $.008$ & $.031$ & $.017$ & \boldmath$.005$ \\
\cline{2-8}
\rotatebox{90}{} & \multirow[c]{3}{*}{\rotatebox{90}{LGBM}} & 10 & $.059$ & $.023$ & $.071$ & $.056$ & \boldmath$.018$ \\
\rotatebox{90}{} & \rotatebox{90}{} & 20 & $.049$ & $.012$ & $.046$ & $.032$ & \boldmath$.010$ \\
\rotatebox{90}{} & \rotatebox{90}{} & 50 & $.041$ & $.007$ & $.033$ & $.018$ & \boldmath$.005$ \\
\cline{2-8}
\rotatebox{90}{} & \multirow[c]{3}{*}{\rotatebox{90}{Ridge}} & 10 & $.063$ & \boldmath$.030$ & $.070$ & $.055$ & $.066$ \\
\rotatebox{90}{} & \rotatebox{90}{} & 20 & $.053$ & \boldmath$.021$ & $.055$ & $.038$ & $.052$ \\
\rotatebox{90}{} & \rotatebox{90}{} & 50 & $.045$ & \boldmath$.016$ & $.043$ & $.027$ & $.045$ \\
\cline{1-8}
\multirow[c]{9}{*}{\rotatebox{90}{Solar Small}} & \multirow[c]{3}{*}{\rotatebox{90}{Forest}} & 10 & \boldmath$.045$ & $.108$ & $.074$ & $.075$ & $.083$ \\
\rotatebox{90}{} & \rotatebox{90}{} & 20 & \boldmath$.026$ & $.074$ & $.046$ & $.045$ & $.055$ \\
\rotatebox{90}{} & \rotatebox{90}{} & 50 & \boldmath$.015$ & $.067$ & $.034$ & $.030$ & $.042$ \\
\cline{2-8}
\rotatebox{90}{} & \multirow[c]{3}{*}{\rotatebox{90}{LGBM}} & 10 & \boldmath$.048$ & $.093$ & $.074$ & $.074$ & $.078$ \\
\rotatebox{90}{} & \rotatebox{90}{} & 20 & \boldmath$.032$ & $.061$ & $.044$ & $.044$ & $.048$ \\
\rotatebox{90}{} & \rotatebox{90}{} & 50 & \boldmath$.020$ & $.051$ & $.030$ & $.029$ & $.035$ \\
\cline{2-8}
\rotatebox{90}{} & \multirow[c]{3}{*}{\rotatebox{90}{Ridge}} & 10 & \boldmath$.051$ & $.100$ & $.065$ & $.058$ & $.064$ \\
\rotatebox{90}{} & \rotatebox{90}{} & 20 & \boldmath$.039$ & $.073$ & $.049$ & \boldmath$.039$ & $.045$ \\
\rotatebox{90}{} & \rotatebox{90}{} & 50 & \boldmath$.032$ & $.066$ & $.041$ & $.033$ & $.038$ \\
\cline{1-8}
\multirow[c]{9}{*}{\rotatebox{90}{Air 10 PM}} & \multirow[c]{3}{*}{\rotatebox{90}{Forest}} & 10 & \boldmath$.033$ & $.049$ & $.124$ & $.075$ & $.106$ \\
\rotatebox{90}{} & \rotatebox{90}{} & 20 & \boldmath$.025$ & $.042$ & $.114$ & $.067$ & $.100$ \\
\rotatebox{90}{} & \rotatebox{90}{} & 50 & \boldmath$.018$ & $.035$ & $.099$ & $.053$ & $.089$ \\
\cline{2-8}
\rotatebox{90}{} & \multirow[c]{3}{*}{\rotatebox{90}{LGBM}} & 10 & $.038$ & \boldmath$.037$ & $.115$ & $.074$ & $.100$ \\
\rotatebox{90}{} & \rotatebox{90}{} & 20 & \boldmath$.030$ & \boldmath$.030$ & $.104$ & $.065$ & $.093$ \\
\rotatebox{90}{} & \rotatebox{90}{} & 50 & \boldmath$.021$ & $.023$ & $.089$ & $.052$ & $.083$ \\
\cline{2-8}
\rotatebox{90}{} & \multirow[c]{3}{*}{\rotatebox{90}{Ridge}} & 10 & $.041$ & \boldmath$.035$ & $.099$ & $.068$ & $.064$ \\
\rotatebox{90}{} & \rotatebox{90}{} & 20 & $.032$ & \boldmath$.028$ & $.089$ & $.059$ & $.057$ \\
\rotatebox{90}{} & \rotatebox{90}{} & 50 & $.024$ & \boldmath$.020$ & $.074$ & $.045$ & $.048$ \\
\cline{1-8}
\multirow[c]{9}{*}{\rotatebox{90}{Air 25 PM}} & \multirow[c]{3}{*}{\rotatebox{90}{Forest}} & 10 & $.076$ & \boldmath$.061$ & $.139$ & $.080$ & $.114$ \\
\rotatebox{90}{} & \rotatebox{90}{} & 20 & $.067$ & \boldmath$.053$ & $.129$ & $.071$ & $.108$ \\
\rotatebox{90}{} & \rotatebox{90}{} & 50 & $.056$ & \boldmath$.044$ & $.113$ & $.056$ & $.096$ \\
\cline{2-8}
\rotatebox{90}{} & \multirow[c]{3}{*}{\rotatebox{90}{LGBM}} & 10 & $.073$ & \boldmath$.051$ & $.124$ & $.080$ & $.113$ \\
\rotatebox{90}{} & \rotatebox{90}{} & 20 & $.064$ & \boldmath$.044$ & $.114$ & $.072$ & $.105$ \\
\rotatebox{90}{} & \rotatebox{90}{} & 50 & $.053$ & \boldmath$.035$ & $.097$ & $.059$ & $.093$ \\
\cline{2-8}
\rotatebox{90}{} & \multirow[c]{3}{*}{\rotatebox{90}{Ridge}} & 10 & $.057$ & \boldmath$.043$ & $.107$ & $.075$ & $.079$ \\
\rotatebox{90}{} & \rotatebox{90}{} & 20 & $.048$ & \boldmath$.035$ & $.096$ & $.066$ & $.072$ \\
\rotatebox{90}{} & \rotatebox{90}{} & 50 & $.039$ & \boldmath$.027$ & $.081$ & $.053$ & $.061$ \\
\cline{1-8}
\multirow[c]{9}{*}{\rotatebox{90}{Sap flow}} & \multirow[c]{3}{*}{\rotatebox{90}{Forest}} & 10 & $.070$ & \boldmath$.058$ & $.107$ & $.076$ & $.073$ \\
\rotatebox{90}{} & \rotatebox{90}{} & 20 & $.062$ & \boldmath$.050$ & $.097$ & $.068$ & $.067$ \\
\rotatebox{90}{} & \rotatebox{90}{} & 50 & $.051$ & \boldmath$.042$ & $.082$ & $.056$ & $.059$ \\
\cline{2-8}
\rotatebox{90}{} & \multirow[c]{3}{*}{\rotatebox{90}{LGBM}} & 10 & $.107$ & \boldmath$.060$ & $.105$ & $.078$ & $.082$ \\
\rotatebox{90}{} & \rotatebox{90}{} & 20 & $.098$ & \boldmath$.052$ & $.094$ & $.069$ & $.075$ \\
\rotatebox{90}{} & \rotatebox{90}{} & 50 & $.086$ & \boldmath$.045$ & $.078$ & $.056$ & $.067$ \\
\cline{2-8}
\rotatebox{90}{} & \multirow[c]{3}{*}{\rotatebox{90}{Ridge}} & 10 & $.097$ & $.286$ & $.102$ & \boldmath$.086$ & $.404$ \\
\rotatebox{90}{} & \rotatebox{90}{} & 20 & $.088$ & $.279$ & $.089$ & \boldmath$.077$ & $.397$ \\
\rotatebox{90}{} & \rotatebox{90}{} & 50 & $.075$ & $.271$ & $.071$ & \boldmath$.063$ & $.390$ \\
\bottomrule
\end{tabular}
\end{table}

\newpage
\section{Theoretical Background} \label{sec:theory-appendix}
\newcommand{\refE}[1]{(\ref{#1})}

We denote the full dataset as $D=(D_1, \dots, D_{t+1})$, where $D_i=(X_i, Y_i)$ is the $i$-th datapoint.
Since we are in the split conformal setting \citep{vovk2005algorithmic}, the full dataset is subdivided into two parts: the calibration and the test data.
The prediction model $\muh$ is already given in this setting and not selected based on $D$.
The prediction interval $\Ch_t^{\alpha}$ for split conformal prediction is given by


\begin{equation}\label{eq:def_splitCP}
\Ch_t^{\alpha}(X_{t+1}) = \muh(X_{t+1})\pm \quant_{1-\alpha}\left(\sum_{i=1}^t
  \tfrac{1}{t+1} \cdot \delta_{R_i} +
  \tfrac{1}{t+1}\cdot\delta_{+\infty}\right), 
\end{equation}
where $R_i =|Y_i - \muh(X_i)|$, $\delta_c$ is a point mass at $c$, and $\quant_{\tau}$ is the $\tau$-quantile of a distribution.
\citet{vovk2005algorithmic} show that this interval provides a \textit{marginal} coverage guarantee in case of \textit{exchangeable} data:

\begin{theorem}[Split Conformal Prediction \citep{vovk2005algorithmic}] \
\\ If the data points $(X_1,Y_1),\dots,(X_t,Y_t),(X_{t+1},Y_{t+1})$ are exchangeable then the split conformal prediction set defined in~\refE{eq:def_splitCP} satisfies 
\[
\PR{Y_{t+1}\in\Ch_t^{\alpha}(X_{t+1})} \geq 1-\alpha.
\]
\end{theorem}

This seminal result cannot be directly applied to our setting, because time series are often non-exchangeable (as we discuss in the introduction) and models often exhibit locally specific errors.

The remainder of this section is split into two parts.
First, we discuss the theoretical connections between the MHN retrieval and CP through the lens of \citet{barber2022conformal}.
Second, we adapt the theory of \citet{xu2021conformal} to be applicable to our regime-based setting.

For the MHN retrieval, we evoke the results of \citet{barber2022conformal}, who introduce the concept of \textit{non-exchangeable split conformal} prediction.
Instead of treating each calibration point equally, they assign an individual weight for each point, to account for non-exchangeability.
Formally, we denote the weight for the $i$-th point by $w_i \in [0,1]$, $i=1,\dots,t$; 
and the respective normalized weights are defined as
\begin{equation}\label{eq:normalized_weights}
\tilde{w}_i = \frac{w_i}{w_1+\dots+w_t+1}, \ i=1,\dots,t, 
\textnormal{ and } \tilde{w}_{t+1}= \frac{1}{w_1+\dots+w_t+1}.
\end{equation}

The prediction interval for non-exchangeable split conformal setting $\Ch_t^{\alpha}$ is given by
\begin{equation}\label{eq:def_splitCP_nonExchange}
\Ch_t^{\alpha}(X_{t+1}) = \muh(X_{t+1})\pm \quant_{1-\alpha} \left(\sum_{i=1}^t \tilde{w}_i \cdot \delta_{R_i} + \tilde{w}_{t+1}\cdot\delta_{+\infty}\right). 
\end{equation}


\citet{barber2022conformal} show that non-exchangeable split conformal prediction satisfies a similar bound as exchangeable split conformal prediction, which is extended by a term that accounts for a potential \textit{coverage gap} $\Delta \ \text{Cov}$.
Formally we define $ \Delta \ \text{Cov} = \alpha - \alpha^\star$, where $\alpha$ is the specified miscoverage and $\alpha^\star$ is the observed miscoverage.

To specify the coverage gap they use the \textit{total variation distance} $\dtv$, which is a distance metric between two probability distributions. 
It is defined as
\begin{equation}
    \dtv(P_1,P_2)=\sup_{A\in\mathcal{F}}|P_1(A)-P_2(A)|,
\end{equation}
where $P_1$ and $P_2$ are arbitrary distributions defined on the space $(\Omega, \mathcal{F})$.

Intuitively, $\dtv$ measures the largest distance between two distributions.
It is connected to the widely used Kullback--Leibler divergence $D_{\mathsf{KL}}$ through Pinsker's inequality \citep{pinsker1964information}:
\begin{equation}
    \dtv(P_1,P_2) \leq \sqrt{\frac{1}{2} D_{\mathsf{KL}}(P_1||P_2)}.
\end{equation}

Given this measure, \citet{barber2022conformal} bound the coverage of non-exchangeable split conformal prediction as follows:

\begin{theorem}[Non-exchangeable split conformal prediction \citep{barber2022conformal}]\label{thm:splitCP_nonExchage} \
\\ The non-exchangeable split conformal method defined in~\eqref{eq:def_splitCP_nonExchange} satisfies 
\[
\PR{Y_{t+1}\in\Ch^{\alpha}_t(X_{t+1})}\geq (1-\alpha) - \sum_{i=1}^t\tilde{w}_i \cdot
\dtv\big(R(D),R(D^i)\big),
\]
where $R(D) = (R_1, R_2, \dots, R_{t+1})$ denotes the sequence of absolute residuals on $D$ and $D^i$ denotes a sequence where the test point $D_{t+1}$ is swapped with the $i$-th calibration point.
\end{theorem}

The coverage gap is represented by the rightmost term.
Therefore,
\[
\Delta \ \text{Cov} \geq - \sum_{i=1}^t\tilde{w}_i \cdot \dtv(R(D),R(D^i)).
\]

\citet{barber2022conformal} prove this for the full-conformal setting, which also directly applies to the split conformal setting (as they note, too).
For the sake of completeness, we state the proof of Theorem~\ref{thm:splitCP_nonExchage} explicitly for split conformal prediction \citep[adapted from][]{barber2022conformal}:\\

By the definition of the non-exchangeable split conformal prediction interval \refE{eq:def_splitCP_nonExchange}, it follows that

\begin{equation}\label{eq:quantile_noncoverage_1}
    Y_{t+1}\not\in\Ch_t^{\alpha}(X_{t+1}) \ \iff \
    R_{t+1} > \quant_{1-\alpha}\left(\sum_{i=1}^t \tilde{w}_i \cdot
      \delta_{R_i} + \tilde{w}_{t+1}\cdot\delta_{+\infty}\right).
    \end{equation}

Next, we show that, for a given $K \in  \{1, \dots, t+1\}$, 
\begin{equation}\label{eq:quantile_noncoverage_2}
    \quant_{1-\alpha}\left(\sum_{i=1}^t \tilde{w}_i \cdot \delta_{R_i} +
      \tilde{w}_{t+1}\cdot\delta_{+\infty}\right) \geq
    \quant_{1-\alpha}\left(\sum_{i=1}^{t+1} \tilde{w}_i \cdot
      \delta_{(R(D^K))_i}\right),
\end{equation}

by considering that
\begin{equation}\label{eq:residual_vector}
    \big(R(D^K)\big)_i =
        \begin{cases}
        R_i, & \textnormal {if } i \not= K \textnormal{ and } i \not= t+1, \\ 
        R_{t+1}, & \textnormal {if } i = K, \\
        R_K, & \textnormal {if } i = t+1.
        \end{cases}
\end{equation}

In the case of $K = t + 1$, this holds trivially since both sides are equivalent.
In the case of $K \leq t$, we can rewrite the left-hand side of inequality~\refE{eq:quantile_noncoverage_2} as 

\begin{align*}
\sum_{i=1}^t \tilde{w}_i \cdot \delta_{R_i} +
\tilde{w}_{t+1}\cdot\delta_{+\infty}& \\
&= \sum_{i=1,\dots,t; i\neq K} \tilde{w}_i \cdot \delta_{R_i} +
\tilde{w}_K (\delta_{R_K} + \delta_{+\infty}) + (\tilde{w}_{t+1} -
\tilde{w}_K)\delta_{+\infty}.
\end{align*}

The right-hand side can be similarly represented as

\begin{align*}
\sum_{i=1}^{t+1} \tilde{w}_i \cdot \delta_{(R(D^K))_i} &=
\sum_{i=1,\dots,t; i\neq K} \tilde{w}_i \cdot \delta_{R_i} +
\tilde{w}_K  \delta_{R_{t+1}} + \tilde{w}_{t+1}\delta_{R_K} \\ 
&=\sum_{i=1,\dots,t; i\neq K} \tilde{w}_i \cdot \delta_{R_i} +
\tilde{w}_K (\delta_{R_K} + \delta_{R_{t+1}}) + (\tilde{w}_{t+1} - \tilde{w}_K)\delta_{R_K}.
\end{align*}

By the definition of the normalized weights \refE{eq:normalized_weights} and the assumption that $w_K \in [0,1]$, it holds that \smash{$\tilde{w}_{t+1}\geq \tilde{w}_K$}.
Applying this to the rewritten forms of the left- and right-hand sides, we can see that inequality~\refE{eq:quantile_noncoverage_2} holds.

Substituting the right hand side of \refE{eq:quantile_noncoverage_1} by using \refE{eq:quantile_noncoverage_2}, we see that

\[
    Y_{t+1}\not\in\Ch_t^{\alpha}(X_{t+1}) \ \implies \ 
    R_{t+1} > \quant_{1-\alpha}\left(\sum_{i=1}^{t+1} \tilde{w}_i \cdot
      \delta_{(R(D^K))_i}\right),
\] 

or equivalently by using \refE{eq:residual_vector}:

\begin{equation}\label{eq:quantile_noncoverage_3}
Y_{t+1}\not\in\Ch_t^{\alpha}(X_{t+1}) \ \implies \ 
(R(D^K))_K >  \quant_{1-\alpha}\left(\sum_{i=1}^{t+1}
  \tilde{w}_i \cdot \delta_{(R(D^K))_i}\right). 
\end{equation}

For the next step we use the definition of the set of so-called \textit{strange points} $\mathcal{S}$:
\begin{equation}\label{eq:strange} 
\mathcal{S}(\Bep) = \left\{i\in[t+1] \ : \ \epsilon_i > \quant_{1-\alpha}\left(\sum_{j=1}^{t+1}
    \tilde{w}_j \cdot\delta_{\epsilon_j}\right)\right\}. 
\end{equation}

A strange point $s \in \mathcal{S}$ therefore refers to a data point with residuals that are outside the prediction interval.
Given the definition of $\mathcal{S}$, we can directly see that

\begin{equation}\label{eq:S_r_alpha}
\sum_{i\in\mathcal{S}(\Bep)} \tilde{w}_i \leq \alpha \textnormal{ for all }
\Bep\in\dR^{t+1}, 
\end{equation}

Combining \refE{eq:quantile_noncoverage_3} and \refE{eq:strange} we see that non-coverage of $Y_{t+1}$ implies strangeness of point $K$:

\begin{equation}\label{eq:noncoverage-implies-strangeness}
Y_{t+1}\not\in\Ch_t^{\alpha}(X_{t+1}) \ \implies \ K
\in\mathcal{S}\big(R(D^K)\big). 
\end{equation}

Given this, we arrive at the final derivation of the coverage bound:

\begin{align}
\PR{K\in\mathcal{S}\big(R(D^K)\big)} 
\notag
&=\sum_{i=1}^{t+1}\PR{K=i\textnormal{ and }
i\in\mathcal{S}\big(R(D^i)\big)} \\
\notag
&=\sum_{i=1}^{t+1}\tilde{w}_i\cdot
\PR{i\in\mathcal{S}\big(R(D^i)\big)}\\ 
\notag
&\leq \sum_{i=1}^{t+1}\tilde{w}_i\cdot
\left(\PR{i\in\mathcal{S}\big(R(D)\big)} +
\dtv\big(R(D),R(D^i)\big)\right)\\ 
\notag
&=\EXP{\sum_{i\in\mathcal{S}(R(D))} \tilde{w}_i} + \sum_{i=1}^t\tilde{w}_i \cdot
\dtv\big(R(D),R(D^i)\big)\\
\notag
&\leq\alpha + \sum_{i=1}^t\tilde{w}_i\cdot\dtv\big(R(D),R(D^i)\big),
\notag
\end{align}

where we use \refE{eq:S_r_alpha} from the penultimate to the last line.
The first to the second line follows from $K \independent D^i$, which holds because $K \independent D$ and $D^i$ is a permutation of $D$ that does not depend on K.

\citet{barber2022conformal} show two additional variations of the coverage gap bound based on Theorem~\ref{thm:splitCP_nonExchage}.
We do not replicate them here. 
One notable observation of theirs, however, is that instead of using the total variation distance based on the residual vectors also the distance of the raw data vector can be used:

\begin{equation}
   \Delta \ \text{Cov} \geq - \sum_i \tilde{w}_i\cdot\dtv(D, D^i).
\end{equation}


This coverage bound suggests that low weights for time steps corresponding to high total variation distance ensure small coverage gaps.
However, lower weights weaken the estimation quality of the prediction intervals.
Hence, all relevant time steps should receive high weights, while all others should receive weights close to zero.

\ourmethod leverages these results by using the association weights $a_i$ from Equation~\ref{eq:hopfield-association} to reflect the weights $\tilde{w_i}$ of non-exchangeable split conformal prediction.
\ourmethod aims to assign high association weights to data points that are from the same error distribution as the current point, and low weights to data points in the opposite case.
By Theorem~\ref{thm:splitCP_nonExchage}, association weights that follow this pattern lead to a small coverage gap.

\paragraph{An asymptotic (conditional) coverage bound} If we assume clearly distinguished error distributions instead of gradually shifting ones, hard weight assignment, i.e., association weights that are either zero or one, are a meaningful choice.
Given this simplification we can directly link \ourmethod to the work of \citet{xu2021conformal}.

\citet{xu2021conformal} introduce a variant of (split) conformal prediction with asymmetric prediction intervals:

\begin{align}\label{def:xu_CI_form}
     \widehat{C}^{\alpha}_{t}(x_{t+1}):=\bigg[
 &\muh_{-(t+1)}(x_{t+1}) +  \quant_{\frac{\alpha}{2}}(\{\hat{\epsilon}_i\}_{i=1}^{t}),
 \muh_{-(t+1)}(x_{t+1}) + \quant_{1-\frac{\alpha}{2}}\big(\{\hat{\epsilon}_i\}_{i=1}^{t}\big)\bigg]
\end{align}

The approach taken by \citet{xu2021conformal} additionally adjusts the share of $\alpha$ between the upper and lower quantiles such that the width of the prediction interval is minimized.
In contrast, \ourmethod adopts a simpler strategy and uses $\frac{\alpha}{2}$ on both sides, which mitigates the need for excessive recalculation of the quantile function. 

Based on the introduced interval calculation procedure, they also provide bounds regarding its coverage properties, in particular about the possible coverage gap.
While \citet{xu2021conformal} base their assumption on error properties only in temporal proximity, we generalize this to proximity in the learned feature representation (which can include but is not limited to information about temporal proximity).
Based on this proximity we assume:

\begin{assumption}[Error regimes]\label{app:local-error-assumption}
Given $\{(\mathbf{Z}_i, y_i, \hat{y}_i)\}^{t+1}_{i=1}$ which represents the (encoded) dataset and the predictions $\hat{y}_i$ of $\muh$, there exists a partition $\mathcal{T} = \{T_i\}_{i \in I}$ of the time steps $\{1,\dots, t\}$, so that
\begin{enumerate}[itemsep=1mm]
    \item[(i)] $\forall i \in I: \{\epsilon_t = y_t - \hat{y}_t \, | \, t \in T_i\}  \sim  E_i$, 
    \item[(ii)] $\exists i \in I: \{ \epsilon_{t+1} \} \sim E_i$, 
    \item[(iii)] the subset $T_i$ corresponding to $t+1$  can be identified based on $\mathbf{Z}_{t+1}$,
    \item[(iv)] $\forall i,j \in I: i = j \vee E_i \neq E_j $,
\end{enumerate}
\end{assumption}

where each $E_i$ is an arbitrary error distribution.

Properties (i) and (ii) imply that any given time step $t+1$ can be associated to a subset $T_i$ where all errors follow a common error distribution $E_i$; property (iii) ensures that the partition can be found or approximated from a set of features. 
Property (iv) is not necessary for the discussion at hand but avoids degenerate solutions, since tight coverage bounds call for assigning all errors from the same error distribution to one large partition.
If the errors of the associated subset were sampled i.i.d.\ from $E_i$, we could use standard split conformal methods \citep{vovk2005algorithmic}, and the variation distance between permutations of the errors in properties (i) and (ii) is zero.
In practice, error noise can make it challenging to find and delineate a given partition. Thus, \ourmethod uses a soft-selection approach to approximate the discrete association of a partition and the current features.


To align \ourmethod with the results of \citet{xu2021conformal}, and thus make use of their proposed coverage bounds for asymmetric prediction intervals, we first introduce their decomposition of the error:
$\epsilon_t = \check{\epsilon}_t + (\mu(\mathbf{x}_t) - \muh(\mathbf{x}_t))$, where $f$ is the true generating function, $\muh$ the prediction model, and $\check{\epsilon}$ a noise term associated with the label.
The notation $\mu_{-t}(\mathbf{x})$ stems from \citet{xu2021conformal} and denotes an evaluation of $\mu$, selected without using $t$.

Next, we adopt their i.i.d.\ assumption about the noise with regard to our approach:

\begin{assumption}[Noise is approximately {\it i.i.d.}\ within an error regime; adaptation of \citet{xu2021conformal}, Assumption~1]\label{ass:error_iid}
Assume $\exists i \in I : \{\check{\epsilon}_t | t \in T_i\} \cup \{\check{\epsilon}_{t+1}\} \overset{\mathrm{iid}}{\sim} E_i$ and that the CDF $\check{F}_{i}$ of $E_i$ is Lipschitz continuous with constant $L_i>0$.
\end{assumption}

This i.i.d.\ assumption within each error regime only concerns the noise of the true generating function.
\citet{xu2021conformal} note that the generating process itself can exhibit arbitrary dependence and be highly non-stationary while still having i.i.d.\ noise.
The assumption is thus relatively weak.

\begin{assumption}[Estimation quality; adaptation of \citet{xu2021conformal}, Assumption~2]\label{ass:quality} 
There exist real sequences $\{\delta_{i,|T_i|}\}_{i,|T_i|\geq 1}$ such that 
\begin{enumerate}
    \item[(i)] $\forall i \in I: \frac{1}{|T_i|} \sum_{t \in T_i} (\muh_{-t}(\Bx_t)- \mu(\Bx_t))^2 \leq \delta_{i,|T_i|}^2$
    \item[(ii)] $\exists i \in I: |\muh_{-(t+1)}(\Bx_{t+1})-\mu(\Bx_{t+1})|\leq \delta_{i,|T_i|}$, where the $i$ for a given $t+1$ is the same $i$ as in assumption~\ref{ass:error_iid}.
\end{enumerate}
\end{assumption}

We want to bound the coverage of \refE{def:xu_CI_form}, which is based on the prediction error $\epsilon$.
However, Assumption~\ref{ass:error_iid} only considers the noise $\check{\epsilon}$ of the true generating function.
\citet{xu2021conformal} bind the difference between the CDF of the true generation function $\check{F_i}$, the \textit{empirical} analogue $\tilde{F_i}(x) := \frac{1}{|T_i|}\sum_{j \in T_i}\mathbf{1}\{\check{\epsilon}_j \leq x\}$, and the empirical CDF of the prediction error $F_i(x) := \frac{1}{|T_i|}\sum_{j \in T_i}\mathbf{1}\{\epsilon_j \leq x\} $.

In particular, they state two lemmas:

\begin{lemma}[Adaptation of \citet{xu2021conformal}, Lemma~1]\label{lem:regu_xu}
Under Assumption \ref{ass:error_iid}, for any subset size $|T_i|$, there is an event $A_T$ which occurs with probability at least $1- \sqrt{\log (16|T_i|)/|T_i|}$, such that conditioning on $A_T$,
\[
     \sup_{x} |\tilde{F}_i(x)-\check{F}_i(x)| \leq \sqrt{\log (16|T_i|)/|T_i|}.
\]
\end{lemma}

\begin{lemma}[Adaptation of \citet{xu2021conformal}, Lemma~2]\label{lem:consis_xu}
Under Assumptions \ref{ass:error_iid} and \ref{ass:quality}, we have
\[
    \sup_x |F_{i}(x)-\tilde{F}_{i}(x)|\leq (L_{i}+1)\delta_{i,|T_i|}^{2/3}+2\sup_x |\tilde{F}_{i}(x)-\check{F}_{i}(x)|.
\]
\end{lemma}

With this result we arrive at a bound for a conditional and a marginal coverage:

\begin{theorem}[Conditional coverage gap; adaptation of \cite{xu2021conformal}, Theorem 1]\label{thm:cond_cov_xu}
Given the prediction interval $\widehat{C}^{\alpha}_t$ defined in \refE{def:xu_CI_form}. Under Assumption \ref{ass:error_iid} and \ref{ass:quality}, for any subset size $|T_i|$, $\alpha \in (0,1)$, and $\beta \in [0,\alpha]$, we have:
\begin{align}
    \exists i \in I: & \big|\PR{Y_{t+1}\in \widehat{C}^{\alpha}_{t}(X_{t+1})|X_{t+1}=\Bx_{t+1}}-(1-\alpha)\big| \nonumber \\
    \leq & 12\sqrt{\log (16|T_i|)/|T_i|} +4 (L_i + 1) (\delta_{i,|T_i|}^{2/3}+\delta_{i,|T_i|}). \label{eq:main}
\end{align}
Furthermore, if $\{\delta_{i,|T_i|}\}_{|T_i|\geq 1}$ converges to zero, the upper bound in \eqref{eq:main} converges to 0 when $|T_i| \rightarrow \infty$, and thus the conditional coverage is asymptotically valid.
\end{theorem}

\begin{theorem}[Marginal coverage gap; adaptation of \cite{xu2021conformal}, Theorem 2]\label{thm:marg_cov_xu}
Given the prediction interval $\Ch^{\alpha}_t$ defined in \refE{def:xu_CI_form}. Under Assumption \ref{ass:error_iid} and \ref{ass:quality}, for any subset size $|T_i|$, $\alpha \in (0,1)$, and $\beta \in [0,\alpha]$, we have:
\begin{align}
    \exists i \in I: \big|\PR{Y_{t+1}\in \widehat{C}^{\alpha}_{t}(X_{t+1})}-(1-\alpha)\big| \nonumber \leq 12\sqrt{\log (16|T_i|)/|T_i|}+4(L_i + 1) (\delta_{i,|T_i|}^{2/3}+\delta_{i,|T_i|}).
\end{align}
\end{theorem}

The proofs of the above theorems follow from \citet{xu2021conformal}.
To apply our adaptions to their poofs, one has to select the error regime matching to the test time step $t+1$, i.e., select the right subset $T_i$ from the partitioning of time steps.
$T_i$ then corresponds to the overall calibration data in the original assumptions, lemmas, theorems, and the respective proofs.
This is valid since Assumption~\ref{ass:quality}-(i) encompasses all subsets and the existing subset in Assumption~\ref{ass:error_iid} and Assumption~\ref{ass:quality}-(ii) is, as noted, the same.

\citet{xu2021conformal} provide additional bounds for two variations of Assumption~\ref{ass:error_iid}, imposing only the requirement of strongly mixing noise or noise following a linear process.
These bounds apply also to our adaption but are omitted here for brevity.

\paragraph{Notes on Exchangeability, Stationarity, Seasonality, and Drifts} 
Comparing \ourmethod to standard CP we see that \ourmethod assumes exchangeability over the weighted time steps within regimes.
Hence, we require that the relationship between inputs and outputs remains stationary within regimes, which is weaker than requiring that the time series as such is stationary.
For example, it allows that time series values decrease or grow over time as long as these changes are reflected in the inputs of the MHN and correspond to a regime.
HopCPT softens the exchangeability requirements of standard CP in two ways:

\begin{enumerate}
    \item The CP mechanism assumes that exchangeability is given within a regime that is selected by the MHN association.
    \item Even if (1) is not the case, the memory of the MHN (and therefore the calibration data) is updated with each new observation. This memory update allows HopCPT to account for shifts — empirically, the benefit is visible in our experiments on the sap flow dataset, as we discuss briefly in Section 3.2 (lines 283-289) and in more detail in Appendix A.3 (Error Trend \& Figure 4).
\end{enumerate}

Regarding seasonality: HopCPT uses covariate features to describe the current time step. Hence, HopCPT implicitly considers seasonality by comparing a given time step with the time steps from the memory. Whenever a time step of a certain season is encountered, it focuses on time steps from the same season. In fact, our experiments already comprise both regular and irregular seasonality patterns. For example, the streamflow dataset contains yearly reoccurring patterns (spring, summer, fall, winter) as well as irregularly reoccurring similar flood events. In both cases, HopCPT can identify related time steps based on the covariates.

Regarding data drifts: an advantage of HopCPT is that the memory is updated with every new observation. Therefore, HopCPT can somewhat incorporate drifts in the target value as long as the covariates remain so that the learned representation is able to focus on the relevant observations. Empirically, the robustness against drifts is visible in our experiments on the sap flow dataset, as we discuss briefly in Section 3.2 (lines 283-289) and in more detail in Appendix A.3 (Error Trend \& Figure 4).

\newpage

\section{Dataset Details}\label{sec:dataset-details}
Datsets from four different domains were used in the experiments.
The following paragraphs summarize the details. 
Table~\ref{tab:dataset-summary} provides a quantitative overview.

\paragraph{Solar.} We conducted experiments on three solar radiation datasets: \solarSmall, \solarOneY, and \solarThreeY.
All three datasets are based on data from the US National Solar Radiation Database \citep[NSDB;][]{sengupta2018national}.
Besides solar radiation as the target variable, the datasets include 8 other environmental features.
\solarSmall is the same dataset as used by \citet{xu2021conformal, xu2022sequential} and focuses on data from 8 cities in California.
To show the scalability of \ourmethod we additionally generated two larger datasets \solarOneY and \solarThreeY, which include data from 50 cities from different parts of the US.
We selected the cities by ordering the areas provided by NSDB according to the population density and picked the top 50 under the condition that no city appears twice in the dataset.

\paragraph{Air quality.}
The datasets \airTen and \airTwenty are air quality measurements from 12 monitoring sites in Beijing, China \citep{zhang2017cautionary}.
The datasets provide two target variables (10PM and 2.5PM measurements) which we use separately in our experiments (the variables are used mutually exclusively in the datasets, e.g., when predicting 10PM we do not use 2.5PM as a feature). 
We encode the wind direction, given as a categorical variable, by mapping the north--south and the east--west components each to a scalar from $-1$ to $1$.

\paragraph{Sap flow.}
The \sapflux dataset is a subset of the Sapflux data project \citep{poyatos2021global}.
This dataset includes sap flow measurements which we use as a target variable, as well as a set of environmental variables.
The available features, the length of the measurements, and the sampling vary between the individual time series.
To get a set of comparable time series we processed the data as follows: (a) We removed all time series where not all of the 10 environmental features are available.
(b) If there was a single missing value between two existing ones, we filled the value with the value before (forward fill).
(c) We cut out all sequences without any missing target or feature values (after step b).
(d) From the resulting set of sequences we removed all that have less than 15,000 or more than 20,000 time steps.

\paragraph{Streamflow.}
For our experiments on the streamflow data, we use the Catchment Attributes and Meteorology for Large-sample Studies dataset \citep[CAMELS;][]{newman2015development,addor2017camels}.
It provides meteorological time series from different data products, corresponding streamflow measurements, and static catchment attributes for catchments across the continental United States.
We ran our experiments on the subset of 531 catchments that were used in previous hydrological benchmarking efforts \citep[e.g.,][]{newman2017benchmarking,kratzert2019regional,kratzert2020note,klotz2022uncertainty}.
Specifically, we trained the LSTM prediction model with the NeuralHydrology Python library \citep{kratzert2022nh} on precipitation, solar radiation, minimum and maximum temperature, and vapor pressure from the NLDAS \citep{xia2012continental}, Maurer \citep{maurer2002long}, and Daymet \citep{thornton1997generating} meteorological data products.
Further, the LSTM received 26 static catchment attributes at each time step that identify the catchment properties. Table~4 in \citet{kratzert2019regional} provides a full list of these attributes.
We trained the prediction model on data from the period Oct 1981 -- Sep 1990 (note that for some catchments, this period starts later due to missing data in the beginning), calibrated the uncertainty models on Oct 1990 -- Sep 1999, and tested them on the period Oct 1999 -- Sep 2008.

\begin{table*}[t]
\caption{Details of the evaluated datasets.}
\label{tab:dataset-summary}
\vskip 0.15in
\centering
\begin{tabular}{@{}lllllll@{}}
\toprule
             & \begin{tabular}[c]{@{}l@{}}Number of\\ Series\end{tabular} & \begin{tabular}[c]{@{}l@{}}Time Steps\\ per Series\end{tabular} & Period  & Sampling & \begin{tabular}[c]{@{}l@{}}Number of\\ Features\end{tabular} & Data Split [\%] \\ \midrule
\solarSmall  & 8                                                          & 2,000                                                            & 01--03 2018    & 60m      & 8                                                            & 60/15/25   \\ 
\solarOneY   & 50                                                         & 8,760                                                            & 2019    & 60m      & 8                                                            & 60/15/25   \\ 
\solarThreeY & 50                                                         & 26,304                                                           & 2018--20 & 60m      & 8                                                            & 34/33/33   \\ \midrule
Air Quality      & 12                                                         & 35,064                                                           & 2013--17 & 60m      & 11                                                           & 34/33/33   \\
(10PM) \\
Air Quality   & 12                                                         & 35,064                                                           & 2013--17 & 60m      & 11                                                           & 34/33/33   \\
(25PM) \\\midrule
\sapflux     & 24                                                         & \begin{tabular}[c]{@{}l@{}}15,000--\\ 20,000\end{tabular}          & 2008--16 & varying  & 10                                                           & 34/33/33   \\ \midrule
\hydro       & 531                                                        & 9,862                                                          & 1981--2008 & 24h      & 41                                                           & 34/33/33   \\ \bottomrule
\end{tabular}
\end{table*}

\section{\knn vs.\ Learned Representation}\label{sec:knn-comparison}
As mentioned in the main paper, we designed \ourmethod with large datasets in mind.
It is only in such a setting that the learned part of our approach can truly play to its strengths and take advantage of nuanced interrelationships in the data. 
\knn provides us with a natural ``fallback'' option for settings where not enough data is available to infer these relationships. 
Our comparisons in Table~\ref{tab:results-knn-solar} and Table~\ref{tab:results-knn-other} substantiate this argument: 
\knn provides competitive results for the small dataset \solarSmall (this can also be seen by contrasting the performance from Table~\ref{tab:results-knn-solar} to the respective results in Appendix~\ref{sec:appendix-full-tables}), but is outperformed by \ourmethod for the larger datasets.

\newpage

\section{Quantile of Sample vs.\ Quantile of Weighted ECDF}

\ourmethod constructs prediction intervals by calculating a quantile on the set of weight-sampled errors (see Equation~\ref{eq:myconfbounds}).
An alternative approach is to calculate the quantile over a weighted empirical CDF of the errors.
This approach would define $q(\tau, \BZ_{t+1})$ in Equation~\ref{eq:myconfbounds} as
\begin{equation}
    q(\tau, \BZ_{t+1}) = \quant_{\tau} \left(\sum_{i=1}^t a_{t+1, i} \delta_{\epsilon_{i}} \right),
\end{equation}

where $\delta_{\epsilon_{i}}$ is a point mass at $|\epsilon_i|$.
Empirically, we find little differences in the performance when comparing the two approaches (Tables~\ref{tab:results-cdf-solar} and~\ref{tab:results-cdf-other}).

\section{Asymmetrical Error Quantiles}
\ourmethod calculates the prediction interval by using equally large quantiles (in terms of distribution mass, not width) for the upper and lower bound (see Equation~\ref{eq:myconfbounds}).
\citet{xu2021conformal} instead search for a $\hat{\beta} \in [0,\alpha]$ and calculate the lower quantile with $\hat{\beta}$ and the upper quantile with $1-\alpha + \hat{\beta}$, which minimizes the overall interval with.
Applying this principle to Equation~\ref{eq:myconfbounds} one would arrive at
\begin{equation}\label{eq:myconfbounds-with-beta}
\begin{aligned}
\hat{\beta} = \argmin_{\beta \in [0,\alpha]} \Bigl[ & \muh(\BX_{t+1}) + q(\beta, \BZ_{t+1}), \muh(\BX_{t+1}) + q(1-\alpha + \beta, \BZ_{t+1}) \Bigr] \\
\Ch_t^{\alpha}(\BZ_{t+1}) = %
   \Bigl[ & \muh(\BX_{t+1}) + q(\hat{\beta}, \BZ_{t+1}), \muh(\BX_{t+1}) + q(1-\alpha + \hat{\beta}, \BZ_{t+1}) \Bigr],
\end{aligned}
\end{equation}
Practically, the search is done by evaluating different $\beta$ values within the potential value range and selecting the one which produces the lowest interval width. 
Theoretically, this could lead to more efficient intervals at the cost of additional computation and potential asymmetry in regard to uncovered samples that are higher and lower than the prediction interval.
Empirically, we do not find a notable improvement when applying this search to \ourmethod (see Table~\ref{tab:results-betasearch-main}).

\section{AdaptiveCI and \ourmethod}\label{sec:adaptive-cp}

As noted in Section~\ref{sec:experiments}, \adaptiveci is orthogonal to \ourmethod, \spic, \enbpi, and \nexcp.
We therefore also evaluated the combined application of the models.
Nevertheless, AdaptiveCI is an independent model on top of an existing model, which is reflected in the way we select the hyperparameters of the combined model:
First, the hyperparameters are selected for each model without adaption through AdaptiveCI (see Appendix~\ref{sec:hyperparam}) --- hence, we use the same hyperparameters as in the main evaluation.
Second, we conduct another hyperparameter search, given the model parameters from the first search, where we only search for the parameter of the adaptive component.

Tables~\ref{tab:results-adaptive-solar} and~\ref{tab:results-adaptive-other} show the results of these experiments.
Overall, the results are slightly better, as the Winkler score (which considers both width and coverage) slightly improves in  most experiments.
The ranking between the different models stays similar to the non-adaptive comparison (see Section~\ref{sec:results-discuss}) with \ourmethod performing best on all but the smallest dataset.

\section{Details on Continuous Modern Hopfield Networks} \label{sec:appendix-hopfield}
Modern Hopfield Networks are associative memory networks.
They learn to associate entries of a memory with a given ``query'' sample in order to produce a prediction.
In \ourmethod, the memory corresponds to representations of past time steps and the query corresponds to a representation of the current time step.
The association happens by means of weights on every memory entry, which are called ``association weights''.
Query--memory sample pairs that have similar representations are assigned high association weights.
In summary, the association weights indicate which samples are closely related --- in \ourmethod, this corresponds to finding the time steps which are part of the same time series regimes and therefore useful to estimate the current uncertainty.
The association mechanism is in that sense closely related to the attention mechanism of Transformers \citep{Vaswani:17}.
However, compared to other trainable networks, the memory-based architecture has the advantage that it does not need to encode all knowledge in network parameters but can dynamically adapt when given new memory entries. 

We adapt the following arguments from \citet{furst2021cloob} and \citet{Ramsauer:21} for a more formal overview.
Associative memory networks have been designed to store and retrieve samples. 
Hopfield networks are energy-based, binary associative memories, which were popularized as artificial neural network architectures in the 1980s \citep{Hopfield:82,Hopfield:84}.
Their storage capacity can be considerably increased by polynomial terms in the energy function \citep{Chen:86,Psaltis:86,Baldi:87,Gardner:87,Abbott:87,Horn:88,Caputo:02,Krotov:16}.
In contrast to these binary memory networks, we use continuous associative memory networks with far higher storage capacity. 
These networks are continuous and differentiable, retrieve with a single update, and have exponential storage capacity \citep[and are therefore scalable, i.e., able tackle large problems;][]{Ramsauer:21}.

Formally, we denote a set of patterns $\{\Bx_1,\ldots,\Bx_N\} \subset \dR^d$
that are stacked as columns to 
the matrix $\BX = \left( \Bx_1,\ldots,\Bx_N \right)$ and a 
state pattern (query) $\Bxi \in \dR^d$ that represents the current state. 
The largest norm of a stored pattern is
$M = \max_{i} \NRM{\Bx_i}$.
Then, the energy $\rE$ of continuous Modern Hopfield Networks with state $\Bxi$ is defined as \citep{Ramsauer:21}

\begin{equation}
\label{eq:energy}
\rE  \ =   \ -  \ \beta^{-1} \ \log \left( \sum_{i=1}^N
\exp(\beta \Bx_i^T \Bxi) \right) +  \ 
\frac{1}{2} \ \Bxi^T \Bxi  \ +  \ \rC,
\end{equation}

where $\rC=\beta^{-1} \log N  \ + \
\frac{1}{2} \ M^2$. For energy $\rE$ and state $\Bxi$, \citet{Ramsauer:21} proved that the update rule 
\begin{equation}
\label{eq:main_update}
\Bxi^{\nn} \ = \  \BX \ \soft ( \beta \BX^T \Bxi)
\end{equation}

converges globally to stationary points of the energy $\rE$ and coincides with the attention mechanisms of Transformers 
\citep{Vaswani:17,Ramsauer:21}.

The {\em separation} $\Delta_i$ of a pattern $\Bx_i$ is its minimal dot product difference to any of the other 
patterns:
\begin{equation}
    \Delta_i = \min_{j,j \not= i} \left( \Bx_i^T \Bx_i - \Bx_i^T \Bx_j \right).
\end{equation}
A pattern is {\em well-separated} from the data if $\Delta_i $ is above a given threshold \citep[specified in][]{Ramsauer:21}.
If the patterns $\Bx_i$ are well-separated, the update rule Equation~\ref{eq:main_update}
converges to a fixed point close to a stored pattern.
If some patterns are similar to one another and, therefore, not well-separated, 
the update rule converges to a fixed point close to the mean of the similar patterns. 

The update rule of a Hopfield network thus identifies sample--sample relations between stored patterns. This enables similarity-based learning methods like nearest neighbor search \citep[see][]{schafl2022hopular}, which \ourmethod leverages to learn a retrieval of samples from similar error regimes.

\section{Potential Social Impact}\label{sec:social-impact}
Reliable uncertainty estimates are crucial, especially for complex time-dependent environmental phenomena.
However, overreliance on these estimates can be dangerous.
For example, unseen regimes might not be properly predicted.
A changing climate that evolves the environment beyond already seen conditions can cause new forms of error regimes which cannot be predicted reliably.
As most machine learning approaches, our method requires accurately labeled training data.
Incorrect labels may lead to unexpected biases and prediction errors.

\section{Code and Data}\label{sec:code}
The code and data to reproduce all of our experiments are available at \repourl.

\newpage

\setTableCaption{Performance of \knn compared to \ourmethod for the miscoverage $\alpha = 0.10$ on the solar datasets. The error term represents the standard deviation over repeated runs.}


\begin{table}
\centering
\caption{\nextTableCaption}
\label{tab:results-knn-solar}
\vskip 0.15in
\small
\begin{tabular}{lllrr}
\toprule
 &  & UC & \knn & \ourmethod \\
Data & FC &  &  &  \\
\midrule
\multirow[c]{12}{*}{\rotatebox{90}{Solar 3Y}} & \multirow[c]{3}{*}{\rotatebox{90}{Forest}} & $\Delta$ Cov & $0.015$ & $0.029^{\pm 0.012}$ \\
\rotatebox{90}{} & \rotatebox{90}{} & PI-Width & $97.2$ & $\textbf{39.0}^{\pm 6.2}$ \\
\rotatebox{90}{} & \rotatebox{90}{} & Winkler & $1.59$ & $\textbf{0.73}^{\pm 0.20}$ \\
\cmidrule{2-5}
\rotatebox{90}{} & \multirow[c]{3}{*}{\rotatebox{90}{LGBM}} & $\Delta$ Cov & $0.012$ & $0.001^{\pm 0.003}$ \\
\rotatebox{90}{} & \rotatebox{90}{} & PI-Width & $108.4$ & $\textbf{37.7}^{\pm 0.7}$ \\
\rotatebox{90}{} & \rotatebox{90}{} & Winkler & $1.74$ & $\textbf{0.57}^{\pm 0.01}$ \\
\cmidrule{2-5}
\rotatebox{90}{} & \multirow[c]{3}{*}{\rotatebox{90}{Ridge}} & $\Delta$ Cov & $-0.009$ & $0.040^{\pm 0.001}$ \\
\rotatebox{90}{} & \rotatebox{90}{} & PI-Width & $136.9$ & $\textbf{44.9}^{\pm 0.5}$ \\
\rotatebox{90}{} & \rotatebox{90}{} & Winkler & $1.99$ & $\textbf{0.64}^{\pm 0.00}$ \\
\cmidrule{2-5}
\rotatebox{90}{} & \multirow[c]{3}{*}{\rotatebox{90}{LSTM}} & $\Delta$ Cov & $-0.002$ & $0.001^{\pm 0.006}$ \\
\rotatebox{90}{} & \rotatebox{90}{} & PI-Width & $22.9$ & $\textbf{17.9}^{\pm 0.6}$ \\
\rotatebox{90}{} & \rotatebox{90}{} & Winkler & $0.51$ & $\textbf{0.30}^{\pm 0.01}$ \\
\cmidrule{1-5}
\multirow[c]{12}{*}{\rotatebox{90}{Solar 1Y}} & \multirow[c]{3}{*}{\rotatebox{90}{Forest}} & $\Delta$ Cov & $0.026$ & $0.047^{\pm 0.004}$ \\
\rotatebox{90}{} & \rotatebox{90}{} & PI-Width & $66.9$ & $\textbf{28.6}^{\pm 1.0}$ \\
\rotatebox{90}{} & \rotatebox{90}{} & Winkler & $1.00$ & $\textbf{0.40}^{\pm 0.04}$ \\
\cmidrule{2-5}
\rotatebox{90}{} & \multirow[c]{3}{*}{\rotatebox{90}{LGBM}} & $\Delta$ Cov & $0.003$ & $-0.003^{\pm 0.009}$ \\
\rotatebox{90}{} & \rotatebox{90}{} & PI-Width & $72.2$ & $\textbf{40.7}^{\pm 2.8}$ \\
\rotatebox{90}{} & \rotatebox{90}{} & Winkler & $1.08$ & $\textbf{0.57}^{\pm 0.08}$ \\
\cmidrule{2-5}
\rotatebox{90}{} & \multirow[c]{3}{*}{\rotatebox{90}{Ridge}} & $\Delta$ Cov & $-0.007$ & $-0.011^{\pm 0.016}$ \\
\rotatebox{90}{} & \rotatebox{90}{} & PI-Width & $128.8$ & $\textbf{59.9}^{\pm 5.6}$ \\
\rotatebox{90}{} & \rotatebox{90}{} & Winkler & $1.63$ & $\textbf{0.92}^{\pm 0.12}$ \\
\cmidrule{2-5}
\rotatebox{90}{} & \multirow[c]{3}{*}{\rotatebox{90}{LSTM}} & $\Delta$ Cov & $0.000$ & $0.028^{\pm 0.010}$ \\
\rotatebox{90}{} & \rotatebox{90}{} & PI-Width & $16.1$ & $\textbf{16.0}^{\pm 0.6}$ \\
\rotatebox{90}{} & \rotatebox{90}{} & Winkler & $0.33$ & $\textbf{0.22}^{\pm 0.01}$ \\
\cmidrule{1-5}
\multirow[c]{9}{*}{\rotatebox{90}{Solar Small}} & \multirow[c]{3}{*}{\rotatebox{90}{Forest}} & $\Delta$ Cov & $0.034$ & $0.008^{\pm 0.006}$ \\
\rotatebox{90}{} & \rotatebox{90}{} & PI-Width & $\textbf{35.0}$ & $38.4^{\pm 3.4}$ \\
\rotatebox{90}{} & \rotatebox{90}{} & Winkler & $\textbf{0.93}$ & $1.09^{\pm 0.08}$ \\
\cmidrule{2-5}
\rotatebox{90}{} & \multirow[c]{3}{*}{\rotatebox{90}{LGBM}} & $\Delta$ Cov & $-0.022$ & $0.007^{\pm 0.012}$ \\
\rotatebox{90}{} & \rotatebox{90}{} & PI-Width & $\textbf{47.2}$ & $48.2^{\pm 1.7}$ \\
\rotatebox{90}{} & \rotatebox{90}{} & Winkler & $\textbf{1.17}$ & $1.24^{\pm 0.08}$ \\
\cmidrule{2-5}
\rotatebox{90}{} & \multirow[c]{3}{*}{\rotatebox{90}{Ridge}} & $\Delta$ Cov & $0.004$ & $-0.016^{\pm 0.022}$ \\
\rotatebox{90}{} & \rotatebox{90}{} & PI-Width & $\textbf{49.1}$ & $78.2^{\pm 24.5}$ \\
\rotatebox{90}{} & \rotatebox{90}{} & Winkler & $\textbf{1.08}$ & $1.78^{\pm 0.59}$ \\
\bottomrule
\end{tabular}
\end{table}

\setTableCaption{Performance of \knn compared to \ourmethod for the miscoverage $\alpha = 0.10$ on the different non-solar datasets. The error term represents the standard deviation over repeated runs.}


\begin{table}
\centering
\caption{\nextTableCaption}
\label{tab:results-knn-other}
\vskip 0.15in
\small
\begin{tabular}{lllrr}
\toprule
 &  & UC & \knn & \ourmethod \\
Data & FC &  &  &  \\
\midrule
\multirow[c]{12}{*}{\rotatebox{90}{Air 10 PM}} & \multirow[c]{3}{*}{\rotatebox{90}{Forest}} & $\Delta$ Cov & $-0.071$ & $0.028^{\pm 0.019}$ \\
\rotatebox{90}{} & \rotatebox{90}{} & PI-Width & $186.3$ & $\textbf{93.9}^{\pm 11.1}$ \\
\rotatebox{90}{} & \rotatebox{90}{} & Winkler & $3.81$ & $\textbf{1.50}^{\pm 0.09}$ \\
\cmidrule{2-5}
\rotatebox{90}{} & \multirow[c]{3}{*}{\rotatebox{90}{LGBM}} & $\Delta$ Cov & $-0.064$ & $0.017^{\pm 0.016}$ \\
\rotatebox{90}{} & \rotatebox{90}{} & PI-Width & $164.6$ & $\textbf{85.6}^{\pm 7.4}$ \\
\rotatebox{90}{} & \rotatebox{90}{} & Winkler & $3.42$ & $\textbf{1.45}^{\pm 0.06}$ \\
\cmidrule{2-5}
\rotatebox{90}{} & \multirow[c]{3}{*}{\rotatebox{90}{Ridge}} & $\Delta$ Cov & $-0.054$ & $0.010^{\pm 0.007}$ \\
\rotatebox{90}{} & \rotatebox{90}{} & PI-Width & $114.8$ & $\textbf{79.9}^{\pm 4.7}$ \\
\rotatebox{90}{} & \rotatebox{90}{} & Winkler & $2.39$ & $\textbf{1.35}^{\pm 0.06}$ \\
\cmidrule{2-5}
\rotatebox{90}{} & \multirow[c]{3}{*}{\rotatebox{90}{LSTM}} & $\Delta$ Cov & $-0.031$ & $-0.002^{\pm 0.005}$ \\
\rotatebox{90}{} & \rotatebox{90}{} & PI-Width & $\textbf{57.2}$ & $62.7^{\pm 1.5}$ \\
\rotatebox{90}{} & \rotatebox{90}{} & Winkler & $\textbf{1.30}$ & $1.33^{\pm 0.01}$ \\
\cmidrule{1-5}
\multirow[c]{12}{*}{\rotatebox{90}{Air 25 PM}} & \multirow[c]{3}{*}{\rotatebox{90}{Forest}} & $\Delta$ Cov & $-0.081$ & $-0.024^{\pm 0.017}$ \\
\rotatebox{90}{} & \rotatebox{90}{} & PI-Width & $158.4$ & $\textbf{48.1}^{\pm 5.6}$ \\
\rotatebox{90}{} & \rotatebox{90}{} & Winkler & $3.77$ & $\textbf{1.12}^{\pm 0.05}$ \\
\cmidrule{2-5}
\rotatebox{90}{} & \multirow[c]{3}{*}{\rotatebox{90}{LGBM}} & $\Delta$ Cov & $-0.073$ & $-0.021^{\pm 0.019}$ \\
\rotatebox{90}{} & \rotatebox{90}{} & PI-Width & $130.4$ & $\textbf{46.8}^{\pm 6.0}$ \\
\rotatebox{90}{} & \rotatebox{90}{} & Winkler & $3.10$ & $\textbf{1.10}^{\pm 0.05}$ \\
\cmidrule{2-5}
\rotatebox{90}{} & \multirow[c]{3}{*}{\rotatebox{90}{Ridge}} & $\Delta$ Cov & $-0.059$ & $-0.005^{\pm 0.026}$ \\
\rotatebox{90}{} & \rotatebox{90}{} & PI-Width & $97.5$ & $\textbf{49.3}^{\pm 10.6}$ \\
\rotatebox{90}{} & \rotatebox{90}{} & Winkler & $2.30$ & $\textbf{1.04}^{\pm 0.11}$ \\
\cmidrule{2-5}
\rotatebox{90}{} & \multirow[c]{3}{*}{\rotatebox{90}{LSTM}} & $\Delta$ Cov & $-0.033$ & $0.007^{\pm 0.008}$ \\
\rotatebox{90}{} & \rotatebox{90}{} & PI-Width & $\textbf{34.7}$ & $40.7^{\pm 4.3}$ \\
\rotatebox{90}{} & \rotatebox{90}{} & Winkler & $0.94$ & $\textbf{0.88}^{\pm 0.05}$ \\
\cmidrule{1-5}
\multirow[c]{12}{*}{\rotatebox{90}{Sap flow}} & \multirow[c]{3}{*}{\rotatebox{90}{Forest}} & $\Delta$ Cov & $-0.056$ & $0.009^{\pm 0.019}$ \\
\rotatebox{90}{} & \rotatebox{90}{} & PI-Width & $3246.1$ & $\textbf{1078.7}^{\pm 73.7}$ \\
\rotatebox{90}{} & \rotatebox{90}{} & Winkler & $1.00$ & $\textbf{0.30}^{\pm 0.01}$ \\
\cmidrule{2-5}
\rotatebox{90}{} & \multirow[c]{3}{*}{\rotatebox{90}{LGBM}} & $\Delta$ Cov & $-0.063$ & $-0.016^{\pm 0.013}$ \\
\rotatebox{90}{} & \rotatebox{90}{} & PI-Width & $2647.5$ & $\textbf{875.5}^{\pm 49.8}$ \\
\rotatebox{90}{} & \rotatebox{90}{} & Winkler & $0.88$ & $\textbf{0.27}^{\pm 0.00}$ \\
\cmidrule{2-5}
\rotatebox{90}{} & \multirow[c]{3}{*}{\rotatebox{90}{Ridge}} & $\Delta$ Cov & $-0.052$ & $-0.025^{\pm 0.023}$ \\
\rotatebox{90}{} & \rotatebox{90}{} & PI-Width & $2808.4$ & $\textbf{1639.6}^{\pm 93.2}$ \\
\rotatebox{90}{} & \rotatebox{90}{} & Winkler & $0.82$ & $\textbf{0.46}^{\pm 0.05}$ \\
\cmidrule{2-5}
\rotatebox{90}{} & \multirow[c]{3}{*}{\rotatebox{90}{LSTM}} & $\Delta$ Cov & $-0.026$ & $0.004^{\pm 0.004}$ \\
\rotatebox{90}{} & \rotatebox{90}{} & PI-Width & $719.5$ & $\textbf{594.3}^{\pm 7.7}$ \\
\rotatebox{90}{} & \rotatebox{90}{} & Winkler & $0.24$ & $\textbf{0.19}^{\pm 0.01}$ \\
\cmidrule{1-5}
\multirow[c]{3}{*}{Streamflow} & \multirow[c]{3}{*}{LSTM} & $\Delta$ Cov & $-0.108$ & $0.001^{\pm 0.041}$ \\
 &  & PI-Width & $1.70$ & $\textbf{1.39}^{\pm 0.17}$ \\
 &  & Winkler & $1.39$ & $\textbf{0.79}^{\pm 0.03}$ \\
\bottomrule
\end{tabular}
\end{table}

\setTableCaption{Performance of the weighted sample quantile (Sample) and the weighted empirical CDF (ECDF) quantile strategies that \ourmethod uses for the miscoverage $\alpha = 0.10$ on the solar datasets. The error term represents the standard deviation over repeated runs.}


\begin{table}
\centering
\caption{\nextTableCaption}
\label{tab:results-cdf-solar}
\vskip 0.15in
\small
\begin{tabular}{lllrr}
\toprule
 &  & UC & \text{Sample} & \text{ECDF} \\
Data & FC &  &  &  \\
\midrule
\multirow[c]{12}{*}{\rotatebox{90}{Solar 3Y}} & \multirow[c]{3}{*}{\rotatebox{90}{Forest}} & $\Delta$ Cov & $0.029^{\pm 0.012}$ & $0.021^{\pm 0.008}$ \\
\rotatebox{90}{} & \rotatebox{90}{} & PI-Width & $39.0^{\pm 6.2}$ & $\textbf{33.1}^{\pm 1.0}$ \\
\rotatebox{90}{} & \rotatebox{90}{} & Winkler & $0.73^{\pm 0.20}$ & $\textbf{0.62}^{\pm 0.03}$ \\
\cmidrule{2-5}
\rotatebox{90}{} & \multirow[c]{3}{*}{\rotatebox{90}{LGBM}} & $\Delta$ Cov & $0.001^{\pm 0.003}$ & $-0.006^{\pm 0.006}$ \\
\rotatebox{90}{} & \rotatebox{90}{} & PI-Width & $37.7^{\pm 0.7}$ & $\textbf{37.1}^{\pm 0.9}$ \\
\rotatebox{90}{} & \rotatebox{90}{} & Winkler & $\textbf{0.57}^{\pm 0.01}$ & $0.58^{\pm 0.01}$ \\
\cmidrule{2-5}
\rotatebox{90}{} & \multirow[c]{3}{*}{\rotatebox{90}{Ridge}} & $\Delta$ Cov & $0.040^{\pm 0.001}$ & $0.041^{\pm 0.001}$ \\
\rotatebox{90}{} & \rotatebox{90}{} & PI-Width & $\textbf{44.9}^{\pm 0.5}$ & $45.0^{\pm 0.5}$ \\
\rotatebox{90}{} & \rotatebox{90}{} & Winkler & $\textbf{0.64}^{\pm 0.00}$ & $\textbf{0.64}^{\pm 0.00}$ \\
\cmidrule{2-5}
\rotatebox{90}{} & \multirow[c]{3}{*}{\rotatebox{90}{LSTM}} & $\Delta$ Cov & $0.001^{\pm 0.006}$ & $0.001^{\pm 0.006}$ \\
\rotatebox{90}{} & \rotatebox{90}{} & PI-Width & $17.9^{\pm 0.6}$ & $\textbf{17.8}^{\pm 0.4}$ \\
\rotatebox{90}{} & \rotatebox{90}{} & Winkler & $\textbf{0.30}^{\pm 0.01}$ & $\textbf{0.30}^{\pm 0.00}$ \\
\cmidrule{1-5}
\multirow[c]{12}{*}{\rotatebox{90}{Solar 1Y}} & \multirow[c]{3}{*}{\rotatebox{90}{Forest}} & $\Delta$ Cov & $0.047^{\pm 0.004}$ & $0.048^{\pm 0.005}$ \\
\rotatebox{90}{} & \rotatebox{90}{} & PI-Width & $\textbf{28.6}^{\pm 1.0}$ & $28.9^{\pm 1.0}$ \\
\rotatebox{90}{} & \rotatebox{90}{} & Winkler & $\textbf{0.40}^{\pm 0.04}$ & $\textbf{0.40}^{\pm 0.04}$ \\
\cmidrule{2-5}
\rotatebox{90}{} & \multirow[c]{3}{*}{\rotatebox{90}{LGBM}} & $\Delta$ Cov & $-0.003^{\pm 0.009}$ & $0.000^{\pm 0.007}$ \\
\rotatebox{90}{} & \rotatebox{90}{} & PI-Width & $40.7^{\pm 2.8}$ & $\textbf{40.5}^{\pm 2.9}$ \\
\rotatebox{90}{} & \rotatebox{90}{} & Winkler & $0.57^{\pm 0.08}$ & $\textbf{0.56}^{\pm 0.08}$ \\
\cmidrule{2-5}
\rotatebox{90}{} & \multirow[c]{3}{*}{\rotatebox{90}{Ridge}} & $\Delta$ Cov & $-0.011^{\pm 0.016}$ & $-0.006^{\pm 0.013}$ \\
\rotatebox{90}{} & \rotatebox{90}{} & PI-Width & $59.9^{\pm 5.6}$ & $\textbf{57.5}^{\pm 2.5}$ \\
\rotatebox{90}{} & \rotatebox{90}{} & Winkler & $0.92^{\pm 0.12}$ & $\textbf{0.88}^{\pm 0.10}$ \\
\cmidrule{2-5}
\rotatebox{90}{} & \multirow[c]{3}{*}{\rotatebox{90}{LSTM}} & $\Delta$ Cov & $0.028^{\pm 0.010}$ & $0.032^{\pm 0.007}$ \\
\rotatebox{90}{} & \rotatebox{90}{} & PI-Width & $\textbf{16.0}^{\pm 0.6}$ & $16.6^{\pm 1.1}$ \\
\rotatebox{90}{} & \rotatebox{90}{} & Winkler & $\textbf{0.22}^{\pm 0.01}$ & $\textbf{0.22}^{\pm 0.01}$ \\
\cmidrule{1-5}
\multirow[c]{9}{*}{\rotatebox{90}{Solar Small}} & \multirow[c]{3}{*}{\rotatebox{90}{Forest}} & $\Delta$ Cov & $0.008^{\pm 0.006}$ & $0.026^{\pm 0.023}$ \\
\rotatebox{90}{} & \rotatebox{90}{} & PI-Width & $38.4^{\pm 3.4}$ & $\textbf{37.1}^{\pm 5.7}$ \\
\rotatebox{90}{} & \rotatebox{90}{} & Winkler & $1.09^{\pm 0.08}$ & $\textbf{0.99}^{\pm 0.06}$ \\
\cmidrule{2-5}
\rotatebox{90}{} & \multirow[c]{3}{*}{\rotatebox{90}{LGBM}} & $\Delta$ Cov & $0.007^{\pm 0.012}$ & $0.011^{\pm 0.013}$ \\
\rotatebox{90}{} & \rotatebox{90}{} & PI-Width & $48.2^{\pm 1.7}$ & $\textbf{46.3}^{\pm 1.3}$ \\
\rotatebox{90}{} & \rotatebox{90}{} & Winkler & $1.24^{\pm 0.08}$ & $\textbf{1.16}^{\pm 0.04}$ \\
\cmidrule{2-5}
\rotatebox{90}{} & \multirow[c]{3}{*}{\rotatebox{90}{Ridge}} & $\Delta$ Cov & $-0.016^{\pm 0.022}$ & $0.012^{\pm 0.011}$ \\
\rotatebox{90}{} & \rotatebox{90}{} & PI-Width & $\textbf{78.2}^{\pm 24.5}$ & $109.0^{\pm 18.6}$ \\
\rotatebox{90}{} & \rotatebox{90}{} & Winkler & $\textbf{1.78}^{\pm 0.59}$ & $2.55^{\pm 0.38}$ \\
\bottomrule
\end{tabular}
\end{table}

\setTableCaption{Comparison of \ourmethod \ (a) with applying the approach proposed by \citep{xu2021conformal, xu2022sequential} to search for a $\beta$ that minimizes the prediction interval width (right column), and (b) without this approach which corresponds to $\beta = \alpha /2$ (left column). The experiments are done with a miscoverage rate $\alpha=0.1$. The error term represents the standard deviation over repeated runs.}
\begin{table}
\centering
\caption{\nextTableCaption}
\label{tab:results-betasearch-main}
\vskip 0.15in
\small
\begin{tabular}{lllll}
\toprule
 &  & UC & $\beta = \alpha / 2$ & With $\beta$ search  \\
Data & FC &  &  &  \\
\midrule
\multirow[c]{12}{*}{\rotatebox{90}{Solar 3Y}} & \multirow[c]{3}{*}{\rotatebox{90}{Forest}} & $\Delta$ Cov & $0.029^{\pm 0.012}$ & $0.010^{\pm 0.018}$ \\
\rotatebox{90}{} & \rotatebox{90}{} & PI-Width & $\textbf{39.0}^{\pm 6.2}$ & $41.0^{\pm 7.6}$ \\
\rotatebox{90}{} & \rotatebox{90}{} & Winkler & $\textbf{0.73}^{\pm 0.20}$ & $0.86^{\pm 0.22}$ \\
\cmidrule{2-5}
\rotatebox{90}{} & \multirow[c]{3}{*}{\rotatebox{90}{LGBM}} & $\Delta$ Cov & $0.001^{\pm 0.003}$ & $-0.013^{\pm 0.003}$ \\
\rotatebox{90}{} & \rotatebox{90}{} & PI-Width & $37.7^{\pm 0.7}$ & $\textbf{36.8}^{\pm 6.2}$ \\
\rotatebox{90}{} & \rotatebox{90}{} & Winkler & $\textbf{0.57}^{\pm 0.01}$ & $0.62^{\pm 0.12}$ \\
\cmidrule{2-5}
\rotatebox{90}{} & \multirow[c]{3}{*}{\rotatebox{90}{Ridge}} & $\Delta$ Cov & $0.040^{\pm 0.001}$ & $0.021^{\pm 0.015}$ \\
\rotatebox{90}{} & \rotatebox{90}{} & PI-Width & $\textbf{44.9}^{\pm 0.5}$ & $50.0^{\pm 28.3}$ \\
\rotatebox{90}{} & \rotatebox{90}{} & Winkler & $\textbf{0.64}^{\pm 0.00}$ & $0.77^{\pm 0.43}$ \\
\cmidrule{2-5}
\rotatebox{90}{} & \multirow[c]{3}{*}{\rotatebox{90}{LSTM}} & $\Delta$ Cov & $0.001^{\pm 0.006}$ & $-0.013^{\pm 0.007}$ \\
\rotatebox{90}{} & \rotatebox{90}{} & PI-Width & $17.9^{\pm 0.6}$ & $\textbf{16.2}^{\pm 0.5}$ \\
\rotatebox{90}{} & \rotatebox{90}{} & Winkler & $\textbf{0.30}^{\pm 0.01}$ & $0.31^{\pm 0.00}$ \\
\cmidrule{1-5}
\multirow[c]{12}{*}{\rotatebox{90}{Air 10 PM}} & \multirow[c]{3}{*}{\rotatebox{90}{Forest}} & $\Delta$ Cov & $0.028^{\pm 0.019}$ & $0.007^{\pm 0.028}$ \\
\rotatebox{90}{} & \rotatebox{90}{} & PI-Width & $93.9^{\pm 11.1}$ & $\textbf{83.4}^{\pm 11.2}$ \\
\rotatebox{90}{} & \rotatebox{90}{} & Winkler & $1.50^{\pm 0.09}$ & $\textbf{1.49}^{\pm 0.07}$ \\
\cmidrule{2-5}
\rotatebox{90}{} & \multirow[c]{3}{*}{\rotatebox{90}{LGBM}} & $\Delta$ Cov & $0.017^{\pm 0.016}$ & $-0.000^{\pm 0.016}$ \\
\rotatebox{90}{} & \rotatebox{90}{} & PI-Width & $85.6^{\pm 7.4}$ & $\textbf{77.3}^{\pm 6.0}$ \\
\rotatebox{90}{} & \rotatebox{90}{} & Winkler & $1.45^{\pm 0.06}$ & $\textbf{1.41}^{\pm 0.04}$ \\
\cmidrule{2-5}
\rotatebox{90}{} & \multirow[c]{3}{*}{\rotatebox{90}{Ridge}} & $\Delta$ Cov & $0.010^{\pm 0.007}$ & $-0.001^{\pm 0.007}$ \\
\rotatebox{90}{} & \rotatebox{90}{} & PI-Width & $79.9^{\pm 4.7}$ & $\textbf{74.6}^{\pm 1.6}$ \\
\rotatebox{90}{} & \rotatebox{90}{} & Winkler & $1.35^{\pm 0.06}$ & $\textbf{1.32}^{\pm 0.02}$ \\
\cmidrule{2-5}
\rotatebox{90}{} & \multirow[c]{3}{*}{\rotatebox{90}{LSTM}} & $\Delta$ Cov & $-0.002^{\pm 0.005}$ & $-0.006^{\pm 0.007}$ \\
\rotatebox{90}{} & \rotatebox{90}{} & PI-Width & $62.7^{\pm 1.5}$ & $\textbf{60.9}^{\pm 2.3}$ \\
\rotatebox{90}{} & \rotatebox{90}{} & Winkler & $\textbf{1.33}^{\pm 0.01}$ & $\textbf{1.33}^{\pm 0.02}$ \\
\cmidrule{1-5}
\multirow[c]{12}{*}{\rotatebox{90}{Sap flow}} & \multirow[c]{3}{*}{\rotatebox{90}{Forest}} & $\Delta$ Cov & $0.009^{\pm 0.019}$ & $-0.044^{\pm 0.025}$ \\
\rotatebox{90}{} & \rotatebox{90}{} & PI-Width & $1078.7^{\pm 73.7}$ & $\textbf{857.2}^{\pm 56.7}$ \\
\rotatebox{90}{} & \rotatebox{90}{} & Winkler & $\textbf{0.30}^{\pm 0.01}$ & $\textbf{0.30}^{\pm 0.01}$ \\
\cmidrule{2-5}
\rotatebox{90}{} & \multirow[c]{3}{*}{\rotatebox{90}{LGBM}} & $\Delta$ Cov & $-0.016^{\pm 0.013}$ & $-0.095^{\pm 0.007}$ \\
\rotatebox{90}{} & \rotatebox{90}{} & PI-Width & $875.5^{\pm 49.8}$ & $\textbf{690.9}^{\pm 31.0}$ \\
\rotatebox{90}{} & \rotatebox{90}{} & Winkler & $\textbf{0.27}^{\pm 0.00}$ & $0.28^{\pm 0.01}$ \\
\cmidrule{2-5}
\rotatebox{90}{} & \multirow[c]{3}{*}{\rotatebox{90}{Ridge}} & $\Delta$ Cov & $-0.025^{\pm 0.023}$ & $-0.050^{\pm 0.020}$ \\
\rotatebox{90}{} & \rotatebox{90}{} & PI-Width & $1639.6^{\pm 93.2}$ & $\textbf{1389.0}^{\pm 128.0}$ \\
\rotatebox{90}{} & \rotatebox{90}{} & Winkler & $0.46^{\pm 0.05}$ & $\textbf{0.42}^{\pm 0.03}$ \\
\cmidrule{2-5}
\rotatebox{90}{} & \multirow[c]{3}{*}{\rotatebox{90}{LSTM}} & $\Delta$ Cov & $0.004^{\pm 0.004}$ & $-0.003^{\pm 0.003}$ \\
\rotatebox{90}{} & \rotatebox{90}{} & PI-Width & $594.3^{\pm 7.7}$ & $\textbf{557.8}^{\pm 8.9}$ \\
\rotatebox{90}{} & \rotatebox{90}{} & Winkler & $\textbf{0.19}^{\pm 0.01}$ & $\textbf{0.19}^{\pm 0.00}$ \\
\bottomrule
\end{tabular}
\end{table}

\setTableCaption{Performance of the weighted sample quantile (Sample) and the weighted empirical CDF (ECDF) quantile strategies that \ourmethod uses for the miscoverage $\alpha = 0.10$ on the different non-solar datasets. The error term represents the standard deviation over repeated runs.}


\begin{table}
\centering
\caption{\nextTableCaption}
\label{tab:results-cdf-other}
\vskip 0.15in
\small
\begin{tabular}{lllrr}
\toprule
 &  & UC & \text{Sample} & \text{ECDF} \\
Data & FC &  &  &  \\
\midrule
\multirow[c]{12}{*}{\rotatebox{90}{Air 10 PM}} & \multirow[c]{3}{*}{\rotatebox{90}{Forest}} & $\Delta$ Cov & $0.028^{\pm 0.019}$ & $0.027^{\pm 0.021}$ \\
\rotatebox{90}{} & \rotatebox{90}{} & PI-Width & $93.9^{\pm 11.1}$ & $\textbf{93.4}^{\pm 11.0}$ \\
\rotatebox{90}{} & \rotatebox{90}{} & Winkler & $\textbf{1.50}^{\pm 0.09}$ & $\textbf{1.50}^{\pm 0.09}$ \\
\cmidrule{2-5}
\rotatebox{90}{} & \multirow[c]{3}{*}{\rotatebox{90}{LGBM}} & $\Delta$ Cov & $0.017^{\pm 0.016}$ & $0.014^{\pm 0.015}$ \\
\rotatebox{90}{} & \rotatebox{90}{} & PI-Width & $85.6^{\pm 7.4}$ & $\textbf{84.5}^{\pm 8.3}$ \\
\rotatebox{90}{} & \rotatebox{90}{} & Winkler & $1.45^{\pm 0.06}$ & $\textbf{1.43}^{\pm 0.09}$ \\
\cmidrule{2-5}
\rotatebox{90}{} & \multirow[c]{3}{*}{\rotatebox{90}{Ridge}} & $\Delta$ Cov & $0.010^{\pm 0.007}$ & $0.013^{\pm 0.010}$ \\
\rotatebox{90}{} & \rotatebox{90}{} & PI-Width & $79.9^{\pm 4.7}$ & $\textbf{79.3}^{\pm 4.6}$ \\
\rotatebox{90}{} & \rotatebox{90}{} & Winkler & $1.35^{\pm 0.06}$ & $\textbf{1.34}^{\pm 0.05}$ \\
\cmidrule{2-5}
\rotatebox{90}{} & \multirow[c]{3}{*}{\rotatebox{90}{LSTM}} & $\Delta$ Cov & $-0.002^{\pm 0.005}$ & $-0.005^{\pm 0.010}$ \\
\rotatebox{90}{} & \rotatebox{90}{} & PI-Width & $\textbf{62.7}^{\pm 1.5}$ & $65.6^{\pm 20.1}$ \\
\rotatebox{90}{} & \rotatebox{90}{} & Winkler & $\textbf{1.33}^{\pm 0.01}$ & $1.35^{\pm 0.15}$ \\
\cmidrule{1-5}
\multirow[c]{12}{*}{\rotatebox{90}{Air 25 PM}} & \multirow[c]{3}{*}{\rotatebox{90}{Forest}} & $\Delta$ Cov & $-0.024^{\pm 0.017}$ & $-0.024^{\pm 0.019}$ \\
\rotatebox{90}{} & \rotatebox{90}{} & PI-Width & $48.1^{\pm 5.6}$ & $\textbf{47.7}^{\pm 4.5}$ \\
\rotatebox{90}{} & \rotatebox{90}{} & Winkler & $1.12^{\pm 0.05}$ & $\textbf{1.11}^{\pm 0.01}$ \\
\cmidrule{2-5}
\rotatebox{90}{} & \multirow[c]{3}{*}{\rotatebox{90}{LGBM}} & $\Delta$ Cov & $-0.021^{\pm 0.019}$ & $-0.023^{\pm 0.019}$ \\
\rotatebox{90}{} & \rotatebox{90}{} & PI-Width & $46.8^{\pm 6.0}$ & $\textbf{46.6}^{\pm 6.1}$ \\
\rotatebox{90}{} & \rotatebox{90}{} & Winkler & $\textbf{1.10}^{\pm 0.05}$ & $\textbf{1.10}^{\pm 0.05}$ \\
\cmidrule{2-5}
\rotatebox{90}{} & \multirow[c]{3}{*}{\rotatebox{90}{Ridge}} & $\Delta$ Cov & $-0.005^{\pm 0.026}$ & $-0.007^{\pm 0.027}$ \\
\rotatebox{90}{} & \rotatebox{90}{} & PI-Width & $49.3^{\pm 10.6}$ & $\textbf{49.1}^{\pm 10.7}$ \\
\rotatebox{90}{} & \rotatebox{90}{} & Winkler & $\textbf{1.04}^{\pm 0.11}$ & $\textbf{1.04}^{\pm 0.11}$ \\
\cmidrule{2-5}
\rotatebox{90}{} & \multirow[c]{3}{*}{\rotatebox{90}{LSTM}} & $\Delta$ Cov & $0.007^{\pm 0.008}$ & $-0.002^{\pm 0.015}$ \\
\rotatebox{90}{} & \rotatebox{90}{} & PI-Width & $40.7^{\pm 4.3}$ & $\textbf{35.0}^{\pm 2.1}$ \\
\rotatebox{90}{} & \rotatebox{90}{} & Winkler & $0.88^{\pm 0.05}$ & $\textbf{0.82}^{\pm 0.05}$ \\
\cmidrule{1-5}
\multirow[c]{12}{*}{\rotatebox{90}{Sap flow}} & \multirow[c]{3}{*}{\rotatebox{90}{Forest}} & $\Delta$ Cov & $0.009^{\pm 0.019}$ & $0.011^{\pm 0.019}$ \\
\rotatebox{90}{} & \rotatebox{90}{} & PI-Width & $\textbf{1078.7}^{\pm 73.7}$ & $1082.3^{\pm 74.0}$ \\
\rotatebox{90}{} & \rotatebox{90}{} & Winkler & $\textbf{0.30}^{\pm 0.01}$ & $\textbf{0.30}^{\pm 0.01}$ \\
\cmidrule{2-5}
\rotatebox{90}{} & \multirow[c]{3}{*}{\rotatebox{90}{LGBM}} & $\Delta$ Cov & $-0.016^{\pm 0.013}$ & $-0.017^{\pm 0.013}$ \\
\rotatebox{90}{} & \rotatebox{90}{} & PI-Width & $\textbf{875.5}^{\pm 49.8}$ & $875.6^{\pm 49.5}$ \\
\rotatebox{90}{} & \rotatebox{90}{} & Winkler & $\textbf{0.27}^{\pm 0.00}$ & $\textbf{0.27}^{\pm 0.00}$ \\
\cmidrule{2-5}
\rotatebox{90}{} & \multirow[c]{3}{*}{\rotatebox{90}{Ridge}} & $\Delta$ Cov & $-0.025^{\pm 0.023}$ & $-0.035^{\pm 0.046}$ \\
\rotatebox{90}{} & \rotatebox{90}{} & PI-Width & $1639.6^{\pm 93.2}$ & $\textbf{1634.0}^{\pm 90.2}$ \\
\rotatebox{90}{} & \rotatebox{90}{} & Winkler & $\textbf{0.46}^{\pm 0.05}$ & $0.50^{\pm 0.13}$ \\
\cmidrule{2-5}
\rotatebox{90}{} & \multirow[c]{3}{*}{\rotatebox{90}{LSTM}} & $\Delta$ Cov & $0.004^{\pm 0.004}$ & $0.005^{\pm 0.003}$ \\
\rotatebox{90}{} & \rotatebox{90}{} & PI-Width & $594.3^{\pm 7.7}$ & $\textbf{592.0}^{\pm 9.7}$ \\
\rotatebox{90}{} & \rotatebox{90}{} & Winkler & $\textbf{0.19}^{\pm 0.01}$ & $\textbf{0.19}^{\pm 0.00}$ \\
\cmidrule{1-5}
\multirow[c]{3}{*}{Streamflow} & \multirow[c]{3}{*}{LSTM} & $\Delta$ Cov & $0.001^{\pm 0.041}$ & $0.028^{\pm 0.002}$ \\
 &  & PI-Width & $\textbf{1.39}^{\pm 0.17}$ & $1.49^{\pm 0.02}$   \\
 &  & Winkler & $0.79^{\pm 0.03}$ & $\textbf{0.77}^{\pm 0.01}$    \\
\bottomrule
\end{tabular}
\end{table}

\setTableCaption{Performance of the CP algorithms \ourmethod, \spic, \enbpi, and \nexcp, each combined with \adaptiveci, for the miscoverage $\alpha = 0.10$ on the solar datasets. Bold numbers correspond to the best result for the respective metric in the experiment (PI-Width and Winkler score). The error term represents the standard deviation over repeated runs. \dag \xspace combined with \adaptiveci.}


\begin{table*}
\centering
\caption{\nextTableCaption}
\label{tab:results-adaptive-solar}
\vskip 0.15in
\small
\begin{tabular}{lllrrrrrr}
\toprule
 &  & UC & \rotatebox{90}{\ourmethod\dag} & \rotatebox{90}{\spic\dag} & \rotatebox{90}{\enbpi\dag} & \rotatebox{90}{\nexcp\dag} & \rotatebox{90}{CopulaCPTS\dag} & \rotatebox{90}{CP/CF-RNN\dag} \\
 &  &  &  &  &  &  &  &  \\
\midrule
\multirow[c]{12}{*}{\rotatebox{90}{Solar 3Y}} & \multirow[c]{3}{*}{\rotatebox{90}{Forest}} & $\Delta$ Cov & $0.004^{\pm 0.001}$ & $0.002^{\pm 0.000}$ & $-0.001$ & $-0.000$ & $0.000$ & $0.000$ \\
\rotatebox{90}{} & \rotatebox{90}{} & PI-Width & $\textbf{29.8}^{\pm 0.7}$ & $97.6^{\pm 0.1}$ & $155.7$ & $170.8$ & $168.6$ & $168.5$ \\
\rotatebox{90}{} & \rotatebox{90}{} & Winkler & $\textbf{0.59}^{\pm 0.03}$ & $1.69^{\pm 0.00}$ & $2.43$ & $2.54$ & $2.52$ & $2.52$ \\
\cmidrule{2-9}
\rotatebox{90}{} & \multirow[c]{3}{*}{\rotatebox{90}{LGBM}} & $\Delta$ Cov & $-0.001^{\pm 0.000}$ & $0.002^{\pm 0.000}$ & $-0.001$ & $-0.000$ & $0.000$ & $0.000$ \\
\rotatebox{90}{} & \rotatebox{90}{} & PI-Width & $\textbf{39.9}^{\pm 7.0}$ & $95.8^{\pm 0.1}$ & $155.0$ & $164.8$ & $162.3$ & $162.4$ \\
\rotatebox{90}{} & \rotatebox{90}{} & Winkler & $\textbf{0.62}^{\pm 0.10}$ & $1.70^{\pm 0.00}$ & $2.49$ & $2.57$ & $2.54$ & $2.54$ \\
\cmidrule{2-9}
\rotatebox{90}{} & \multirow[c]{3}{*}{\rotatebox{90}{Ridge}} & $\Delta$ Cov & $0.002^{\pm 0.000}$ & $-0.000^{\pm 0.000}$ & $-0.006$ & $0.000$ & $0.000$ & $-0.000$ \\
\rotatebox{90}{} & \rotatebox{90}{} & PI-Width & $\textbf{36.0}^{\pm 0.3}$ & $110.6^{\pm 0.1}$ & $166.3$ & $177.3$ & $172.8$ & $172.4$ \\
\rotatebox{90}{} & \rotatebox{90}{} & Winkler & $\textbf{0.62}^{\pm 0.00}$ & $1.77^{\pm 0.00}$ & $2.28$ & $2.63$ & $2.61$ & $2.58$ \\
\cmidrule{2-9}
\rotatebox{90}{} & \multirow[c]{3}{*}{\rotatebox{90}{LSTM}} & $\Delta$ Cov & $0.000^{\pm 0.000}$ & $0.001^{\pm 0.000}$ & $-0.002$ & $-0.000$ & $0.001$ & $0.001$ \\
\rotatebox{90}{} & \rotatebox{90}{} & PI-Width & $\textbf{17.8}^{\pm 0.2}$ & $25.5^{\pm 0.0}$ & $30.6$ & $30.6$ & $28.8$ & $28.8$ \\
\rotatebox{90}{} & \rotatebox{90}{} & Winkler & $\textbf{0.30}^{\pm 0.00}$ & $0.59^{\pm 0.00}$ & $0.63$ & $0.62$ & $0.63$ & $0.63$ \\
\cmidrule{1-9} \cmidrule{2-9}
\multirow[c]{12}{*}{\rotatebox{90}{Solar 1Y}} & \multirow[c]{3}{*}{\rotatebox{90}{Forest}} & $\Delta$ Cov & $0.007^{\pm 0.001}$ & $0.011^{\pm 0.000}$ & $-0.002$ & $-0.001$ & $0.002$ & $0.003$ \\
\rotatebox{90}{} & \rotatebox{90}{} & PI-Width & $\textbf{21.5}^{\pm 1.2}$ & $72.6^{\pm 0.1}$ & $109.5$ & $126.7$ & $128.9$ & $129.5$ \\
\rotatebox{90}{} & \rotatebox{90}{} & Winkler & $\textbf{0.40}^{\pm 0.02}$ & $1.14^{\pm 0.00}$ & $1.63$ & $1.85$ & $1.84$ & $1.84$ \\
\cmidrule{2-9}
\rotatebox{90}{} & \multirow[c]{3}{*}{\rotatebox{90}{LGBM}} & $\Delta$ Cov & $0.000^{\pm 0.001}$ & $0.005^{\pm 0.000}$ & $-0.001$ & $-0.000$ & $0.002$ & $0.003$ \\
\rotatebox{90}{} & \rotatebox{90}{} & PI-Width & $\textbf{41.6}^{\pm 1.9}$ & $69.2^{\pm 0.2}$ & $106.2$ & $117.1$ & $119.1$ & $119.4$ \\
\rotatebox{90}{} & \rotatebox{90}{} & Winkler & $\textbf{0.56}^{\pm 0.04}$ & $1.15^{\pm 0.00}$ & $1.63$ & $1.78$ & $1.77$ & $1.77$ \\
\cmidrule{2-9}
\rotatebox{90}{} & \multirow[c]{3}{*}{\rotatebox{90}{Ridge}} & $\Delta$ Cov & $-0.003^{\pm 0.005}$ & $0.004^{\pm 0.000}$ & $-0.002$ & $-0.000$ & $-0.000$ & $-0.000$ \\
\rotatebox{90}{} & \rotatebox{90}{} & PI-Width & $\textbf{61.5}^{\pm 9.9}$ & $98.9^{\pm 0.2}$ & $167.8$ & $174.8$ & $174.9$ & $175.6$ \\
\rotatebox{90}{} & \rotatebox{90}{} & Winkler & $\textbf{0.87}^{\pm 0.11}$ & $1.32^{\pm 0.00}$ & $2.01$ & $2.08$ & $2.08$ & $2.08$ \\
\cmidrule{2-9}
\rotatebox{90}{} & \multirow[c]{3}{*}{\rotatebox{90}{LSTM}} & $\Delta$ Cov & $0.001^{\pm 0.000}$ & $0.003^{\pm 0.000}$ & $-0.001$ & $-0.001$ & $0.001$ & $0.001$ \\
\rotatebox{90}{} & \rotatebox{90}{} & PI-Width & $\textbf{14.3}^{\pm 0.7}$ & $18.9^{\pm 0.0}$ & $21.8$ & $20.4$ & $19.9$ & $20.0$ \\
\rotatebox{90}{} & \rotatebox{90}{} & Winkler & $\textbf{0.22}^{\pm 0.01}$ & $0.40^{\pm 0.00}$ & $0.42$ & $0.41$ & $0.41$ & $0.41$ \\
\cmidrule{1-9} \cmidrule{2-9}
\multirow[c]{9}{*}{\rotatebox{90}{Solar Small}} & \multirow[c]{3}{*}{\rotatebox{90}{Forest}} & $\Delta$ Cov & $0.006^{\pm 0.003}$ & $-0.023^{\pm 0.002}$ & $-0.003$ & $-0.014$ & $-0.007$ & $-0.009$ \\
\rotatebox{90}{} & \rotatebox{90}{} & PI-Width & $\textbf{33.6}^{\pm 2.9}$ & $64.7^{\pm 0.7}$ & $92.9$ & $127.3$ & $131.1$ & $128.1$ \\
\rotatebox{90}{} & \rotatebox{90}{} & Winkler & $\textbf{1.01}^{\pm 0.04}$ & $1.95^{\pm 0.01}$ & $2.38$ & $3.27$ & $3.25$ & $3.23$ \\
\cmidrule{2-9}
\rotatebox{90}{} & \multirow[c]{3}{*}{\rotatebox{90}{LGBM}} & $\Delta$ Cov & $-0.001^{\pm 0.002}$ & $-0.015^{\pm 0.001}$ & $-0.005$ & $-0.005$ & $-0.008$ & $-0.010$ \\
\rotatebox{90}{} & \rotatebox{90}{} & PI-Width & $\textbf{44.8}^{\pm 1.0}$ & $60.9^{\pm 0.6}$ & $96.4$ & $135.7$ & $128.3$ & $125.0$ \\
\rotatebox{90}{} & \rotatebox{90}{} & Winkler & $\textbf{1.18}^{\pm 0.04}$ & $1.95^{\pm 0.01}$ & $2.54$ & $3.32$ & $3.33$ & $3.31$ \\
\cmidrule{2-9}
\rotatebox{90}{} & \multirow[c]{3}{*}{\rotatebox{90}{Ridge}} & $\Delta$ Cov & $0.008^{\pm 0.004}$ & $-0.015^{\pm 0.001}$ & $-0.003$ & $0.008$ & $0.008$ & $0.006$ \\
\rotatebox{90}{} & \rotatebox{90}{} & PI-Width & $104.7^{\pm 19.8}$ & $\textbf{69.3}^{\pm 0.7}$ & $102.5$ & $97.7$ & $96.2$ & $93.0$ \\
\rotatebox{90}{} & \rotatebox{90}{} & Winkler & $2.44^{\pm 0.42}$ & $\textbf{1.92}^{\pm 0.01}$ & $2.56$ & $2.75$ & $2.75$ & $2.72$ \\
\bottomrule
\end{tabular}
\end{table*}

\setTableCaption{Performance of the CP algorithms \ourmethod, \spic, \enbpi, and \nexcp, each combined with \adaptiveci, for the miscoverage $\alpha = 0.10$  on the different non-solar datasets. Bold numbers correspond to the best result for the respective metric in the experiment (PI-Width and Winkler score). The error term represents the standard deviation over repeated runs. \dag \xspace combined with \adaptiveci.}


\begin{table*}
\centering
\caption{\nextTableCaption}
\label{tab:results-adaptive-other}
\vskip 0.15in
\small
\begin{tabular}{lllrrrrrr}
\toprule
 &  & UC & \rotatebox{90}{\ourmethod\dag} & \rotatebox{90}{\spic\dag} & \rotatebox{90}{\enbpi\dag} & \rotatebox{90}{\nexcp\dag} & \rotatebox{90}{CopulaCPTS\dag} & \rotatebox{90}{CP/CF-RNN\dag} \\
 &  &  &  &  &  &  &  &  \\
\midrule
\multirow[c]{12}{*}{\rotatebox{90}{Air 10 PM}} & \multirow[c]{3}{*}{\rotatebox{90}{Forest}} & $\Delta$ Cov & $0.002^{\pm 0.001}$ & $-0.002^{\pm 0.000}$ & $-0.002$ & $0.000$ & $-0.003$ & $-0.005$ \\
\rotatebox{90}{} & \rotatebox{90}{} & PI-Width & $\textbf{80.7}^{\pm 2.6}$ & $112.6^{\pm 0.1}$ & $250.4$ & $270.5$ & $228.7$ & $228.3$ \\
\rotatebox{90}{} & \rotatebox{90}{} & Winkler & $\textbf{1.45}^{\pm 0.06}$ & $1.96^{\pm 0.00}$ & $3.70$ & $3.92$ & $3.37$ & $3.39$ \\
\cmidrule{2-9}
\rotatebox{90}{} & \multirow[c]{3}{*}{\rotatebox{90}{LGBM}} & $\Delta$ Cov & $0.001^{\pm 0.001}$ & $0.001^{\pm 0.000}$ & $-0.001$ & $0.000$ & $-0.002$ & $-0.003$ \\
\rotatebox{90}{} & \rotatebox{90}{} & PI-Width & $\textbf{78.5}^{\pm 2.2}$ & $102.5^{\pm 0.2}$ & $220.0$ & $233.2$ & $196.3$ & $196.3$ \\
\rotatebox{90}{} & \rotatebox{90}{} & Winkler & $\textbf{1.38}^{\pm 0.04}$ & $1.83^{\pm 0.00}$ & $3.38$ & $3.54$ & $3.01$ & $3.01$ \\
\cmidrule{2-9}
\rotatebox{90}{} & \multirow[c]{3}{*}{\rotatebox{90}{Ridge}} & $\Delta$ Cov & $0.001^{\pm 0.001}$ & $0.000^{\pm 0.000}$ & $-0.001$ & $0.001$ & $0.000$ & $0.000$ \\
\rotatebox{90}{} & \rotatebox{90}{} & PI-Width & $\textbf{75.3}^{\pm 0.8}$ & $83.7^{\pm 0.1}$ & $147.2$ & $159.5$ & $142.7$ & $142.3$ \\
\rotatebox{90}{} & \rotatebox{90}{} & Winkler & $\textbf{1.30}^{\pm 0.02}$ & $1.46^{\pm 0.00}$ & $2.46$ & $2.67$ & $2.41$ & $2.42$ \\
\cmidrule{2-9}
\rotatebox{90}{} & \multirow[c]{3}{*}{\rotatebox{90}{LSTM}} & $\Delta$ Cov & $-0.001^{\pm 0.000}$ & $0.000^{\pm 0.000}$ & $0.000$ & $0.001$ & $0.001$ & $0.000$ \\
\rotatebox{90}{} & \rotatebox{90}{} & PI-Width & $65.4^{\pm 0.3}$ & $\textbf{59.2}^{\pm 0.1}$ & $69.2$ & $65.7$ & $62.9$ & $62.5$ \\
\rotatebox{90}{} & \rotatebox{90}{} & Winkler & $1.30^{\pm 0.01}$ & $\textbf{1.18}^{\pm 0.00}$ & $1.32$ & $1.29$ & $1.29$ & $1.30$ \\
\cmidrule{1-9} \cmidrule{2-9}
\multirow[c]{12}{*}{\rotatebox{90}{Air 25 PM}} & \multirow[c]{3}{*}{\rotatebox{90}{Forest}} & $\Delta$ Cov & $-0.001^{\pm 0.001}$ & $-0.002^{\pm 0.000}$ & $-0.003$ & $-0.001$ & $-0.004$ & $-0.008$ \\
\rotatebox{90}{} & \rotatebox{90}{} & PI-Width & $\textbf{53.4}^{\pm 1.1}$ & $86.5^{\pm 0.1}$ & $221.7$ & $245.1$ & $205.2$ & $200.8$ \\
\rotatebox{90}{} & \rotatebox{90}{} & Winkler & $\textbf{1.10}^{\pm 0.02}$ & $1.76^{\pm 0.00}$ & $3.68$ & $3.97$ & $3.43$ & $3.47$ \\
\cmidrule{2-9}
\rotatebox{90}{} & \multirow[c]{3}{*}{\rotatebox{90}{LGBM}} & $\Delta$ Cov & $-0.000^{\pm 0.001}$ & $0.000^{\pm 0.000}$ & $-0.001$ & $0.000$ & $-0.003$ & $-0.007$ \\
\rotatebox{90}{} & \rotatebox{90}{} & PI-Width & $\textbf{52.5}^{\pm 1.7}$ & $76.1^{\pm 0.1}$ & $180.6$ & $191.7$ & $151.1$ & $147.2$ \\
\rotatebox{90}{} & \rotatebox{90}{} & Winkler & $\textbf{1.10}^{\pm 0.05}$ & $1.69^{\pm 0.00}$ & $3.17$ & $3.38$ & $2.73$ & $2.74$ \\
\cmidrule{2-9}
\rotatebox{90}{} & \multirow[c]{3}{*}{\rotatebox{90}{Ridge}} & $\Delta$ Cov & $-0.000^{\pm 0.001}$ & $-0.001^{\pm 0.000}$ & $-0.001$ & $0.001$ & $0.001$ & $0.000$ \\
\rotatebox{90}{} & \rotatebox{90}{} & PI-Width & $\textbf{50.4}^{\pm 3.9}$ & $63.2^{\pm 0.1}$ & $130.0$ & $143.6$ & $130.3$ & $129.5$ \\
\rotatebox{90}{} & \rotatebox{90}{} & Winkler & $\textbf{1.04}^{\pm 0.10}$ & $1.33^{\pm 0.00}$ & $2.36$ & $2.64$ & $2.53$ & $2.52$ \\
\cmidrule{2-9}
\rotatebox{90}{} & \multirow[c]{3}{*}{\rotatebox{90}{LSTM}} & $\Delta$ Cov & $-0.001^{\pm 0.001}$ & $-0.001^{\pm 0.000}$ & $0.000$ & $0.001$ & $-0.001$ & $-0.000$ \\
\rotatebox{90}{} & \rotatebox{90}{} & PI-Width & $40.3^{\pm 4.1}$ & $\textbf{36.3}^{\pm 0.0}$ & $44.6$ & $40.8$ & $38.8$ & $38.9$ \\
\rotatebox{90}{} & \rotatebox{90}{} & Winkler & $\textbf{0.89}^{\pm 0.06}$ & $0.89^{\pm 0.00}$ & $0.96$ & $0.94$ & $0.95$ & $0.93$ \\
\cmidrule{1-9} \cmidrule{2-9}
\multirow[c]{12}{*}{\rotatebox{90}{Sap flow}} & \multirow[c]{3}{*}{\rotatebox{90}{Forest}} & $\Delta$ Cov & $-0.004^{\pm 0.007}$ & $0.004^{\pm 0.000}$ & $-0.006$ & $0.000$ & $0.004$ & $0.003$ \\
\rotatebox{90}{} & \rotatebox{90}{} & PI-Width & $\textbf{1006.9}^{\pm 44.5}$ & $1651.0^{\pm 1.9}$ & $4142.8$ & $6151.7$ & $6034.6$ & $6022.0$ \\
\rotatebox{90}{} & \rotatebox{90}{} & Winkler & $\textbf{0.29}^{\pm 0.01}$ & $0.51^{\pm 0.00}$ & $1.11$ & $1.54$ & $1.53$ & $1.53$ \\
\cmidrule{2-9}
\rotatebox{90}{} & \multirow[c]{3}{*}{\rotatebox{90}{LGBM}} & $\Delta$ Cov & $-0.001^{\pm 0.003}$ & $0.007^{\pm 0.000}$ & $-0.005$ & $-0.000$ & $0.006$ & $0.008$ \\
\rotatebox{90}{} & \rotatebox{90}{} & PI-Width & $\textbf{919.1}^{\pm 25.8}$ & $1601.7^{\pm 3.0}$ & $3327.0$ & $4845.1$ & $4787.0$ & $5016.1$ \\
\rotatebox{90}{} & \rotatebox{90}{} & Winkler & $\textbf{0.26}^{\pm 0.01}$ & $0.45^{\pm 0.00}$ & $0.88$ & $1.23$ & $1.22$ & $1.28$ \\
\cmidrule{2-9}
\rotatebox{90}{} & \multirow[c]{3}{*}{\rotatebox{90}{Ridge}} & $\Delta$ Cov & $-0.006^{\pm 0.002}$ & $-0.121^{\pm 0.000}$ & $-0.003$ & $-0.000$ & $-0.004$ & $-0.126$ \\
\rotatebox{90}{} & \rotatebox{90}{} & PI-Width & $\textbf{1492.5}^{\pm 29.9}$ & $2974.7^{\pm 6.9}$ & $3431.4$ & $10665.6$ & $10426.2$ & $9521.9$ \\
\rotatebox{90}{} & \rotatebox{90}{} & Winkler & $\textbf{0.39}^{\pm 0.01}$ & $1.53^{\pm 0.00}$ & $0.87$ & $2.39$ & $2.36$ & $2.87$ \\
\cmidrule{2-9}
\rotatebox{90}{} & \multirow[c]{3}{*}{\rotatebox{90}{LSTM}} & $\Delta$ Cov & $0.000^{\pm 0.000}$ & $0.002^{\pm 0.000}$ & $-0.001$ & $-0.000$ & $0.001$ & $0.002$ \\
\rotatebox{90}{} & \rotatebox{90}{} & PI-Width & $\textbf{595.0}^{\pm 7.5}$ & $639.5^{\pm 1.8}$ & $863.1$ & $1019.1$ & $980.1$ & $992.6$ \\
\rotatebox{90}{} & \rotatebox{90}{} & Winkler & $\textbf{0.19}^{\pm 0.00}$ & $0.22^{\pm 0.00}$ & $0.28$ & $0.32$ & $0.31$ & $0.32$ \\
\bottomrule
\end{tabular}
\end{table*}

\setTableCaption{Performance of the evaluated CP algorithms on the \solarThreeY datasets for the miscoverage levels $\alpha \in \{0.05, 0.10, 0.15\}$. The column FC specifies the prediction algorithm used for the experiment (Forest: Random Forest, LGBM: LightGBM, Ridge: Ridge Regression, LSTM: LSTM neural network). Bold numbers correspond to the best result for the respective metric in the experiment (PI-Width and Winkler score), given that $\Delta \text{Cov} \ge - 0.25 \alpha$, i.e., the specific algorithm reached at least approximate coverage (the result is grayed out otherwise). The error term represents the standard deviation over repeated runs (results without an error term are from deterministic models).}


\begin{table*}
\centering
\caption{\nextTableCaption}
\label{tab:nsdb-60m-2018-20-overall-v2}
\vskip 0.15in
\small
\begin{tabular}{lllrrrrrr|r}
\toprule
 &  &  & \rotatebox{90}{\text{\ourmethod}} & \rotatebox{90}{\text{SPCI}} & \rotatebox{90}{\text{EnbPI}} & \rotatebox{90}{\text{NexCP}} & \rotatebox{90}{\text{CopulaCPTS}} & \rotatebox{90}{\text{CP/CF-RNN}} & \rotatebox{90}{\text{AdaptiveCI}} \\
\text{FC} & \text{$\alpha$} &  &  &  &  &  &  &  &  \\
\midrule
\multirow[c]{9}{*}{\rotatebox{90}{\text{Forest}}} & \multirow[c]{3}{*}{\rotatebox{90}{0.05}} & \text{$\Delta$ Cov} & $0.006^{\pm 0.006}$ & $0.007^{\pm 0.000}$ & $\textcolor{gray}{\text{-}0.030}$ & $\text{-}0.002$ & $0.002$ & $0.002$ &   \\
\rotatebox{90}{} & \rotatebox{90}{} & \text{PI-Width} & $\textbf{47.8}^{\pm 9.5}$ & $149.2^{\pm 0.1}$ & $\textcolor{gray}{173.0}$ & $216.5$ & $237.2$ & $236.7$ &   \\
\rotatebox{90}{} & \rotatebox{90}{} & \text{Winkler} & $\textbf{0.99}^{\pm 0.35}$ & $2.22^{\pm 0.00}$ & $\textcolor{gray}{3.01}$ & $2.95$ & $3.30$ & $3.30$ &   \\
\cmidrule{2-10}
\rotatebox{90}{} & \multirow[c]{3}{*}{\rotatebox{90}{0.10}} & \text{$\Delta$ Cov} & $0.029^{\pm 0.012}$ & $0.012^{\pm 0.000}$ & $\textcolor{gray}{\text{-}0.031}$ & $\text{-}0.002$ & $0.005$ & $0.004$ &   \\
\rotatebox{90}{} & \rotatebox{90}{} & \text{PI-Width} & $\textbf{39.0}^{\pm 6.2}$ & $103.1^{\pm 0.1}$ & $\textcolor{gray}{131.1}$ & $166.6$ & $174.9$ & $174.6$ &   \\
\rotatebox{90}{} & \rotatebox{90}{} & \text{Winkler} & $\textbf{0.73}^{\pm 0.20}$ & $1.74^{\pm 0.00}$ & $\textcolor{gray}{2.47}$ & $2.53$ & $2.75$ & $2.76$ &   \\
\cmidrule{2-10}
\rotatebox{90}{} & \multirow[c]{3}{*}{\rotatebox{90}{0.15}} & \text{$\Delta$ Cov} & $0.052^{\pm 0.018}$ & $0.014^{\pm 0.000}$ & $\text{-}0.027$ & $\text{-}0.002$ & $0.006$ & $0.006$ &   \\
\rotatebox{90}{} & \rotatebox{90}{} & \text{PI-Width} & $\textbf{33.6}^{\pm 4.6}$ & $75.3^{\pm 0.1}$ & $101.4$ & $129.1$ & $132.2$ & $132.0$ &   \\
\rotatebox{90}{} & \rotatebox{90}{} & \text{Winkler} & $\textbf{0.61}^{\pm 0.15}$ & $1.46^{\pm 0.00}$ & $2.12$ & $2.23$ & $2.38$ & $2.39$ &   \\
\cmidrule{1-10} \cmidrule{2-10}
\multirow[c]{9}{*}{\rotatebox{90}{\text{LGBM}}} & \multirow[c]{3}{*}{\rotatebox{90}{0.05}} & \text{$\Delta$ Cov} & $\text{-}0.008^{\pm 0.002}$ & $0.007^{\pm 0.000}$ & $\textcolor{gray}{\text{-}0.022}$ & $\text{-}0.002$ & $0.002$ & $0.002$ & $0.001$ \\
\rotatebox{90}{} & \rotatebox{90}{} & \text{PI-Width} & $\textbf{45.6}^{\pm 0.8}$ & $148.0^{\pm 0.1}$ & $\textcolor{gray}{183.0}$ & $215.9$ & $237.9$ & $237.8$ & $88.1$ \\
\rotatebox{90}{} & \rotatebox{90}{} & \text{Winkler} & $\textbf{0.72}^{\pm 0.02}$ & $2.25^{\pm 0.00}$ & $\textcolor{gray}{3.09}$ & $3.06$ & $3.45$ & $3.46$ & $1.36$ \\
\cmidrule{2-10}
\rotatebox{90}{} & \multirow[c]{3}{*}{\rotatebox{90}{0.10}} & \text{$\Delta$ Cov} & $0.001^{\pm 0.003}$ & $0.014^{\pm 0.000}$ & $\text{-}0.023$ & $\text{-}0.002$ & $0.006$ & $0.006$ & $0.001$ \\
\rotatebox{90}{} & \rotatebox{90}{} & \text{PI-Width} & $\textbf{37.7}^{\pm 0.7}$ & $102.2^{\pm 0.1}$ & $133.6$ & $159.9$ & $169.8$ & $170.2$ & $67.1$ \\
\rotatebox{90}{} & \rotatebox{90}{} & \text{Winkler} & $\textbf{0.57}^{\pm 0.01}$ & $1.75^{\pm 0.00}$ & $2.52$ & $2.55$ & $2.80$ & $2.81$ & $1.19$ \\
\cmidrule{2-10}
\rotatebox{90}{} & \multirow[c]{3}{*}{\rotatebox{90}{0.15}} & \text{$\Delta$ Cov} & $0.009^{\pm 0.003}$ & $0.017^{\pm 0.000}$ & $\text{-}0.022$ & $\text{-}0.002$ & $0.008$ & $0.009$ & $0.001$ \\
\rotatebox{90}{} & \rotatebox{90}{} & \text{PI-Width} & $\textbf{32.7}^{\pm 0.7}$ & $75.5^{\pm 0.1}$ & $100.6$ & $121.3$ & $126.6$ & $127.2$ & $55.1$ \\
\rotatebox{90}{} & \rotatebox{90}{} & \text{Winkler} & $\textbf{0.50}^{\pm 0.01}$ & $1.46^{\pm 0.00}$ & $2.15$ & $2.21$ & $2.39$ & $2.40$ & $1.11$ \\
\cmidrule{1-10} \cmidrule{2-10}
\multirow[c]{9}{*}{\rotatebox{90}{\text{Ridge}}} & \multirow[c]{3}{*}{\rotatebox{90}{0.05}} & \text{$\Delta$ Cov} & $0.010^{\pm 0.001}$ & $\text{-}0.003^{\pm 0.000}$ & $\textcolor{gray}{\text{-}0.080}$ & $\text{-}0.002$ & $0.003$ & $0.003$ &   \\
\rotatebox{90}{} & \rotatebox{90}{} & \text{PI-Width} & $\textbf{52.7}^{\pm 0.6}$ & $146.3^{\pm 0.0}$ & $\textcolor{gray}{151.0}$ & $226.2$ & $223.8$ & $224.8$ &   \\
\rotatebox{90}{} & \rotatebox{90}{} & \text{Winkler} & $\textbf{0.78}^{\pm 0.00}$ & $2.31^{\pm 0.00}$ & $\textcolor{gray}{3.04}$ & $3.20$ & $3.44$ & $3.46$ &   \\
\cmidrule{2-10}
\rotatebox{90}{} & \multirow[c]{3}{*}{\rotatebox{90}{0.10}} & \text{$\Delta$ Cov} & $0.040^{\pm 0.001}$ & $0.002^{\pm 0.000}$ & $\textcolor{gray}{\text{-}0.074}$ & $\text{-}0.001$ & $0.004$ & $0.005$ &   \\
\rotatebox{90}{} & \rotatebox{90}{} & \text{PI-Width} & $\textbf{44.9}^{\pm 0.5}$ & $108.2^{\pm 0.0}$ & $\textcolor{gray}{131.1}$ & $171.0$ & $166.0$ & $167.7$ &   \\
\rotatebox{90}{} & \rotatebox{90}{} & \text{Winkler} & $\textbf{0.64}^{\pm 0.00}$ & $1.82^{\pm 0.00}$ & $\textcolor{gray}{2.49}$ & $2.66$ & $2.73$ & $2.74$ &   \\
\cmidrule{2-10}
\rotatebox{90}{} & \multirow[c]{3}{*}{\rotatebox{90}{0.15}} & \text{$\Delta$ Cov} & $0.070^{\pm 0.001}$ & $0.009^{\pm 0.000}$ & $\textcolor{gray}{\text{-}0.069}$ & $\text{-}0.000$ & $0.004$ & $0.006$ &   \\
\rotatebox{90}{} & \rotatebox{90}{} & \text{PI-Width} & $\textbf{39.6}^{\pm 0.5}$ & $89.4^{\pm 0.0}$ & $\textcolor{gray}{114.7}$ & $142.2$ & $141.2$ & $142.7$ &   \\
\rotatebox{90}{} & \rotatebox{90}{} & \text{Winkler} & $\textbf{0.56}^{\pm 0.00}$ & $1.56^{\pm 0.00}$ & $\textcolor{gray}{2.21}$ & $2.34$ & $2.37$ & $2.37$ &   \\
\cmidrule{1-10} \cmidrule{2-10}
\multirow[c]{9}{*}{\rotatebox{90}{\text{\text{LSTM}}}} & \multirow[c]{3}{*}{\rotatebox{90}{0.05}} & \text{$\Delta$ Cov} & $\text{-}0.003^{\pm 0.005}$ & $0.004^{\pm 0.000}$ & $\textcolor{gray}{\text{-}0.018}$ & $\text{-}0.001$ & $0.002$ & $0.001$ &   \\
\rotatebox{90}{} & \rotatebox{90}{} & \text{PI-Width} & $\textbf{22.7}^{\pm 0.8}$ & $47.7^{\pm 0.0}$ & $\textcolor{gray}{41.0}$ & $47.3$ & $54.1$ & $54.7$ &   \\
\rotatebox{90}{} & \rotatebox{90}{} & \text{Winkler} & $\textbf{0.38}^{\pm 0.01}$ & $0.87^{\pm 0.00}$ & $\textcolor{gray}{0.90}$ & $0.87$ & $0.96$ & $0.97$ &   \\
\cmidrule{2-10}
\rotatebox{90}{} & \multirow[c]{3}{*}{\rotatebox{90}{0.10}} & \text{$\Delta$ Cov} & $0.001^{\pm 0.006}$ & $0.014^{\pm 0.000}$ & $\text{-}0.018$ & $\text{-}0.001$ & $0.007$ & $0.007$ &   \\
\rotatebox{90}{} & \rotatebox{90}{} & \text{PI-Width} & $\textbf{17.9}^{\pm 0.6}$ & $27.8^{\pm 0.0}$ & $24.6$ & $28.2$ & $31.9$ & $33.0$ &   \\
\rotatebox{90}{} & \rotatebox{90}{} & \text{Winkler} & $\textbf{0.30}^{\pm 0.01}$ & $0.62^{\pm 0.00}$ & $0.64$ & $0.63$ & $0.68$ & $0.70$ &   \\
\cmidrule{2-10}
\rotatebox{90}{} & \multirow[c]{3}{*}{\rotatebox{90}{0.15}} & \text{$\Delta$ Cov} & $0.002^{\pm 0.007}$ & $0.024^{\pm 0.000}$ & $\text{-}0.017$ & $\text{-}0.001$ & $0.010$ & $0.009$ &   \\
\rotatebox{90}{} & \rotatebox{90}{} & \text{PI-Width} & $\textbf{15.0}^{\pm 0.4}$ & $18.1^{\pm 0.0}$ & $16.2$ & $18.6$ & $20.5$ & $21.3$ &   \\
\rotatebox{90}{} & \rotatebox{90}{} & \text{Winkler} & $\textbf{0.26}^{\pm 0.00}$ & $0.48^{\pm 0.00}$ & $0.50$ & $0.50$ & $0.54$ & $0.55$ &   \\
\bottomrule
\end{tabular}
\end{table*}

\setTableCaption{Performance of the evaluated CP algorithms on the \solarOneY dataset for the miscoverage levels $\alpha \in \{0.05, 0.10, 0.15\}$. The column FC specifies the prediction algorithm used for the experiment (Forest: Random Forest, LGBM: LightGBM, Ridge: Ridge Regression, LSTM: LSTM neural network). Bold numbers correspond to the best result for the respective metric in the experiment (PI-Width and Winkler score), given that $\Delta \text{Cov} \ge - 0.25 \alpha$, i.e., the specific algorithm reached at least approximate coverage (the result is grayed out otherwise). The error term represents the standard deviation over repeated runs (results without an error term are from deterministic models).}


\begin{table*}
\centering
\caption{\nextTableCaption}
\label{tab:nsdb-60m-2019-overall-v2}
\vskip 0.15in
\small
\begin{tabular}{lllrrrrrr|r}
\toprule
 &  &  & \rotatebox{90}{\text{\ourmethod}} & \rotatebox{90}{\text{SPCI}} & \rotatebox{90}{\text{EnbPI}} & \rotatebox{90}{\text{NexCP}} & \rotatebox{90}{\text{CopulaCPTS}} & \rotatebox{90}{\text{CP/CF-RNN}} & \rotatebox{90}{\text{AdaptiveCI}} \\
\text{FC} & \text{$\alpha$} &  &  &  &  &  &  &  &  \\
\midrule
\multirow[c]{9}{*}{\rotatebox{90}{\text{Forest}}} & \multirow[c]{3}{*}{\rotatebox{90}{0.05}} & \text{$\Delta$ Cov} & $0.016^{\pm 0.002}$ & $0.029^{\pm 0.000}$ & $\textcolor{gray}{\text{-}0.017}$ & $0.002$ & $0.034$ & $0.035$ &   \\
\rotatebox{90}{} & \rotatebox{90}{} & \text{PI-Width} & $\textbf{35.0}^{\pm 1.2}$ & $147.8^{\pm 0.3}$ & $\textcolor{gray}{133.4}$ & $173.9$ & $241.8$ & $262.9$ &   \\
\rotatebox{90}{} & \rotatebox{90}{} & \text{Winkler} & $\textbf{0.50}^{\pm 0.05}$ & $1.71^{\pm 0.00}$ & $\textcolor{gray}{1.97}$ & $2.18$ & $2.61$ & $2.82$ &   \\
\cmidrule{2-10}
\rotatebox{90}{} & \multirow[c]{3}{*}{\rotatebox{90}{0.10}} & \text{$\Delta$ Cov} & $0.047^{\pm 0.004}$ & $0.045^{\pm 0.000}$ & $\text{-}0.018$ & $0.002$ & $0.056$ & $0.063$ &   \\
\rotatebox{90}{} & \rotatebox{90}{} & \text{PI-Width} & $\textbf{28.6}^{\pm 1.0}$ & $97.1^{\pm 0.2}$ & $98.8$ & $127.8$ & $182.4$ & $204.9$ &   \\
\rotatebox{90}{} & \rotatebox{90}{} & \text{Winkler} & $\textbf{0.40}^{\pm 0.04}$ & $1.26^{\pm 0.00}$ & $1.65$ & $1.84$ & $2.10$ & $2.30$ &   \\
\cmidrule{2-10}
\rotatebox{90}{} & \multirow[c]{3}{*}{\rotatebox{90}{0.15}} & \text{$\Delta$ Cov} & $0.078^{\pm 0.006}$ & $0.052^{\pm 0.000}$ & $\text{-}0.016$ & $0.002$ & $0.070$ & $0.086$ &   \\
\rotatebox{90}{} & \rotatebox{90}{} & \text{PI-Width} & $\textbf{24.6}^{\pm 0.8}$ & $67.7^{\pm 0.1}$ & $73.1$ & $93.6$ & $138.5$ & $161.2$ &   \\
\rotatebox{90}{} & \rotatebox{90}{} & \text{Winkler} & $\textbf{0.35}^{\pm 0.03}$ & $1.00^{\pm 0.00}$ & $1.42$ & $1.59$ & $1.76$ & $1.93$ &   \\
\cmidrule{1-10} \cmidrule{2-10}
\multirow[c]{9}{*}{\rotatebox{90}{\text{LGBM}}} & \multirow[c]{3}{*}{\rotatebox{90}{0.05}} & \text{$\Delta$ Cov} & $\text{-}0.010^{\pm 0.006}$ & $0.029^{\pm 0.000}$ & $\textcolor{gray}{\text{-}0.019}$ & $0.001$ & $0.034$ & $0.036$ & $\text{-}0.000$ \\
\rotatebox{90}{} & \rotatebox{90}{} & \text{PI-Width} & $\textbf{51.0}^{\pm 3.4}$ & $147.8^{\pm 0.2}$ & $\textcolor{gray}{131.5}$ & $165.0$ & $237.9$ & $265.2$ & $62.2$ \\
\rotatebox{90}{} & \rotatebox{90}{} & \text{Winkler} & $\textbf{0.71}^{\pm 0.10}$ & $1.72^{\pm 0.00}$ & $\textcolor{gray}{2.05}$ & $2.16$ & $2.60$ & $2.86$ & $0.78$ \\
\cmidrule{2-10}
\rotatebox{90}{} & \multirow[c]{3}{*}{\rotatebox{90}{0.10}} & \text{$\Delta$ Cov} & $\text{-}0.003^{\pm 0.009}$ & $0.045^{\pm 0.000}$ & $\text{-}0.017$ & $0.002$ & $0.056$ & $0.065$ & $0.000$ \\
\rotatebox{90}{} & \rotatebox{90}{} & \text{PI-Width} & $\textbf{40.7}^{\pm 2.8}$ & $96.5^{\pm 0.2}$ & $93.9$ & $118.0$ & $171.1$ & $196.6$ & $49.5$ \\
\rotatebox{90}{} & \rotatebox{90}{} & \text{Winkler} & $\textbf{0.57}^{\pm 0.08}$ & $1.26^{\pm 0.00}$ & $1.65$ & $1.78$ & $2.02$ & $2.24$ & $0.69$ \\
\cmidrule{2-10}
\rotatebox{90}{} & \multirow[c]{3}{*}{\rotatebox{90}{0.15}} & \text{$\Delta$ Cov} & $0.001^{\pm 0.012}$ & $0.053^{\pm 0.000}$ & $\text{-}0.014$ & $0.002$ & $0.068$ & $0.086$ & $0.000$ \\
\rotatebox{90}{} & \rotatebox{90}{} & \text{PI-Width} & $\textbf{34.5}^{\pm 2.5}$ & $68.0^{\pm 0.1}$ & $68.2$ & $85.7$ & $125.5$ & $149.1$ & $42.5$ \\
\rotatebox{90}{} & \rotatebox{90}{} & \text{Winkler} & $\textbf{0.50}^{\pm 0.07}$ & $1.01^{\pm 0.00}$ & $1.40$ & $1.52$ & $1.66$ & $1.83$ & $0.67$ \\
\cmidrule{1-10} \cmidrule{2-10}
\multirow[c]{9}{*}{\rotatebox{90}{\text{Ridge}}} & \multirow[c]{3}{*}{\rotatebox{90}{0.05}} & \text{$\Delta$ Cov} & $\textcolor{gray}{\text{-}0.025}^{\pm 0.012}$ & $0.018^{\pm 0.000}$ & $\textcolor{gray}{\text{-}0.025}$ & $\text{-}0.004$ & $0.008$ & $0.011$ &   \\
\rotatebox{90}{} & \rotatebox{90}{} & \text{PI-Width} & $\textcolor{gray}{70.3}^{\pm 6.5}$ & $\textbf{151.0}^{\pm 0.2}$ & $\textcolor{gray}{178.0}$ & $196.3$ & $207.9$ & $219.8$ &   \\
\rotatebox{90}{} & \rotatebox{90}{} & \text{Winkler} & $\textcolor{gray}{1.15}^{\pm 0.18}$ & $\textbf{1.78}^{\pm 0.00}$ & $\textcolor{gray}{2.26}$ & $2.29$ & $2.35$ & $2.49$ &   \\
\cmidrule{2-10}
\rotatebox{90}{} & \multirow[c]{3}{*}{\rotatebox{90}{0.10}} & \text{$\Delta$ Cov} & $\text{-}0.011^{\pm 0.016}$ & $0.031^{\pm 0.000}$ & $\textcolor{gray}{\text{-}0.031}$ & $\text{-}0.006$ & $\text{-}0.010$ & $\text{-}0.015$ &   \\
\rotatebox{90}{} & \rotatebox{90}{} & \text{PI-Width} & $\textbf{59.9}^{\pm 5.6}$ & $112.8^{\pm 0.1}$ & $\textcolor{gray}{158.4}$ & $171.7$ & $171.7$ & $172.0$ &   \\
\rotatebox{90}{} & \rotatebox{90}{} & \text{Winkler} & $\textbf{0.92}^{\pm 0.12}$ & $1.40^{\pm 0.00}$ & $\textcolor{gray}{2.06}$ & $2.09$ & $2.12$ & $2.17$ &   \\
\cmidrule{2-10}
\rotatebox{90}{} & \multirow[c]{3}{*}{\rotatebox{90}{0.15}} & \text{$\Delta$ Cov} & $0.000^{\pm 0.020}$ & $0.035^{\pm 0.001}$ & $\text{-}0.034$ & $\text{-}0.008$ & $\text{-}0.024$ & $\textcolor{gray}{\text{-}0.040}$ &   \\
\rotatebox{90}{} & \rotatebox{90}{} & \text{PI-Width} & $\textbf{52.9}^{\pm 4.9}$ & $92.5^{\pm 0.1}$ & $144.1$ & $154.6$ & $150.2$ & $\textcolor{gray}{146.6}$ &   \\
\rotatebox{90}{} & \rotatebox{90}{} & \text{Winkler} & $\textbf{0.81}^{\pm 0.10}$ & $1.21^{\pm 0.00}$ & $1.92$ & $1.96$ & $1.98$ & $\textcolor{gray}{2.02}$ &   \\
\cmidrule{1-10} \cmidrule{2-10}
\multirow[c]{9}{*}{\rotatebox{90}{\text{\text{LSTM}}}} & \multirow[c]{3}{*}{\rotatebox{90}{0.05}} & \text{$\Delta$ Cov} & $0.019^{\pm 0.003}$ & $0.006^{\pm 0.000}$ & $\textcolor{gray}{\text{-}0.018}$ & $\text{-}0.001$ & $0.010$ & $0.013$ &   \\
\rotatebox{90}{} & \rotatebox{90}{} & \text{PI-Width} & $\textbf{21.4}^{\pm 0.7}$ & $37.0^{\pm 0.0}$ & $\textcolor{gray}{29.5}$ & $33.6$ & $39.4$ & $42.2$ &   \\
\rotatebox{90}{} & \rotatebox{90}{} & \text{Winkler} & $\textbf{0.27}^{\pm 0.01}$ & $0.58^{\pm 0.00}$ & $\textcolor{gray}{0.60}$ & $0.57$ & $0.59$ & $0.61$ &   \\
\cmidrule{2-10}
\rotatebox{90}{} & \multirow[c]{3}{*}{\rotatebox{90}{0.10}} & \text{$\Delta$ Cov} & $0.028^{\pm 0.010}$ & $0.018^{\pm 0.000}$ & $\text{-}0.018$ & $\text{-}0.001$ & $0.018$ & $0.025$ &   \\
\rotatebox{90}{} & \rotatebox{90}{} & \text{PI-Width} & $\textbf{16.0}^{\pm 0.6}$ & $22.5^{\pm 0.0}$ & $17.4$ & $19.5$ & $23.1$ & $25.0$ &   \\
\rotatebox{90}{} & \rotatebox{90}{} & \text{Winkler} & $\textbf{0.22}^{\pm 0.01}$ & $0.41^{\pm 0.00}$ & $0.42$ & $0.41$ & $0.43$ & $0.43$ &   \\
\cmidrule{2-10}
\rotatebox{90}{} & \multirow[c]{3}{*}{\rotatebox{90}{0.15}} & \text{$\Delta$ Cov} & $0.029^{\pm 0.017}$ & $0.032^{\pm 0.000}$ & $\text{-}0.014$ & $0.001$ & $0.030$ & $0.040$ &   \\
\rotatebox{90}{} & \rotatebox{90}{} & \text{PI-Width} & $13.1^{\pm 0.5}$ & $15.3^{\pm 0.0}$ & $\textbf{11.1}$ & $12.6$ & $15.2$ & $16.9$ &   \\
\rotatebox{90}{} & \rotatebox{90}{} & \text{Winkler} & $\textbf{0.19}^{\pm 0.00}$ & $0.32^{\pm 0.00}$ & $0.33$ & $0.33$ & $0.33$ & $0.34$ &   \\
\bottomrule
\end{tabular}
\end{table*}

\setTableCaption{Performance of the evaluated CP algorithms on the \solarSmall dataset for the miscoverage levels $\alpha \in \{0.05, 0.10, 0.15\}$. The column FC specifies the prediction algorithm used for the experiment (Forest: Random Forest, LGBM: LightGBM, Ridge: Ridge Regression). Bold numbers correspond to the best result for the respective metric in the experiment (PI-Width and Winkler score), given that $\Delta \text{Cov} \ge - 0.25 \alpha$, i.e., the specific algorithm reached at least approximate coverage (the result is grayed out otherwise). The error term represents the standard deviation over repeated runs (results without an error term are from deterministic models).}


\begin{table*}
\centering
\caption{\nextTableCaption}
\label{tab:solar-overall-v2}
\vskip 0.15in
\small
\begin{tabular}{lllrrrrrr|r}
\toprule
 &  &  & \rotatebox{90}{\text{\ourmethod}} & \rotatebox{90}{\text{SPCI}} & \rotatebox{90}{\text{EnbPI}} & \rotatebox{90}{\text{NexCP}} & \rotatebox{90}{\text{CopulaCPTS}} & \rotatebox{90}{\text{CP/CF-RNN}} & \rotatebox{90}{\text{AdaptiveCI}} \\
\text{FC} & \text{$\alpha$} &  &  &  &  &  &  &  &  \\
\midrule
\multirow[c]{9}{*}{\rotatebox{90}{\text{Forest}}} & \multirow[c]{3}{*}{\rotatebox{90}{0.05}} & \text{$\Delta$ Cov} & $\text{-}0.002^{\pm 0.005}$ & $\textcolor{gray}{\text{-}0.074}^{\pm 0.001}$ & $\textcolor{gray}{\text{-}0.022}$ & $\textcolor{gray}{\text{-}0.013}$ & $\textcolor{gray}{\text{-}0.022}$ & $\textcolor{gray}{\text{-}0.028}$ &   \\
\rotatebox{90}{} & \rotatebox{90}{} & \text{PI-Width} & $\textbf{47.5}^{\pm 5.0}$ & $\textcolor{gray}{57.9}^{\pm 0.3}$ & $\textcolor{gray}{110.7}$ & $\textcolor{gray}{160.2}$ & $\textcolor{gray}{162.0}$ & $\textcolor{gray}{155.0}$ &   \\
\rotatebox{90}{} & \rotatebox{90}{} & \text{Winkler} & $\textbf{1.29}^{\pm 0.11}$ & $\textcolor{gray}{2.62}^{\pm 0.02}$ & $\textcolor{gray}{2.92}$ & $\textcolor{gray}{3.70}$ & $\textcolor{gray}{3.95}$ & $\textcolor{gray}{4.04}$ &   \\
\cmidrule{2-10}
\rotatebox{90}{} & \multirow[c]{3}{*}{\rotatebox{90}{0.10}} & \text{$\Delta$ Cov} & $0.008^{\pm 0.006}$ & $\textcolor{gray}{\text{-}0.064}^{\pm 0.002}$ & $\text{-}0.022$ & $\text{-}0.021$ & $\textcolor{gray}{\text{-}0.027}$ & $\textcolor{gray}{\text{-}0.025}$ &   \\
\rotatebox{90}{} & \rotatebox{90}{} & \text{PI-Width} & $\textbf{38.4}^{\pm 3.4}$ & $\textcolor{gray}{38.8}^{\pm 0.3}$ & $86.0$ & $122.9$ & $\textcolor{gray}{110.4}$ & $\textcolor{gray}{111.4}$ &   \\
\rotatebox{90}{} & \rotatebox{90}{} & \text{Winkler} & $\textbf{1.09}^{\pm 0.08}$ & $\textcolor{gray}{1.82}^{\pm 0.01}$ & $2.54$ & $3.28$ & $\textcolor{gray}{3.47}$ & $\textcolor{gray}{3.48}$ &   \\
\cmidrule{2-10}
\rotatebox{90}{} & \multirow[c]{3}{*}{\rotatebox{90}{0.15}} & \text{$\Delta$ Cov} & $0.020^{\pm 0.010}$ & $\textcolor{gray}{\text{-}0.052}^{\pm 0.002}$ & $\text{-}0.028$ & $\text{-}0.019$ & $\text{-}0.020$ & $\text{-}0.017$ &   \\
\rotatebox{90}{} & \rotatebox{90}{} & \text{PI-Width} & $\textbf{32.6}^{\pm 2.8}$ & $\textcolor{gray}{26.7}^{\pm 0.3}$ & $64.8$ & $91.0$ & $77.5$ & $79.6$ &   \\
\rotatebox{90}{} & \rotatebox{90}{} & \text{Winkler} & $\textbf{0.99}^{\pm 0.07}$ & $\textcolor{gray}{1.45}^{\pm 0.02}$ & $2.40$ & $2.91$ & $2.99$ & $2.98$ &   \\
\cmidrule{1-10} \cmidrule{2-10}
\multirow[c]{9}{*}{\rotatebox{90}{\text{LGBM}}} & \multirow[c]{3}{*}{\rotatebox{90}{0.05}} & \text{$\Delta$ Cov} & $0.002^{\pm 0.008}$ & $\textcolor{gray}{\text{-}0.063}^{\pm 0.001}$ & $\textcolor{gray}{\text{-}0.024}$ & $\textcolor{gray}{\text{-}0.014}$ & $\textcolor{gray}{\text{-}0.018}$ & $\textcolor{gray}{\text{-}0.019}$ & $\text{-}0.000$ \\
\rotatebox{90}{} & \rotatebox{90}{} & \text{PI-Width} & $\textbf{59.7}^{\pm 3.4}$ & $\textcolor{gray}{56.3}^{\pm 0.3}$ & $\textcolor{gray}{111.7}$ & $\textcolor{gray}{161.7}$ & $\textcolor{gray}{164.5}$ & $\textcolor{gray}{162.1}$ & $61.8$ \\
\rotatebox{90}{} & \rotatebox{90}{} & \text{Winkler} & $1.49^{\pm 0.15}$ & $\textcolor{gray}{2.66}^{\pm 0.02}$ & $\textcolor{gray}{3.10}$ & $\textcolor{gray}{3.87}$ & $\textcolor{gray}{4.13}$ & $\textcolor{gray}{4.18}$ & $\textbf{1.45}$ \\
\cmidrule{2-10}
\rotatebox{90}{} & \multirow[c]{3}{*}{\rotatebox{90}{0.10}} & \text{$\Delta$ Cov} & $0.007^{\pm 0.012}$ & $\textcolor{gray}{\text{-}0.052}^{\pm 0.002}$ & $\text{-}0.022$ & $\text{-}0.020$ & $\text{-}0.022$ & $\text{-}0.021$ & $\text{-}0.004$ \\
\rotatebox{90}{} & \rotatebox{90}{} & \text{PI-Width} & $\textbf{48.2}^{\pm 1.7}$ & $\textcolor{gray}{37.5}^{\pm 0.3}$ & $84.3$ & $119.9$ & $107.6$ & $106.9$ & $49.2$ \\
\rotatebox{90}{} & \rotatebox{90}{} & \text{Winkler} & $\textbf{1.24}^{\pm 0.08}$ & $\textcolor{gray}{1.84}^{\pm 0.01}$ & $2.67$ & $3.35$ & $3.52$ & $3.53$ & $1.26$ \\
\cmidrule{2-10}
\rotatebox{90}{} & \multirow[c]{3}{*}{\rotatebox{90}{0.15}} & \text{$\Delta$ Cov} & $0.005^{\pm 0.016}$ & $\textcolor{gray}{\text{-}0.043}^{\pm 0.002}$ & $\text{-}0.024$ & $\text{-}0.019$ & $\text{-}0.013$ & $\text{-}0.011$ & $\text{-}0.007$ \\
\rotatebox{90}{} & \rotatebox{90}{} & \text{PI-Width} & $\textbf{41.4}^{\pm 1.3}$ & $\textcolor{gray}{26.9}^{\pm 0.2}$ & $60.8$ & $85.1$ & $75.8$ & $77.5$ & $44.1$ \\
\rotatebox{90}{} & \rotatebox{90}{} & \text{Winkler} & $\textbf{1.11}^{\pm 0.07}$ & $\textcolor{gray}{1.47}^{\pm 0.01}$ & $2.48$ & $2.93$ & $2.98$ & $2.97$ & $1.22$ \\
\cmidrule{1-10} \cmidrule{2-10}
\multirow[c]{9}{*}{\rotatebox{90}{\text{Ridge}}} & \multirow[c]{3}{*}{\rotatebox{90}{0.05}} & \text{$\Delta$ Cov} & $\textcolor{gray}{\text{-}0.017}^{\pm 0.014}$ & $\textcolor{gray}{\text{-}0.064}^{\pm 0.001}$ & $\textcolor{gray}{\text{-}0.021}$ & $0.004$ & $0.009$ & $\text{-}0.002$ &   \\
\rotatebox{90}{} & \rotatebox{90}{} & \text{PI-Width} & $\textcolor{gray}{92.7}^{\pm 32.4}$ & $\textcolor{gray}{67.2}^{\pm 0.3}$ & $\textcolor{gray}{110.1}$ & $144.8$ & $156.1$ & $\textbf{133.6}$ &   \\
\rotatebox{90}{} & \rotatebox{90}{} & \text{Winkler} & $\textcolor{gray}{1.99}^{\pm 0.69}$ & $\textcolor{gray}{2.49}^{\pm 0.02}$ & $\textcolor{gray}{3.07}$ & $3.82$ & $3.78$ & $\textbf{3.75}$ &   \\
\cmidrule{2-10}
\rotatebox{90}{} & \multirow[c]{3}{*}{\rotatebox{90}{0.10}} & \text{$\Delta$ Cov} & $\text{-}0.016^{\pm 0.022}$ & $\textcolor{gray}{\text{-}0.056}^{\pm 0.002}$ & $\text{-}0.020$ & $0.007$ & $0.014$ & $\text{-}0.003$ &   \\
\rotatebox{90}{} & \rotatebox{90}{} & \text{PI-Width} & $78.2^{\pm 24.5}$ & $\textcolor{gray}{51.8}^{\pm 0.3}$ & $84.9$ & $85.2$ & $88.6$ & $\textbf{78.1}$ &   \\
\rotatebox{90}{} & \rotatebox{90}{} & \text{Winkler} & $\textbf{1.78}^{\pm 0.59}$ & $\textcolor{gray}{1.91}^{\pm 0.01}$ & $2.66$ & $2.85$ & $2.83$ & $2.81$ &   \\
\cmidrule{2-10}
\rotatebox{90}{} & \multirow[c]{3}{*}{\rotatebox{90}{0.15}} & \text{$\Delta$ Cov} & $\text{-}0.009^{\pm 0.021}$ & $\textcolor{gray}{\text{-}0.039}^{\pm 0.002}$ & $\text{-}0.017$ & $0.003$ & $0.014$ & $0.003$ &   \\
\rotatebox{90}{} & \rotatebox{90}{} & \text{PI-Width} & $\textbf{67.4}^{\pm 19.0}$ & $\textcolor{gray}{43.2}^{\pm 0.2}$ & $71.6$ & $70.9$ & $71.9$ & $70.7$ &   \\
\rotatebox{90}{} & \rotatebox{90}{} & \text{Winkler} & $\textbf{1.61}^{\pm 0.51}$ & $\textcolor{gray}{1.59}^{\pm 0.00}$ & $2.29$ & $2.34$ & $2.34$ & $2.33$ &   \\
\bottomrule
\end{tabular}
\end{table*}

\setTableCaption{Performance of the evaluated CP algorithms on the \airTen dataset for the miscoverage levels $\alpha \in \{0.05, 0.10, 0.15\}$. The column FC specifies the prediction algorithm used for the experiment (Forest: Random Forest, LGBM: LightGBM, Ridge: Ridge Regression, LSTM: LSTM neural network). Bold numbers correspond to the best result for the respective metric in the experiment (PI-Width and Winkler score), given that $\Delta \text{Cov} \ge - 0.25 \alpha$, i.e., the specific algorithm reached at least approximate coverage (the result is grayed out otherwise). The error term represents the standard deviation over repeated runs (results without an error term are from deterministic models).}


\begin{table*}
\centering
\caption{\nextTableCaption}
\label{tab:air-10-overall-v2}
\vskip 0.15in
\small
\begin{tabular}{lllrrrrrr|r}
\toprule
 &  &  & \rotatebox{90}{\text{\ourmethod}} & \rotatebox{90}{\text{SPCI}} & \rotatebox{90}{\text{EnbPI}} & \rotatebox{90}{\text{NexCP}} & \rotatebox{90}{\text{CopulaCPTS}} & \rotatebox{90}{\text{CP/CF-RNN}} & \rotatebox{90}{\text{AdaptiveCI}} \\
\text{FC} & \text{$\alpha$} &  &  &  &  &  &  &  &  \\
\midrule
\multirow[c]{9}{*}{\rotatebox{90}{\text{Forest}}} & \multirow[c]{3}{*}{\rotatebox{90}{0.05}} & \text{$\Delta$ Cov} & $0.006^{\pm 0.013}$ & $\text{-}0.001^{\pm 0.000}$ & $\textcolor{gray}{\text{-}0.054}$ & $\text{-}0.005$ & $\textcolor{gray}{\text{-}0.016}$ & $\textcolor{gray}{\text{-}0.027}$ &   \\
\rotatebox{90}{} & \rotatebox{90}{} & \text{PI-Width} & $\textbf{116.7}^{\pm 14.4}$ & $152.7^{\pm 0.1}$ & $\textcolor{gray}{242.9}$ & $321.5$ & $\textcolor{gray}{322.5}$ & $\textcolor{gray}{297.1}$ &   \\
\rotatebox{90}{} & \rotatebox{90}{} & \text{Winkler} & $\textbf{1.94}^{\pm 0.10}$ & $2.86^{\pm 0.00}$ & $\textcolor{gray}{4.97}$ & $4.82$ & $\textcolor{gray}{6.47}$ & $\textcolor{gray}{6.61}$ &   \\
\cmidrule{2-10}
\rotatebox{90}{} & \multirow[c]{3}{*}{\rotatebox{90}{0.10}} & \text{$\Delta$ Cov} & $0.028^{\pm 0.019}$ & $0.008^{\pm 0.000}$ & $\textcolor{gray}{\text{-}0.066}$ & $\text{-}0.004$ & $\text{-}0.019$ & $\textcolor{gray}{\text{-}0.033}$ &   \\
\rotatebox{90}{} & \rotatebox{90}{} & \text{PI-Width} & $\textbf{93.9}^{\pm 11.1}$ & $118.5^{\pm 0.1}$ & $\textcolor{gray}{202.8}$ & $263.5$ & $243.1$ & $\textcolor{gray}{229.8}$ &   \\
\rotatebox{90}{} & \rotatebox{90}{} & \text{Winkler} & $\textbf{1.50}^{\pm 0.09}$ & $2.23^{\pm 0.00}$ & $\textcolor{gray}{4.16}$ & $4.03$ & $4.94$ & $\textcolor{gray}{4.98}$ &   \\
\cmidrule{2-10}
\rotatebox{90}{} & \multirow[c]{3}{*}{\rotatebox{90}{0.15}} & \text{$\Delta$ Cov} & $0.050^{\pm 0.025}$ & $0.018^{\pm 0.000}$ & $\textcolor{gray}{\text{-}0.073}$ & $\text{-}0.002$ & $\text{-}0.021$ & $\textcolor{gray}{\text{-}0.039}$ &   \\
\rotatebox{90}{} & \rotatebox{90}{} & \text{PI-Width} & $\textbf{80.5}^{\pm 9.4}$ & $99.8^{\pm 0.1}$ & $\textcolor{gray}{175.7}$ & $229.4$ & $207.1$ & $\textcolor{gray}{198.2}$ &   \\
\rotatebox{90}{} & \rotatebox{90}{} & \text{Winkler} & $\textbf{1.28}^{\pm 0.08}$ & $1.92^{\pm 0.00}$ & $\textcolor{gray}{3.72}$ & $3.61$ & $4.18$ & $\textcolor{gray}{4.20}$ &   \\
\cmidrule{1-10} \cmidrule{2-10}
\multirow[c]{9}{*}{\rotatebox{90}{\text{LGBM}}} & \multirow[c]{3}{*}{\rotatebox{90}{0.05}} & \text{$\Delta$ Cov} & $\text{-}0.000^{\pm 0.011}$ & $0.008^{\pm 0.000}$ & $\textcolor{gray}{\text{-}0.047}$ & $\text{-}0.005$ & $\textcolor{gray}{\text{-}0.014}$ & $\textcolor{gray}{\text{-}0.022}$ & $\text{-}0.000$ \\
\rotatebox{90}{} & \rotatebox{90}{} & \text{PI-Width} & $\textbf{106.3}^{\pm 9.9}$ & $146.1^{\pm 0.1}$ & $\textcolor{gray}{219.5}$ & $285.7$ & $\textcolor{gray}{281.1}$ & $\textcolor{gray}{264.6}$ & $228.0$ \\
\rotatebox{90}{} & \rotatebox{90}{} & \text{Winkler} & $\textbf{1.87}^{\pm 0.07}$ & $2.49^{\pm 0.00}$ & $\textcolor{gray}{4.56}$ & $4.46$ & $\textcolor{gray}{5.72}$ & $\textcolor{gray}{5.79}$ & $3.61$ \\
\cmidrule{2-10}
\rotatebox{90}{} & \multirow[c]{3}{*}{\rotatebox{90}{0.10}} & \text{$\Delta$ Cov} & $0.017^{\pm 0.016}$ & $0.023^{\pm 0.000}$ & $\textcolor{gray}{\text{-}0.057}$ & $\text{-}0.004$ & $\text{-}0.017$ & $\textcolor{gray}{\text{-}0.028}$ & $\text{-}0.001$ \\
\rotatebox{90}{} & \rotatebox{90}{} & \text{PI-Width} & $\textbf{85.6}^{\pm 7.4}$ & $113.2^{\pm 0.1}$ & $\textcolor{gray}{178.3}$ & $224.8$ & $206.7$ & $\textcolor{gray}{196.5}$ & $186.4$ \\
\rotatebox{90}{} & \rotatebox{90}{} & \text{Winkler} & $\textbf{1.45}^{\pm 0.06}$ & $1.94^{\pm 0.00}$ & $\textcolor{gray}{3.69}$ & $3.64$ & $4.33$ & $\textcolor{gray}{4.36}$ & $3.00$ \\
\cmidrule{2-10}
\rotatebox{90}{} & \multirow[c]{3}{*}{\rotatebox{90}{0.15}} & \text{$\Delta$ Cov} & $0.033^{\pm 0.019}$ & $0.038^{\pm 0.000}$ & $\textcolor{gray}{\text{-}0.061}$ & $\text{-}0.004$ & $\text{-}0.017$ & $\text{-}0.029$ & $\text{-}0.001$ \\
\rotatebox{90}{} & \rotatebox{90}{} & \text{PI-Width} & $\textbf{73.4}^{\pm 6.2}$ & $94.9^{\pm 0.1}$ & $\textcolor{gray}{151.4}$ & $188.9$ & $168.2$ & $161.9$ & $161.1$ \\
\rotatebox{90}{} & \rotatebox{90}{} & \text{Winkler} & $\textbf{1.24}^{\pm 0.06}$ & $1.66^{\pm 0.00}$ & $\textcolor{gray}{3.23}$ & $3.20$ & $3.63$ & $3.65$ & $2.71$ \\
\cmidrule{1-10} \cmidrule{2-10}
\multirow[c]{9}{*}{\rotatebox{90}{\text{Ridge}}} & \multirow[c]{3}{*}{\rotatebox{90}{0.05}} & \text{$\Delta$ Cov} & $0.001^{\pm 0.003}$ & $0.010^{\pm 0.000}$ & $\textcolor{gray}{\text{-}0.037}$ & $\text{-}0.003$ & $0.005$ & $0.006$ &   \\
\rotatebox{90}{} & \rotatebox{90}{} & \text{PI-Width} & $\textbf{100.6}^{\pm 6.2}$ & $120.2^{\pm 0.1}$ & $\textcolor{gray}{152.6}$ & $202.3$ & $215.7$ & $219.3$ &   \\
\rotatebox{90}{} & \rotatebox{90}{} & \text{Winkler} & $\textbf{1.70}^{\pm 0.08}$ & $1.93^{\pm 0.00}$ & $\textcolor{gray}{3.32}$ & $3.42$ & $3.96$ & $3.98$ &   \\
\cmidrule{2-10}
\rotatebox{90}{} & \multirow[c]{3}{*}{\rotatebox{90}{0.10}} & \text{$\Delta$ Cov} & $0.010^{\pm 0.007}$ & $0.024^{\pm 0.000}$ & $\textcolor{gray}{\text{-}0.045}$ & $\text{-}0.002$ & $0.010$ & $0.012$ &   \\
\rotatebox{90}{} & \rotatebox{90}{} & \text{PI-Width} & $\textbf{79.9}^{\pm 4.7}$ & $93.9^{\pm 0.1}$ & $\textcolor{gray}{120.0}$ & $153.3$ & $153.7$ & $155.3$ &   \\
\rotatebox{90}{} & \rotatebox{90}{} & \text{Winkler} & $\textbf{1.35}^{\pm 0.06}$ & $1.52^{\pm 0.00}$ & $\textcolor{gray}{2.60}$ & $2.68$ & $2.95$ & $2.96$ &   \\
\cmidrule{2-10}
\rotatebox{90}{} & \multirow[c]{3}{*}{\rotatebox{90}{0.15}} & \text{$\Delta$ Cov} & $0.016^{\pm 0.011}$ & $0.038^{\pm 0.001}$ & $\textcolor{gray}{\text{-}0.049}$ & $\text{-}0.003$ & $0.015$ & $0.018$ &   \\
\rotatebox{90}{} & \rotatebox{90}{} & \text{PI-Width} & $\textbf{68.1}^{\pm 4.1}$ & $79.4^{\pm 0.1}$ & $\textcolor{gray}{100.5}$ & $126.9$ & $125.7$ & $127.0$ &   \\
\rotatebox{90}{} & \rotatebox{90}{} & \text{Winkler} & $\textbf{1.17}^{\pm 0.05}$ & $1.31^{\pm 0.00}$ & $\textcolor{gray}{2.23}$ & $2.31$ & $2.46$ & $2.46$ &   \\
\cmidrule{1-10} \cmidrule{2-10}
\multirow[c]{9}{*}{\rotatebox{90}{\text{\text{LSTM}}}} & \multirow[c]{3}{*}{\rotatebox{90}{0.05}} & \text{$\Delta$ Cov} & $\text{-}0.001^{\pm 0.001}$ & $0.003^{\pm 0.000}$ & $\textcolor{gray}{\text{-}0.021}$ & $\text{-}0.002$ & $\text{-}0.002$ & $\text{-}0.001$ &   \\
\rotatebox{90}{} & \rotatebox{90}{} & \text{PI-Width} & $90.7^{\pm 1.9}$ & $\textbf{86.1}^{\pm 0.0}$ & $\textcolor{gray}{80.8}$ & $88.8$ & $86.8$ & $88.1$ &   \\
\rotatebox{90}{} & \rotatebox{90}{} & \text{Winkler} & $1.83^{\pm 0.03}$ & $\textbf{1.63}^{\pm 0.00}$ & $\textcolor{gray}{1.81}$ & $1.77$ & $1.86$ & $1.86$ &   \\
\cmidrule{2-10}
\rotatebox{90}{} & \multirow[c]{3}{*}{\rotatebox{90}{0.10}} & \text{$\Delta$ Cov} & $\text{-}0.002^{\pm 0.005}$ & $0.010^{\pm 0.000}$ & $\textcolor{gray}{\text{-}0.025}$ & $\text{-}0.002$ & $0.001$ & $0.004$ &   \\
\rotatebox{90}{} & \rotatebox{90}{} & \text{PI-Width} & $62.7^{\pm 1.5}$ & $62.2^{\pm 0.0}$ & $\textcolor{gray}{58.1}$ & $62.4$ & $\textbf{61.8}$ & $63.0$ &   \\
\rotatebox{90}{} & \rotatebox{90}{} & \text{Winkler} & $1.33^{\pm 0.01}$ & $\textbf{1.21}^{\pm 0.00}$ & $\textcolor{gray}{1.32}$ & $1.29$ & $1.34$ & $1.34$ &   \\
\cmidrule{2-10}
\rotatebox{90}{} & \multirow[c]{3}{*}{\rotatebox{90}{0.15}} & \text{$\Delta$ Cov} & $\text{-}0.002^{\pm 0.010}$ & $0.017^{\pm 0.000}$ & $\text{-}0.028$ & $\text{-}0.002$ & $0.005$ & $0.009$ &   \\
\rotatebox{90}{} & \rotatebox{90}{} & \text{PI-Width} & $49.4^{\pm 1.7}$ & $50.2^{\pm 0.0}$ & $\textbf{46.5}$ & $49.6$ & $49.6$ & $50.8$ &   \\
\rotatebox{90}{} & \rotatebox{90}{} & \text{Winkler} & $1.09^{\pm 0.01}$ & $\textbf{1.01}^{\pm 0.00}$ & $1.08$ & $1.07$ & $1.10$ & $1.10$ &   \\
\bottomrule
\end{tabular}
\end{table*}

\setTableCaption{Performance of the evaluated CP algorithms on the \airTwenty dataset for the miscoverage levels $\alpha \in \{0.05, 0.10, 0.15\}$. The column FC specifies the prediction algorithm used for the experiment (Forest: Random Forest, LGBM: LightGBM, Ridge: Ridge Regression, LSTM: LSTM neural network). Bold numbers correspond to the best result for the respective metric in the experiment (PI-Width and Winkler score), given that $\Delta \text{Cov} \ge - 0.25 \alpha$, i.e., the specific algorithm reached at least approximate coverage (the result is grayed out otherwise). The error term represents the standard deviation over repeated runs (results without an error term are from deterministic models).}


\begin{table*}
\centering
\caption{\nextTableCaption}
\label{tab:air-25-overall-v2}
\vskip 0.15in
\small
\begin{tabular}{lllrrrrrr|r}
\toprule
 &  &  & \rotatebox{90}{\text{\ourmethod}} & \rotatebox{90}{\text{SPCI}} & \rotatebox{90}{\text{EnbPI}} & \rotatebox{90}{\text{NexCP}} & \rotatebox{90}{\text{CopulaCPTS}} & \rotatebox{90}{\text{CP/CF-RNN}} & \rotatebox{90}{\text{AdaptiveCI}} \\
\text{FC} & \text{$\alpha$} &  &  &  &  &  &  &  &  \\
\midrule
\multirow[c]{9}{*}{\rotatebox{90}{\text{Forest}}} & \multirow[c]{3}{*}{\rotatebox{90}{0.05}} & \text{$\Delta$ Cov} & $\textcolor{gray}{\text{-}0.033}^{\pm 0.013}$ & $\textcolor{gray}{\text{-}0.015}^{\pm 0.000}$ & $\textcolor{gray}{\text{-}0.065}$ & $\text{-}0.009$ & $\textcolor{gray}{\text{-}0.018}$ & $\textcolor{gray}{\text{-}0.031}$ &   \\
\rotatebox{90}{} & \rotatebox{90}{} & \text{PI-Width} & $\textcolor{gray}{58.7}^{\pm 6.9}$ & $\textcolor{gray}{102.6}^{\pm 0.1}$ & $\textcolor{gray}{211.3}$ & $\textbf{283.9}$ & $\textcolor{gray}{271.4}$ & $\textcolor{gray}{249.8}$ &   \\
\rotatebox{90}{} & \rotatebox{90}{} & \text{Winkler} & $\textcolor{gray}{1.51}^{\pm 0.04}$ & $\textcolor{gray}{2.67}^{\pm 0.00}$ & $\textcolor{gray}{5.01}$ & $\textbf{4.71}$ & $\textcolor{gray}{6.47}$ & $\textcolor{gray}{6.59}$ &   \\
\cmidrule{2-10}
\rotatebox{90}{} & \multirow[c]{3}{*}{\rotatebox{90}{0.10}} & \text{$\Delta$ Cov} & $\text{-}0.024^{\pm 0.017}$ & $\text{-}0.009^{\pm 0.000}$ & $\textcolor{gray}{\text{-}0.079}$ & $\text{-}0.007$ & $\text{-}0.025$ & $\textcolor{gray}{\text{-}0.042}$ &   \\
\rotatebox{90}{} & \rotatebox{90}{} & \text{PI-Width} & $\textbf{48.1}^{\pm 5.6}$ & $81.5^{\pm 0.0}$ & $\textcolor{gray}{177.3}$ & $235.6$ & $212.6$ & $\textcolor{gray}{203.5}$ &   \\
\rotatebox{90}{} & \rotatebox{90}{} & \text{Winkler} & $\textbf{1.12}^{\pm 0.05}$ & $2.02^{\pm 0.00}$ & $\textcolor{gray}{4.31}$ & $4.05$ & $4.94$ & $\textcolor{gray}{4.98}$ &   \\
\cmidrule{2-10}
\rotatebox{90}{} & \multirow[c]{3}{*}{\rotatebox{90}{0.15}} & \text{$\Delta$ Cov} & $\text{-}0.014^{\pm 0.019}$ & $\text{-}0.000^{\pm 0.001}$ & $\textcolor{gray}{\text{-}0.084}$ & $\text{-}0.005$ & $\text{-}0.029$ & $\textcolor{gray}{\text{-}0.045}$ &   \\
\rotatebox{90}{} & \rotatebox{90}{} & \text{PI-Width} & $\textbf{41.4}^{\pm 4.9}$ & $70.8^{\pm 0.0}$ & $\textcolor{gray}{153.0}$ & $206.9$ & $185.5$ & $\textcolor{gray}{179.0}$ &   \\
\rotatebox{90}{} & \rotatebox{90}{} & \text{Winkler} & $\textbf{0.94}^{\pm 0.05}$ & $1.71^{\pm 0.00}$ & $\textcolor{gray}{3.88}$ & $3.67$ & $4.23$ & $\textcolor{gray}{4.25}$ &   \\
\cmidrule{1-10} \cmidrule{2-10}
\multirow[c]{9}{*}{\rotatebox{90}{\text{LGBM}}} & \multirow[c]{3}{*}{\rotatebox{90}{0.05}} & \text{$\Delta$ Cov} & $\textcolor{gray}{\text{-}0.031}^{\pm 0.014}$ & $\text{-}0.007^{\pm 0.000}$ & $\textcolor{gray}{\text{-}0.054}$ & $\text{-}0.006$ & $\textcolor{gray}{\text{-}0.020}$ & $\textcolor{gray}{\text{-}0.032}$ & $\text{-}0.001$ \\
\rotatebox{90}{} & \rotatebox{90}{} & \text{PI-Width} & $\textcolor{gray}{57.2}^{\pm 7.8}$ & $\textbf{93.5}^{\pm 0.1}$ & $\textcolor{gray}{176.4}$ & $239.5$ & $\textcolor{gray}{212.7}$ & $\textcolor{gray}{196.0}$ & $208.8$ \\
\rotatebox{90}{} & \rotatebox{90}{} & \text{Winkler} & $\textcolor{gray}{1.48}^{\pm 0.05}$ & $\textbf{2.45}^{\pm 0.00}$ & $\textcolor{gray}{4.30}$ & $4.35$ & $\textcolor{gray}{5.47}$ & $\textcolor{gray}{5.58}$ & $3.99$ \\
\cmidrule{2-10}
\rotatebox{90}{} & \multirow[c]{3}{*}{\rotatebox{90}{0.10}} & \text{$\Delta$ Cov} & $\text{-}0.021^{\pm 0.019}$ & $0.002^{\pm 0.000}$ & $\textcolor{gray}{\text{-}0.063}$ & $\text{-}0.007$ & $\textcolor{gray}{\text{-}0.027}$ & $\textcolor{gray}{\text{-}0.042}$ & $\text{-}0.001$ \\
\rotatebox{90}{} & \rotatebox{90}{} & \text{PI-Width} & $\textbf{46.8}^{\pm 6.0}$ & $73.3^{\pm 0.0}$ & $\textcolor{gray}{142.1}$ & $182.2$ & $\textcolor{gray}{154.5}$ & $\textcolor{gray}{143.9}$ & $163.3$ \\
\rotatebox{90}{} & \rotatebox{90}{} & \text{Winkler} & $\textbf{1.10}^{\pm 0.05}$ & $1.83^{\pm 0.00}$ & $\textcolor{gray}{3.54}$ & $3.55$ & $\textcolor{gray}{4.09}$ & $\textcolor{gray}{4.14}$ & $3.35$ \\
\cmidrule{2-10}
\rotatebox{90}{} & \multirow[c]{3}{*}{\rotatebox{90}{0.15}} & \text{$\Delta$ Cov} & $\text{-}0.010^{\pm 0.022}$ & $0.010^{\pm 0.001}$ & $\textcolor{gray}{\text{-}0.066}$ & $\text{-}0.006$ & $\text{-}0.029$ & $\textcolor{gray}{\text{-}0.045}$ & $\text{-}0.001$ \\
\rotatebox{90}{} & \rotatebox{90}{} & \text{PI-Width} & $\textbf{40.2}^{\pm 5.1}$ & $62.4^{\pm 0.0}$ & $\textcolor{gray}{118.8}$ & $147.6$ & $124.2$ & $\textcolor{gray}{117.1}$ & $133.3$ \\
\rotatebox{90}{} & \rotatebox{90}{} & \text{Winkler} & $\textbf{0.92}^{\pm 0.05}$ & $1.55^{\pm 0.00}$ & $\textcolor{gray}{3.07}$ & $3.08$ & $3.41$ & $\textcolor{gray}{3.43}$ & $2.99$ \\
\cmidrule{1-10} \cmidrule{2-10}
\multirow[c]{9}{*}{\rotatebox{90}{\text{Ridge}}} & \multirow[c]{3}{*}{\rotatebox{90}{0.05}} & \text{$\Delta$ Cov} & $\textcolor{gray}{\text{-}0.018}^{\pm 0.018}$ & $\text{-}0.001^{\pm 0.000}$ & $\textcolor{gray}{\text{-}0.043}$ & $\text{-}0.005$ & $\text{-}0.002$ & $\text{-}0.004$ &   \\
\rotatebox{90}{} & \rotatebox{90}{} & \text{PI-Width} & $\textcolor{gray}{60.2}^{\pm 13.1}$ & $\textbf{84.0}^{\pm 0.1}$ & $\textcolor{gray}{132.5}$ & $177.6$ & $180.5$ & $177.1$ &   \\
\rotatebox{90}{} & \rotatebox{90}{} & \text{Winkler} & $\textcolor{gray}{1.35}^{\pm 0.12}$ & $\textbf{1.82}^{\pm 0.00}$ & $\textcolor{gray}{3.11}$ & $3.31$ & $3.82$ & $3.85$ &   \\
\cmidrule{2-10}
\rotatebox{90}{} & \multirow[c]{3}{*}{\rotatebox{90}{0.10}} & \text{$\Delta$ Cov} & $\text{-}0.005^{\pm 0.026}$ & $0.011^{\pm 0.000}$ & $\textcolor{gray}{\text{-}0.051}$ & $\text{-}0.006$ & $\text{-}0.002$ & $\text{-}0.004$ &   \\
\rotatebox{90}{} & \rotatebox{90}{} & \text{PI-Width} & $\textbf{49.3}^{\pm 10.6}$ & $65.5^{\pm 0.0}$ & $\textcolor{gray}{106.2}$ & $134.3$ & $127.4$ & $125.3$ &   \\
\rotatebox{90}{} & \rotatebox{90}{} & \text{Winkler} & $\textbf{1.04}^{\pm 0.11}$ & $1.40^{\pm 0.00}$ & $\textcolor{gray}{2.54}$ & $2.66$ & $2.92$ & $2.93$ &   \\
\cmidrule{2-10}
\rotatebox{90}{} & \multirow[c]{3}{*}{\rotatebox{90}{0.15}} & \text{$\Delta$ Cov} & $0.008^{\pm 0.032}$ & $0.021^{\pm 0.000}$ & $\textcolor{gray}{\text{-}0.053}$ & $\text{-}0.007$ & $\text{-}0.001$ & $\text{-}0.003$ &   \\
\rotatebox{90}{} & \rotatebox{90}{} & \text{PI-Width} & $\textbf{42.6}^{\pm 9.3}$ & $55.6^{\pm 0.0}$ & $\textcolor{gray}{88.7}$ & $109.7$ & $102.5$ & $101.6$ &   \\
\rotatebox{90}{} & \rotatebox{90}{} & \text{Winkler} & $\textbf{0.89}^{\pm 0.10}$ & $1.20^{\pm 0.00}$ & $\textcolor{gray}{2.21}$ & $2.30$ & $2.46$ & $2.46$ &   \\
\cmidrule{1-10} \cmidrule{2-10}
\multirow[c]{9}{*}{\rotatebox{90}{\text{\text{LSTM}}}} & \multirow[c]{3}{*}{\rotatebox{90}{0.05}} & \text{$\Delta$ Cov} & $0.005^{\pm 0.005}$ & $\textcolor{gray}{\text{-}0.015}^{\pm 0.000}$ & $\textcolor{gray}{\text{-}0.023}$ & $\text{-}0.003$ & $\textcolor{gray}{\text{-}0.015}$ & $\textcolor{gray}{\text{-}0.021}$ &   \\
\rotatebox{90}{} & \rotatebox{90}{} & \text{PI-Width} & $57.1^{\pm 5.6}$ & $\textcolor{gray}{44.9}^{\pm 0.0}$ & $\textcolor{gray}{50.8}$ & $\textbf{56.0}$ & $\textcolor{gray}{48.4}$ & $\textcolor{gray}{46.1}$ &   \\
\rotatebox{90}{} & \rotatebox{90}{} & \text{Winkler} & $\textbf{1.19}^{\pm 0.08}$ & $\textcolor{gray}{1.29}^{\pm 0.00}$ & $\textcolor{gray}{1.33}$ & $1.27$ & $\textcolor{gray}{1.39}$ & $\textcolor{gray}{1.40}$ &   \\
\cmidrule{2-10}
\rotatebox{90}{} & \multirow[c]{3}{*}{\rotatebox{90}{0.10}} & \text{$\Delta$ Cov} & $0.007^{\pm 0.008}$ & $\text{-}0.017^{\pm 0.000}$ & $\textcolor{gray}{\text{-}0.028}$ & $\text{-}0.003$ & $\text{-}0.019$ & $\textcolor{gray}{\text{-}0.025}$ &   \\
\rotatebox{90}{} & \rotatebox{90}{} & \text{PI-Width} & $40.7^{\pm 4.3}$ & $\textbf{32.4}^{\pm 0.0}$ & $\textcolor{gray}{35.9}$ & $38.6$ & $34.0$ & $\textcolor{gray}{32.8}$ &   \\
\rotatebox{90}{} & \rotatebox{90}{} & \text{Winkler} & $\textbf{0.88}^{\pm 0.05}$ & $0.93^{\pm 0.00}$ & $\textcolor{gray}{0.97}$ & $0.94$ & $0.99$ & $\textcolor{gray}{0.99}$ &   \\
\cmidrule{2-10}
\rotatebox{90}{} & \multirow[c]{3}{*}{\rotatebox{90}{0.15}} & \text{$\Delta$ Cov} & $0.005^{\pm 0.011}$ & $\text{-}0.016^{\pm 0.000}$ & $\text{-}0.029$ & $\text{-}0.003$ & $\text{-}0.019$ & $\text{-}0.025$ &   \\
\rotatebox{90}{} & \rotatebox{90}{} & \text{PI-Width} & $32.5^{\pm 3.7}$ & $\textbf{26.1}^{\pm 0.0}$ & $28.4$ & $30.2$ & $26.8$ & $26.1$ &   \\
\rotatebox{90}{} & \rotatebox{90}{} & \text{Winkler} & $\textbf{0.74}^{\pm 0.04}$ & $0.76^{\pm 0.00}$ & $0.79$ & $0.77$ & $0.80$ & $0.81$ &   \\
\bottomrule
\end{tabular}
\end{table*}

\setTableCaption{Performance of the evaluated CP algorithms on the \sapflux dataset for the miscoverage levels $\alpha \in \{0.05, 0.10, 0.15\}$. The column FC specifies the prediction algorithm used for the experiment (Forest: Random Forest, LGBM: LightGBM, Ridge: Ridge Regression, LSTM: LSTM neural network). Bold numbers correspond to the best result for the respective metric in the experiment (PI-Width and Winkler score), given that $\Delta \text{Cov} \ge - 0.25 \alpha$, i.e.\ the specific algorithm reached at least approximate coverage (the result is grayed out otherwise). The error term represents the standard deviation over repeated runs (results without an error term are from deterministic models).}


\begin{table*}
\centering
\caption{\nextTableCaption}
\label{tab:sapflux-solo3-large-overall-v2}
\vskip 0.15in
\small
\begin{tabular}{lllrrrrrr|r}
\toprule
 &  &  & \rotatebox{90}{\text{\ourmethod}} & \rotatebox{90}{\text{SPCI}} & \rotatebox{90}{\text{EnbPI}} & \rotatebox{90}{\text{NexCP}} & \rotatebox{90}{\text{CopulaCPTS}} & \rotatebox{90}{\text{CP/CF-RNN}} & \rotatebox{90}{\text{AdaptiveCI}} \\
\text{FC} & \text{$\alpha$} &  &  &  &  &  &  &  &  \\
\midrule
\multirow[c]{9}{*}{\rotatebox{90}{\text{Forest}}} & \multirow[c]{3}{*}{\rotatebox{90}{0.05}} & \text{$\Delta$ Cov} & $\text{-}0.001^{\pm 0.014}$ & $\text{-}0.002^{\pm 0.000}$ & $\textcolor{gray}{\text{-}0.047}$ & $\text{-}0.003$ & $0.010$ & $0.003$ &   \\
\rotatebox{90}{} & \rotatebox{90}{} & \text{PI-Width} & $\textbf{1333.2}^{\pm 93.3}$ & $2193.8^{\pm 2.1}$ & $\textcolor{gray}{4264.1}$ & $7290.8$ & $8805.8$ & $8970.9$ &   \\
\rotatebox{90}{} & \rotatebox{90}{} & \text{Winkler} & $\textbf{0.38}^{\pm 0.01}$ & $0.74^{\pm 0.00}$ & $\textcolor{gray}{1.42}$ & $1.75$ & $2.04$ & $2.12$ &   \\
\cmidrule{2-10}
\rotatebox{90}{} & \multirow[c]{3}{*}{\rotatebox{90}{0.10}} & \text{$\Delta$ Cov} & $0.009^{\pm 0.019}$ & $0.007^{\pm 0.000}$ & $\textcolor{gray}{\text{-}0.042}$ & $0.000$ & $0.014$ & $0.005$ &   \\
\rotatebox{90}{} & \rotatebox{90}{} & \text{PI-Width} & $\textbf{1078.7}^{\pm 73.7}$ & $1741.8^{\pm 2.4}$ & $\textcolor{gray}{3671.6}$ & $6137.1$ & $7131.1$ & $7201.5$ &   \\
\rotatebox{90}{} & \rotatebox{90}{} & \text{Winkler} & $\textbf{0.30}^{\pm 0.01}$ & $0.59^{\pm 0.00}$ & $\textcolor{gray}{1.24}$ & $1.56$ & $1.76$ & $1.80$ &   \\
\cmidrule{2-10}
\rotatebox{90}{} & \multirow[c]{3}{*}{\rotatebox{90}{0.15}} & \text{$\Delta$ Cov} & $0.017^{\pm 0.024}$ & $0.014^{\pm 0.001}$ & $\text{-}0.034$ & $0.002$ & $0.017$ & $0.009$ &   \\
\rotatebox{90}{} & \rotatebox{90}{} & \text{PI-Width} & $\textbf{921.7}^{\pm 61.9}$ & $1456.7^{\pm 2.2}$ & $3200.5$ & $5261.8$ & $5980.4$ & $6062.0$ &   \\
\rotatebox{90}{} & \rotatebox{90}{} & \text{Winkler} & $\textbf{0.26}^{\pm 0.01}$ & $0.50^{\pm 0.00}$ & $1.12$ & $1.43$ & $1.57$ & $1.60$ &   \\
\cmidrule{1-10} \cmidrule{2-10}
\multirow[c]{9}{*}{\rotatebox{90}{\text{LGBM}}} & \multirow[c]{3}{*}{\rotatebox{90}{0.05}} & \text{$\Delta$ Cov} & $\textcolor{gray}{\text{-}0.027}^{\pm 0.010}$ & $\text{-}0.006^{\pm 0.000}$ & $\textcolor{gray}{\text{-}0.043}$ & $\text{-}0.006$ & $0.006$ & $\text{-}0.002$ & $0.008$ \\
\rotatebox{90}{} & \rotatebox{90}{} & \text{PI-Width} & $\textcolor{gray}{1072.4}^{\pm 65.8}$ & $\textbf{1988.0}^{\pm 1.0}$ & $\textcolor{gray}{3453.3}$ & $5737.6$ & $6929.6$ & $7087.5$ & $7160.4$ \\
\rotatebox{90}{} & \rotatebox{90}{} & \text{Winkler} & $\textcolor{gray}{0.35}^{\pm 0.00}$ & $\textbf{0.62}^{\pm 0.00}$ & $\textcolor{gray}{1.12}$ & $1.41$ & $1.66$ & $1.72$ & $1.64$ \\
\cmidrule{2-10}
\rotatebox{90}{} & \multirow[c]{3}{*}{\rotatebox{90}{0.10}} & \text{$\Delta$ Cov} & $\text{-}0.016^{\pm 0.013}$ & $0.003^{\pm 0.000}$ & $\textcolor{gray}{\text{-}0.040}$ & $\text{-}0.003$ & $0.006$ & $\text{-}0.007$ & $0.010$ \\
\rotatebox{90}{} & \rotatebox{90}{} & \text{PI-Width} & $\textbf{875.5}^{\pm 49.8}$ & $1582.3^{\pm 1.0}$ & $\textcolor{gray}{2924.1}$ & $4805.3$ & $5588.5$ & $5614.8$ & $6273.5$ \\
\rotatebox{90}{} & \rotatebox{90}{} & \text{Winkler} & $\textbf{0.27}^{\pm 0.00}$ & $0.49^{\pm 0.00}$ & $\textcolor{gray}{0.96}$ & $1.25$ & $1.43$ & $1.46$ & $1.50$ \\
\cmidrule{2-10}
\rotatebox{90}{} & \multirow[c]{3}{*}{\rotatebox{90}{0.15}} & \text{$\Delta$ Cov} & $\text{-}0.006^{\pm 0.016}$ & $0.010^{\pm 0.001}$ & $\text{-}0.036$ & $\text{-}0.002$ & $0.005$ & $\text{-}0.012$ & $0.010$ \\
\rotatebox{90}{} & \rotatebox{90}{} & \text{PI-Width} & $\textbf{751.7}^{\pm 41.8}$ & $1347.1^{\pm 1.3}$ & $2531.2$ & $4147.4$ & $4586.3$ & $4544.3$ & $5564.7$ \\
\rotatebox{90}{} & \rotatebox{90}{} & \text{Winkler} & $\textbf{0.23}^{\pm 0.00}$ & $0.42^{\pm 0.00}$ & $0.87$ & $1.14$ & $1.27$ & $1.30$ & $1.40$ \\
\cmidrule{1-10} \cmidrule{2-10}
\multirow[c]{9}{*}{\rotatebox{90}{\text{Ridge}}} & \multirow[c]{3}{*}{\rotatebox{90}{0.05}} & \text{$\Delta$ Cov} & $\textcolor{gray}{\text{-}0.021}^{\pm 0.014}$ & $\textcolor{gray}{\text{-}0.231}^{\pm 0.000}$ & $\textcolor{gray}{\text{-}0.040}$ & $\text{-}0.012$ & $\textcolor{gray}{\text{-}0.174}$ & $\textcolor{gray}{\text{-}0.331}$ &   \\
\rotatebox{90}{} & \rotatebox{90}{} & \text{PI-Width} & $\textcolor{gray}{1971.3}^{\pm 115.4}$ & $\textcolor{gray}{2539.3}^{\pm 1.6}$ & $\textcolor{gray}{3595.3}$ & $\textbf{11318.6}$ & $\textcolor{gray}{10230.3}$ & $\textcolor{gray}{8462.1}$ &   \\
\rotatebox{90}{} & \rotatebox{90}{} & \text{Winkler} & $\textcolor{gray}{0.54}^{\pm 0.05}$ & $\textcolor{gray}{3.97}^{\pm 0.01}$ & $\textcolor{gray}{1.04}$ & $\textbf{2.54}$ & $\textcolor{gray}{3.53}$ & $\textcolor{gray}{8.20}$ &   \\
\cmidrule{2-10}
\rotatebox{90}{} & \multirow[c]{3}{*}{\rotatebox{90}{0.10}} & \text{$\Delta$ Cov} & $\text{-}0.025^{\pm 0.023}$ & $\textcolor{gray}{\text{-}0.241}^{\pm 0.000}$ & $\textcolor{gray}{\text{-}0.041}$ & $\text{-}0.015$ & $\textcolor{gray}{\text{-}0.251}$ & $\textcolor{gray}{\text{-}0.358}$ &   \\
\rotatebox{90}{} & \rotatebox{90}{} & \text{PI-Width} & $\textbf{1639.6}^{\pm 93.2}$ & $\textcolor{gray}{2060.5}^{\pm 1.6}$ & $\textcolor{gray}{3117.5}$ & $10628.9$ & $\textcolor{gray}{8943.3}$ & $\textcolor{gray}{7148.7}$ &   \\
\rotatebox{90}{} & \rotatebox{90}{} & \text{Winkler} & $\textbf{0.46}^{\pm 0.05}$ & $\textcolor{gray}{2.52}^{\pm 0.00}$ & $\textcolor{gray}{0.92}$ & $2.42$ & $\textcolor{gray}{3.31}$ & $\textcolor{gray}{5.85}$ &   \\
\cmidrule{2-10}
\rotatebox{90}{} & \multirow[c]{3}{*}{\rotatebox{90}{0.15}} & \text{$\Delta$ Cov} & $\text{-}0.026^{\pm 0.031}$ & $\textcolor{gray}{\text{-}0.235}^{\pm 0.001}$ & $\textcolor{gray}{\text{-}0.042}$ & $\text{-}0.016$ & $\textcolor{gray}{\text{-}0.292}$ & $\textcolor{gray}{\text{-}0.375}$ &   \\
\rotatebox{90}{} & \rotatebox{90}{} & \text{PI-Width} & $\textbf{1427.1}^{\pm 80.7}$ & $\textcolor{gray}{1792.0}^{\pm 1.8}$ & $\textcolor{gray}{2775.8}$ & $10073.3$ & $\textcolor{gray}{7959.3}$ & $\textcolor{gray}{6155.9}$ &   \\
\rotatebox{90}{} & \rotatebox{90}{} & \text{Winkler} & $\textbf{0.42}^{\pm 0.05}$ & $\textcolor{gray}{1.93}^{\pm 0.00}$ & $\textcolor{gray}{0.85}$ & $2.34$ & $\textcolor{gray}{3.17}$ & $\textcolor{gray}{4.89}$ &   \\
\cmidrule{1-10} \cmidrule{2-10}
\multirow[c]{9}{*}{\rotatebox{90}{\text{\text{LSTM}}}} & \multirow[c]{3}{*}{\rotatebox{90}{0.05}} & \text{$\Delta$ Cov} & $0.001^{\pm 0.002}$ & $0.000^{\pm 0.000}$ & $\textcolor{gray}{\text{-}0.018}$ & $\text{-}0.001$ & $\text{-}0.012$ & $\textcolor{gray}{\text{-}0.024}$ &   \\
\rotatebox{90}{} & \rotatebox{90}{} & \text{PI-Width} & $\textbf{783.5}^{\pm 9.2}$ & $897.7^{\pm 0.9}$ & $\textcolor{gray}{1020.5}$ & $1338.9$ & $1300.1$ & $\textcolor{gray}{1194.5}$ &   \\
\rotatebox{90}{} & \rotatebox{90}{} & \text{Winkler} & $\textbf{0.25}^{\pm 0.01}$ & $0.33^{\pm 0.00}$ & $\textcolor{gray}{0.36}$ & $0.40$ & $0.45$ & $\textcolor{gray}{0.47}$ &   \\
\cmidrule{2-10}
\rotatebox{90}{} & \multirow[c]{3}{*}{\rotatebox{90}{0.10}} & \text{$\Delta$ Cov} & $0.004^{\pm 0.004}$ & $0.004^{\pm 0.000}$ & $\text{-}0.019$ & $\text{-}0.000$ & $\text{-}0.022$ & $\textcolor{gray}{\text{-}0.042}$ &   \\
\rotatebox{90}{} & \rotatebox{90}{} & \text{PI-Width} & $\textbf{594.3}^{\pm 7.7}$ & $628.2^{\pm 0.9}$ & $768.0$ & $990.0$ & $903.9$ & $\textcolor{gray}{817.2}$ &   \\
\rotatebox{90}{} & \rotatebox{90}{} & \text{Winkler} & $\textbf{0.19}^{\pm 0.01}$ & $0.24^{\pm 0.00}$ & $0.28$ & $0.32$ & $0.35$ & $\textcolor{gray}{0.36}$ &   \\
\cmidrule{2-10}
\rotatebox{90}{} & \multirow[c]{3}{*}{\rotatebox{90}{0.15}} & \text{$\Delta$ Cov} & $0.005^{\pm 0.005}$ & $0.006^{\pm 0.000}$ & $\text{-}0.019$ & $0.002$ & $\text{-}0.026$ & $\textcolor{gray}{\text{-}0.048}$ &   \\
\rotatebox{90}{} & \rotatebox{90}{} & \text{PI-Width} & $\textbf{489.9}^{\pm 7.0}$ & $493.2^{\pm 0.5}$ & $618.2$ & $780.6$ & $681.5$ & $\textcolor{gray}{620.4}$ &   \\
\rotatebox{90}{} & \rotatebox{90}{} & \text{Winkler} & $\textbf{0.17}^{\pm 0.01}$ & $0.20^{\pm 0.00}$ & $0.24$ & $0.27$ & $0.29$ & $\textcolor{gray}{0.30}$ &   \\
\bottomrule
\end{tabular}
\end{table*}

\setTableCaption{Performance of the evaluated CP algorithms and the CMAL baseline (see Appendix~\ref{sec:appendix-full-tables}) on the \hydro dataset for the miscoverage levels $\alpha \in \{0.05, 0.10, 0.15\}$. Bold numbers correspond to the best result for the respective metric in the experiment (PI-Width and Winkler score), given that $\Delta \text{Cov} \ge - 0.25 \alpha$, i.e.\ the specific algorithm reached at least approximate coverage (the result is grayed out otherwise). The error term represents the standard deviation over repeated runs (results without an error term are from deterministic models).}


\begin{table*}
\centering
\caption{\nextTableCaption}
\label{tab:hydro-531-basin-list-overall-v2}
\vskip 0.15in
\small
\begin{tabular}{llrrrrrr|r}
\toprule
 & \text{} & \rotatebox{90}{\text{\ourmethod}} & \rotatebox{90}{\text{SPCI}} & \rotatebox{90}{\text{EnbPI}} & \rotatebox{90}{\text{NexCP}} & \rotatebox{90}{\text{CopulaCPTS}} & \rotatebox{90}{\text{CP/CF-RNN}} & \rotatebox{90}{CMAL} \\
\text{$\alpha$} &  &  &  &  &  &  &  &  \\
\midrule
\multirow[c]{3}{*}{\rotatebox{90}{0.05}} & \text{$\Delta$ Cov} & $\text{-}0.002^{\pm 0.022}$ & $0.013^{\pm 0.000}$ & $\textcolor{gray}{\text{-}0.042}$ & $\text{-}0.001$ & $0.003$ & $0.006$ & $\text{-}0.004^{\pm 0.004}$ \\
 & \text{PI-Width} & $\textbf{1.91}^{\pm 0.20}$ & $2.57^{\pm 0.00}$ & $\textcolor{gray}{2.53}$ & $3.23$ & $3.44$ & $3.63$ &  $2.4^{\pm 0.08}$ \\
 & \text{Winkler} & $1.05^{\pm 0.03}$ & $1.38^{\pm 0.00}$ & $\textcolor{gray}{1.91}$ & $1.80$ & $1.93$ & $1.94$ & $\textbf{0.78}^{\pm 0.01}$ \\
\cmidrule{1-9}
\multirow[c]{3}{*}{\rotatebox{90}{0.10}} & \text{$\Delta$ Cov} & $0.001^{\pm 0.041}$ & $0.027^{\pm 0.000}$ & $\textcolor{gray}{\text{-}0.054}$ & $\text{-}0.000$ & $0.005$ & $0.009$ & $\text{-}0.004^{\pm 0.008}$ \\
 & \text{PI-Width} & $\textbf{1.39}^{\pm 0.17}$ & $1.58^{\pm 0.00}$ & $\textcolor{gray}{1.55}$ & $1.94$ & $1.99$ & $2.08$ & $1.90^{\pm 0.06}$  \\
 & \text{Winkler} & $0.79^{\pm 0.03}$ & $0.91^{\pm 0.00}$ & $\textcolor{gray}{1.27}$ & $1.21$ & $1.28$ & $1.29$ &  $\textbf{0.63}^{\pm 0.01}$ \\
\cmidrule{1-9}
\multirow[c]{3}{*}{\rotatebox{90}{0.15}} & \text{$\Delta$ Cov} & $0.003^{\pm 0.056}$ & $0.038^{\pm 0.000}$ & $\textcolor{gray}{\text{-}0.061}$ & $0.001$ & $0.005$ & $0.009$ &  $\text{-}0.006^{\pm 0.011}$  \\
 & \text{PI-Width} & $\textbf{1.11}^{\pm 0.15}$ & $1.17^{\pm 0.00}$ & $\textcolor{gray}{1.12}$ & $1.39$ & $1.39$ & $1.45$ &  $1.60^{\pm 0.05}$ \\
 & \text{Winkler} & $0.66^{\pm 0.03}$ & $0.71^{\pm 0.00}$ & $\textcolor{gray}{0.98}$ & $0.95$ & $0.99$ & $1.00$ &  $\textbf{0.55}^{\pm 0.03}$ \\
\bottomrule
\end{tabular}
\end{table*}

\end{document}